\documentclass[lettersize,journal]{IEEEtran}
\usepackage[linesnumbered,ruled,vlined]{algorithm2e}
\usepackage{amsmath,amsfonts}
\usepackage{algorithmic}
\usepackage{booktabs}
\usepackage{array}
\usepackage[caption=false,font=normalsize,labelfont=sf,textfont=sf]{subfig}
\usepackage{textcomp}
\usepackage{stfloats}
\usepackage{url}
\usepackage{verbatim}
\usepackage{graphicx}
\usepackage{cite}
\usepackage{multirow}
\usepackage{makecell}
\usepackage{threeparttable}
\usepackage{xcolor}
\usepackage{colortbl}
\usepackage{graphicx}
\usepackage{subfloat}
\begin{document}

\title{Proto-EVFL: Enhanced Vertical Federated Learning via Dual Prototype with Extremely Unaligned Data}

\author{$\text{Wei Guo}^{\dag}$, $\text{Yiyang Duan}^{\dag}$, Zhaojun Hu, Yiqi Tong, Fuzhen Zhuang*, Xiao Zhang*, Jin Dong, \\ Ruofan Wu, Tengfei Liu, Yifan Sun 
\thanks{$\dag$ indicates equal contribution, * represents the corresponding authors}
\thanks{Wei Guo, Yiyang Duan and Fuzhen Zhuang are with the School of Artificial Intelligence, Beihang University, Beijing 100083, China (e-mail: \{guowei$\underline{~}$, duanyiyang, zhuangfuzhen$\}$@buaa.edu.cn).} 
\thanks{Xiao Zhang is with the School of Computer Science and Technology, Shandong University, Shandong 266237, China (e-mail: xiaozhang@sdu.edu.cn).}
\thanks{Zhaojun Hu is with the Center for Applied Statistics, School of Statistics, Renmin University of China, Beijing 100872, China (e-mail: huzhaojun@ruc.edu.cn).}
\thanks{Yiqi Tong is with the School of Computer Science and Engineering, Beihang University, Beijing 100083, China (e-mail: yqtong@buaa.edu.cn).} 
\thanks{Jin Dong is with the Beijing Academy of Blockchain and Edge Computing, Beijing 100080, China (e-mail: dongjin@baec.org.cn).}
\thanks{Ruofan Wu is with the Department of Statistics, Fudan University, Shanghai 200433, China (e-mail: wuruofan1989@gmail.com).}
\thanks{Tengfei Liu is with the Department of Computer Science and Engineering, Hong Kong University of Science and Technology, Hong Kong 999077, China (e-mail: liutf2005@gmail.com).}
\thanks{Yifan Sun is with the Center for Applied Statistics, School of Statistics, Renmin University of China, Beijing 100872, China; Beijing Advanced Innovation Center for Future Blockchain and Privacy Computing, China (e-mail: sunyifan@ruc.edu.cn).}
}

\markboth{Journal of \LaTeX\ Class Files,~Vol.~14, No.~8, August~2024}%
{Shell \MakeLowercase{\textit{et al.}}: A Sample Article Using IEEEtran.cls for IEEE Journals}

\IEEEpubid{}

\maketitle

\begin{abstract}
 In vertical federated learning (VFL), multiple enterprises address aligned sample scarcity by leveraging massive locally unaligned samples to facilitate collaborative learning. However, unaligned samples across different parties in VFL can be extremely class-imbalanced, leading to insufficient feature representation and limited model prediction space. Specifically, class-imbalanced problems consist of \textbf{\textit{intra-party class imbalance}} and \textbf{\textit{inter-party class imbalance}}, which can further cause local model bias and feature contribution inconsistency issues, respectively. To address the above challenges, we propose Proto-EVFL, an enhanced VFL framework via dual prototypes. We first introduce class prototypes for each party to learn relationships between classes in the latent space, allowing the active party to predict unseen classes. We further design a \textit{probabilistic dual prototype learning scheme} to dynamically select unaligned samples by conditional optimal transport cost with class prior probability. Moreover, a \textit{mixed prior guided module} guides this selection process by combining local and global class prior probabilities. Finally, we adopt an \textit{adaptive gated feature aggregation strategy} to mitigate feature contribution inconsistency by dynamically weighting and aggregating local features across different parties. We proved that Proto-EVFL, as the first bi-level optimization framework in VFL, has a convergence rate of $1/\sqrt T$. Extensive experiments on various datasets validate the superiority of our Proto-EVFL. Even in a zero-shot scenario with one unseen class, it outperforms baselines by at least 6.97\%.
\end{abstract}


\section{Introduction}
\begin{figure}[t]
  \centering
  \includegraphics[scale=0.30]{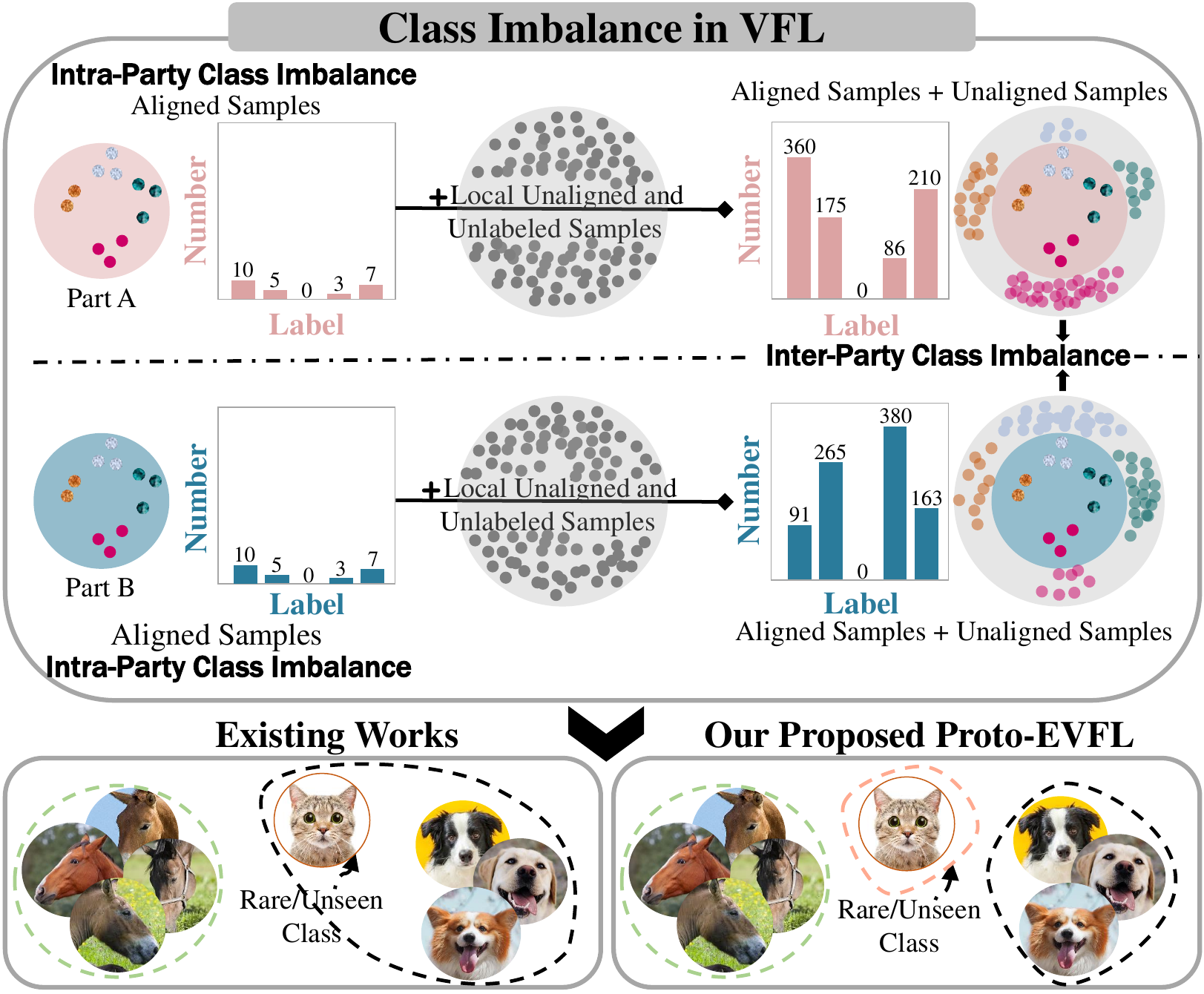}
  \caption{Intra-party class imbalance and inter-party class imbalance in VFL. Using unaligned data creates heterogeneous sample spaces among parties in VFL, causing inter-party class imbalance, and worsening existing intra-party class imbalance in only aligned data. Current VFL methods struggle to accurately classify rare or unseen classes.}
\label{fig: intro}
\end{figure}
\begin{figure*}[t]
  \centering
  \includegraphics[scale=0.27]{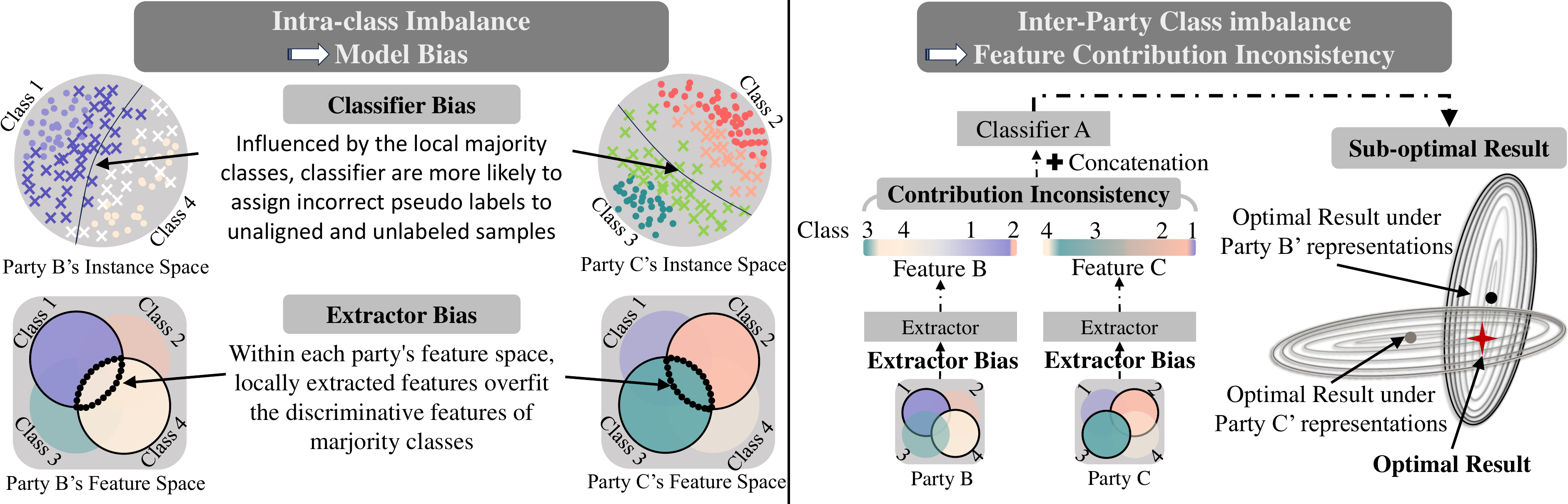}
  \caption{Model bias and inconsistent feature contribution issues in the VFL. The purple, white, green, and red represent the different classes, respectively. On the left side of the figure, points shaped as $\cdot$ denote aligned and labeled samples, and points shaped as $\times$ refer to unaligned and unlabeled samples. For features B and C on the right, the size of the area occupied by each color represents the amount of discriminant features belonging to the corresponding class. }
  \label{fig: issue}
\end{figure*}
\IEEEPARstart{F}{ederated} learning (FL) \cite{mcmahan2017communication} is a novel distributed learning paradigm addressing data isolation \cite{cheng2020federated}, where horizontal federated learning (HFL) \cite{yang2019federated} and vertical federated learning (VFL) \cite{liu2022vertical} being the most common frameworks. HFL involves parties with different samples but the same features \cite{lim2020federated}, while VFL assumes parties have the same samples but different features \cite{liu2022vertical}. For VFL, current studies mainly face two significant challenges: aligned sample scarcity and labeled sample scarcity \cite{sun2023communication,kang2022fedcvt}. In real-world scenarios, it is difficult for all parties to find the same samples due to the regional or business difference \cite{liu2022vertical,wei2022vertical}, leading to available aligned data being scarce. On the other hand, due to the high cost of labor and time, or professional barriers, unaligned data between parties is often unlabeled \cite{liu2022vertical}. 


\textbf{Limitations of Existing Methods.} In response to aligned and labeled sample scarcity of VFL, existing methods mainly focus on utilizing unaligned samples to augment the local training sets \cite{feng2022semi,kang2022fedcvt,he2022hybrid,liang2021self,castiglia2022self,li2022vertical,yang2022multi}. Nonetheless, we note that all these previous methods ignore the class-imbalanced problems when unaligned samples are utilized. The use of unaligned data disrupts the consistent sample setting of traditional VFL. This leads to biased feature representation capability and limited prediction space, causing the active-side classifier cannot predict rare or unseen classes well. Different from the class imbalance in HFL, which occurs within a uniform feature space, these problems arise in the heterogeneous feature space setting of VFL. Specifically, class imbalance in VFL includes \textbf{\textit{intra-party class imbalance}} and \textbf{\textit{inter-party class imbalance}}. As shown in Fig. \ref{fig: intro}, intra-party class imbalance refers to the class distribution imbalance within each party, while inter-party class imbalance represents a class distribution inconsistency among parties. Moreover, these class imbalances can cause local model bias and global feature contribution inconsistency issues, respectively. To illustrate these issues clearly, we visualize them in Fig. \ref{fig: issue}. Among them, model bias includes classifier bias and extractor bias. Classifier bias refers that unaligned samples are more likely to be misclassified to majority classes by a biased decision boundary. Extractor bias means that the extracted features are overfitted to local majority classes' discriminative features. Consequently, due to different intra-party class imbalances, aggregating local biased features of different parties to complete the active party's target task will further cause feature contribution inconsistency, producing sub-optimal prediction results.

\textbf{Proposed Methods.} To bridge these gaps, this paper proposes an enhanced VFL framework called Proto-EVFL via dual prototypes. Specifically, it first designs a \textit{probabilistic dual prototype learning scheme} that introduces dual optimal transport cost derived from class prior distribution to select unaligned data for augmentation, aiming to optimize feature extraction ability across parties. Furthermore, a \textit{mixed prior guided module} is proposed to guide the above selection process using estimated local and global prior probability distributions, allowing the local feature extractor to focus on discriminative features of less frequent classes. Finally, an \textit{adaptive gated feature aggregation strategy} is employed to mitigate feature contribution inconsistency, thereby improving the model generalization of the active party.

\textbf{Main Contributions:} 

\begin{itemize}
    \item 
    We take the first step to explore the class imbalance and even class invisibility problem in extremely unaligned data to address the     
    aligned sample scarcity in VFL, which could result in biased feature representation capabilities and limited model prediction space.
    \item A novel VFL framework named Proto-EVFL is proposed. It employs a probabilistic dual prototype learning scheme to enlarge local training sets, which effectively enhancing feature representation for each party and expanding the model's predictive capability and space.
    \item The proposed mixed prior guidance module combines local and global class prior probabilities to mitigate local model bias from intra-party class imbalance. Additionally, a gating network is introduced to reduce feature contribution inconsistency due to inter-party class imbalance.
    \item We provide a complete convergence proof for Proto-EVFL, the first bi-level optimization framework in VFL, with a convergence rate of $1/\sqrt{T}$. Extensive experiments demonstrate the superiority of our proposed Proto-EVFL under various class-imbalanced settings. Even in a zero-shot scenario where one class is unseen, it can outperform the baselines by at least 6.97\%.
\end{itemize}

\section{Related Work}
\label{sec:related work}
\subsection{Vertical Federated Learning}
Vertical federated learning (VFL) builds a joint model using features distributed among parties while protecting privacy \cite{yang2019federated}. In VFL, each party holds data within its own feature space, while the active party additionally possesses the labels and initiates training \cite{liu2022vertical}. Traditional VFL methods rely on large amounts of aligned samples to train models by combining features across parties \cite{feng2022vertical,wu2022practical}. However, since aligned samples are often scarce and unlabeled, some works \cite{kang2022fedcvt,feng2022semi,yang2022multi,li2022vertical,feng2022vertical,he2022hybrid} propose using abundant unaligned samples to enlarge available training sets. However, the use of unaligned data breaks the consistent sample space setting in VFL, and the differences of local unaligned samples lead to imbalanced class distribution among parties. This issue is overlooked by existing works. Therefore, we introduce a probabilistic dual prototype learning scheme to filter unaligned samples based on the estimated class prior and optimal transport cost.

\subsection{Class Imbalance}
Traditional methods like over-sampling \cite{sharma2022review}, under-sampling \cite{shelke2017review}, and HFL approaches such as filtering samples for consistent distributions \cite{Tuor2020DataSF} or clustering parties with similar distributions \cite{mansour2020three} can handle intra-party class imbalance. However, they often assume true labels are available, which is not applicable for VFL with no labels among passive parties. Moreover, these HFL methods are implemented in the consistent feature space, which conflicts with heterogeneous feature space in VFL. To address this challenge, we propose a dual optimal transport distance based on conditional probability and a mixed prior module to avoid biased extractors caused by the intra-party class imbalance. Furthermore, we introduce an adaptive gated feature aggregation strategy to prevent inconsistent feature contributions caused by inter-party class imbalance in the aggregation process of the active party. 

\subsection{Zero-shot Learning}
Zero-shot learning recognizes unseen classes by bridging them with seen classes using semantics \cite{lampert2013attribute,larochelle2008zero,huang2015learning}. Current zero-shot learning works can be divided into three directions: direct attribute prediction-based \cite{6552193,NIPS2014_1f1baa5b,6571196}, embedded model-based \cite{8451907,8411154,8272453}, and generative model-based approaches \cite{DBLP:journals/corr/abs-1712-00981,Schonfeld_2019_CVPR}. The embedded model-based approach is the most common due to its straightforwardness and efficiency in forming the mapping relationship between semantic features and sample features, which is also applied to this paper. We introduce class prototypes as semantic information to learn the relationship with sample intermediate features, and then leverage these prototypes to design a prototype learning scheme in VFL with heterogeneous feature spaces. Unlike existing VFL works \cite{wu2022practical,liu2022vertical}, our method effectively improves the predictive capability for extreme class-imbalanced scenarios with unseen classes.

\subsection{Prototype Learning}
Prototype learning has been widely used in HFL to tackle data heterogeneous challenges. Specifically, FedProto\cite{tan2022fedproto} uses class prototypes to improve the heterogeneity tolerance. FedFA \cite{zhou2023fedfa} aligns class prototypes and classifiers across different parties during the local training process. The studies \cite{xu2023personalized,dai2023tackling} address local-global prototype alignment by using global semantic knowledge or contrastive learning methods \cite{mu2023fedproc,yu2023contrastive}. However, these methods only work in HFL with homogeneous feature spaces, allowing direct data distribution alignment via global prototypes. They could not apply to VFL with heterogeneous prototypes among parties. Meanwhile, these methods fail to explicitly consider class probability distributions when using prototypes to define loss functions. 

\section{Problem definition}
In a VFL setting with $M$ parties, each party collaborates to obtain the intermediate representation of aligned data $f_{a}$ for the active party's task \cite{huang2023vertical,ren2022improving}. The active party serves as a task curator ($m=1$), and each party maintains its private training data $X^{m} \in \mathbb{R}^{N^{m}\times D^{m}}$ ($m = 1,2, ..., M$), where $N^{m}$ and $D^{m}$ are the number of samples and feature dimensions of party $m$, respectively. Each party's training set is divided into an aligned dataset $X^{m}_{a} \in \mathbb{R}^{N_{a}\times D^{m}}$ and an unaligned dataset $X^{m}_{u} \in \mathbb{R}^{N^{m}_{u}\times D^{m}}$ ($N^{m} = N_{a} + N^{m}_{u}$). Aligned samples across parties have identical IDs, which can be identified using techniques from previous works \cite{nock2018entity,hardy2017private}. The active party only holds aligned samples' ground truth labels $Y_{a} \in {(0, 1)}^{N_{a}\times Z}$, where $Z$ represents the number of classes. Since aligned data $N_{a}$ is often scarce in reality, each party can enlarge its training set with local unaligned and unlabeled samples $X^{m}_{u}$ to enhance its feature extraction ability of local extractor $E_{m}$. However, using unaligned data in VFL breaks the assumption of sample space consistency across parties, resulting in heterogeneous sample space within the heterogeneous feature space: 
\begin{small}
\begin{equation}
\label{equ:inconsistent sample space}
    \mathcal{F}^{m} \neq \mathcal{F}^{m+1}, \mathcal{I}_{a+u}^{m} \neq \mathcal{I}_{a+u}^{m+1}, \mathcal{Y}_{a+u}^{m} = \mathcal{Y}_{a+u}^{m+1}, \forall m \in (0,M-1)
\end{equation}
\end{small}

\noindent where $\mathcal{F}^{m}$, $\mathcal{I}_{a}$, $\mathcal{I}_{a+u}^{m}$ represent the feature space, the sample space of only using aligned data, the new sample space of using unaligned data of party $m$, respectively.

Proto-EVFL facilitates knowledge sharing between parties using prototypes without gradient transfer, which mitigates privacy leakage issues in VFL. Therefore, different from existing VFL models, the training process of Proto-EVFL can be defined as a bi-level optimization problem:
\begin{small}
\begin{equation}
    F(\theta_{C},\theta^{*}_{E}) = \mathop{\rm min}_{\theta_{C}}\frac{1}{N_{a}}\sum^{N_{a}}_{i=1}\mathcal{L}_{\rm global}(F(\theta^{*}_{E};x^{1}_{a,i},x^{2}_{a,i},...,x_{a,i}^{M});y_{a,i}),
\end{equation}
\end{small}

\begin{footnotesize}
\begin{equation}
    s.t. \theta^{*}_{E^{m}} = \arg\min_{\theta_{E^{m}}}\frac{1}{N^{m}}\sum^{N^{m}}_{i=1}\mathcal{L}^{m}_{\rm local}(F(\theta_{E^{m}};x^{1}_{a,i},x^{2}_{a,i},...,x_{a,i}^{M});\mu^{m}),
\end{equation}
\end{footnotesize}

\noindent where $\mathcal{L}_{\rm global}$ and $\mathcal{L}^{m}_{\rm local}$ are loss functions of the active party and passive party $m$, respectively. $\theta_{C}$ denotes the parameter of the active-party side classifier. For party $m$, $\theta_{E^{m}}$ is the extractor parameter. $\theta^{*}_{E^{m}}$ represents the optimal solution. $\theta^{*}_{E}$ is the set of $\theta^{*}_{E^{m}}$ for all parties. $\mu^{m}$ denotes class prototypes. We summarize our notations in Appendix \ref{sec:notations}.

\section{Methodology}
\begin{figure*}[ht]
  \centering
  \includegraphics[scale=0.17]{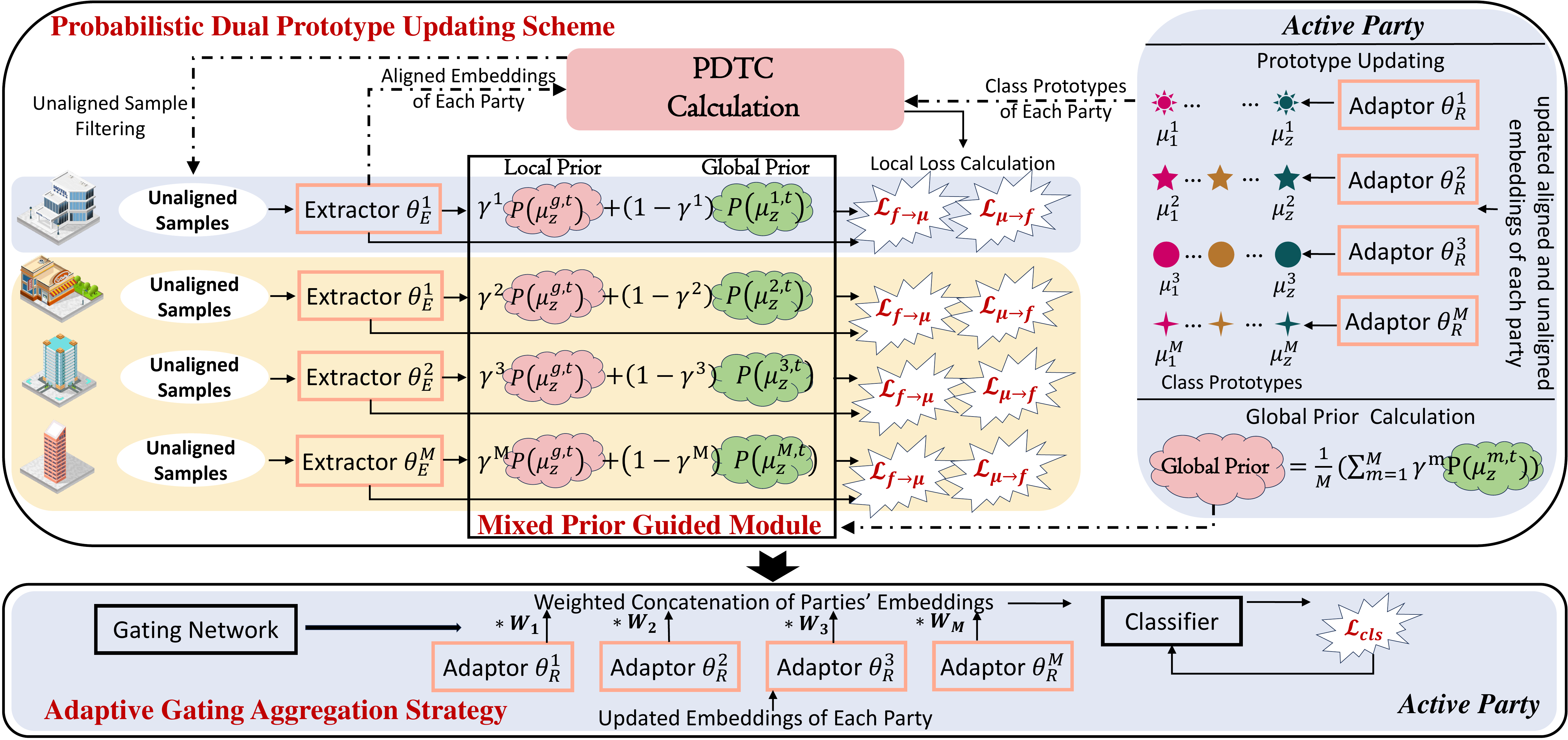}
  \caption{The overview framework of Proto-EVFL. The diagram shows the active party on a blue background and the passive party on a yellow background. The active party obtains each party's prototypes using their aligned data embeddings. Parties train their feature extractors with PDTC based on a mixed prior guidance module and optimized prototypes. Finally, the active party optimizes the target classifier with an adaptive gated feature aggregation strategy.}
  \label{fig: framework}
\end{figure*}
\subsection{Overview}
The overall framework of our proposed Proto-EVFL is shown in Fig. \ref{fig: framework}, which consists of three major components: probabilistic dual prototype learning scheme, mixed prior guided module, and adaptive gated feature aggregation strategy. Specifically, each party utilizes probabilistic dual transport cost (PDTC) to filter unaligned samples to obtain an unbiased local extractor, extracting intermediate representations of aligned samples. In this process, updating the local class prior distribution synchronously mixes the global prior, mitigating the bias of imbalanced local class prior in unaligned sample selection. Finally, an adaptive gated feature aggregation strategy is designed to adjust parties' feature contributions for classifier optimization. We describe details of the components in the following sections.
\subsection{Probabilistic Dual Prototype Leaning Scheme}
\label{sec:dual prototype}
Due to the scarcity of alignment samples, it is natural to leverage local unaligned data to enlarge the available training set of local extractors. For the first time, we introduce the optimal transport (OT) distance to address the unaligned sample selection problem. This method incorporates the local class prior distribution into the loss function to control the influence of class imbalance on extractor optimization, alleviating the biased problem caused by intra-party imbalanced distribution. Existing OT methods \cite{chiang2023optimal,li2024global} calculate the distance between distributions by measuring a single directional distance. Different from them, inspired by \cite{sun2024spectr}, which computes the distance between distributions using conditional probability, we apply a dual form of OT, including the distance from samples to prototypes and the distance from prototypes to samples. This aims to improve the confidence of pseudo-labels while mitigating the class omission problem. 

The proposed scheme assumes that the party's data distribution consists of separated clusters, with samples in a cluster likely sharing the same labels. Thus, each cluster's center acts as a class prototype. To aggregate global feature information, we update each party's prototypes on the active party, then use these updated prototypes to select local unaligned samples globally. This scheme aims to minimize the expected PDTC between unaligned representations and class prototypes through the following two dual steps:

\subsubsection{From Unaligned Representations to Class Prototypes}
The expected transport cost from unaligned representations $f^{m,n}_{u}$ ($n \in (1, N^{m}_{u})$) to class prototypes $\mu^{m}_{z}$ ($z \in (1, Z)$) can be factorized into the joint distribution $p(f^{m,n}_{u}\pi_{\theta_{E}}(\mu^{m}_{z}|f^{m,n}_{u}))$ using the chain rule. On the other hand, unaligned representation $f^{m,n}_{u}$ is obtained by randomly selecting an unaligned sample $X^{m,n}_{u} \sim X^{m}_{u}$ and applying function $F_{\theta_{E}}(X^{m,n}_{u})$. The conditional distribution $\pi_{\theta_{E}}(\mu^{m}_{z}|f^{m,n}_{u})$ represents the probability of moving from $f^{m,n}_{u}$ to $\mu^{m}_{z}$ for party $m$:
\begin{equation}
    \pi_{\theta_{E}}(\mu^{m}_{z}|f^{m,n}_{u}) = \frac{p(\mu^{m}_{z}){\rm exp}(\mu^{m,t}_{z}f^{m,n}_{u})}{\sum^{Z}_{z'=1}p(\mu^{m}_{z'}){\rm exp}(\mu^{m,t}_{z'}f^{m,n}_{u})},
\end{equation}
where $t$ denotes the $t$-th communication round ($t \in (1, T)$). Based on Bayes' theorem \cite{joyce2003bayes}, $p(\mu^{m}_{z})$ represents the discrete prior probability distribution across $Z$ classes. As the unaligned samples are unlabeled, the local class prior probability $\{p(\mu^{m}_{z})\}^{Z}_{z=1}$ is unavailable and will be estimated using the mixed prior module (see Section \ref{sec:mixed prior}). And, ${\rm exp}(\mu^{m,t}_{z}f^{m,n}_{u})$ measures the similarity between target class prototypes and unaligned representations. ${p(\mu^{m,t}_{z})}^{Z}_{z=1}$ will be further estimated according to Eq. \ref{equ:mixed prior}. Overall, the expected cost of moving unaligned representations to prototypes is expressed as:
\begin{equation}
\label{equ:Lf--mu}
    \mathcal{L}^{m}_{f \rightarrow \mu} = \mathbb{E}_{f^{m,n}_{u} \sim f^{m}_{u}}\mathbb{E}_{\mu^{m}_{z}\sim \pi_{\theta_{E}}(\mu^{m}_{z}|f^{m,n}_{u}))}[c(\mu^{m}_{z}, f^{m,n}_{u})],
\end{equation}
where $c(\mu^{m}_{z}, f^{m,n}_{u})$ is defined by the cosine dissimilarity.
\subsubsection{From Class Prototypes to Unaligned Representations}
The above step minimizes entropy (see Appendix \ref{sec:entropy minimization}). However, entropy minimization alone may overlook certain patterns (class prototypes) \cite{morerio2017minimal,wu2020entropy}. Therefore, we add a reverse direction loss (from class prototypes to unaligned representations) to ensure each class prototype has nearby data points and prevents omitting any class prototypes, meanwhile improving the confidence of results. Given a mini-batch of unaligned samples $\{X^{m,n}_{u}\}_{n=1}^{B}$, where $B$ represents the batch size, we calculate the probabilities of moving from class prototypes to unaligned representations for part $m$ as follows:
\begin{equation}
    \pi_{\theta_{E}}(f^{m,n}_{u}|\mu^{m}_{z}) = \frac{p(\mu^{m}_{z}){\rm exp}(\mu^{m,t}_{z}f^{m,n}_{u})}{\sum^{B}_{n=1}p(\mu^{m}_{n}){\rm exp}(\mu^{m,t}_{z}f^{m,n}_{u})}.
\end{equation}
We further provide the expected cost of moving from class prototypes to unaligned representations:
\begin{equation}
\label{equ:Lmu--f}
\begin{split}
    \mathcal{L}^{m}_{\mu \rightarrow f} =& \mathbb{E}_{\{f^{m,n}_{u}\}^{B}_{n=1} \sim f^{m}_{u}}\mathbb{E}_{\mu^{m}_{z} \sim p(\mu^{m}_{z})}\mathbb{E}_{f^{m,n}_{u}\sim \pi_{\theta_{E}}(f^{m,n}_{u})|\mu^{m}_{z})}\\
    &[c(\mu^{m}_{z}, f^{m,n}_{u})].
\end{split}
\end{equation}
Finally, the extractor $E^{m}$ is updated by the local loss $\mathcal{L}_{\rm local}$:
\begin{equation}
    \mathcal{L}^{m}_{\rm local} = \mathcal{L}_{f \rightarrow \mu} + \mathcal{L}_{\mu \rightarrow f} + \frac{\varphi}{2}{\Vert\theta_{E^{m}}\Vert}^{2}.
\end{equation}
The last regularization term restricts the local extractor's updates avoiding overfitting, where $\varphi$ controls this term. The class prototypes are updated by Eq.\ref{equ:prototype update} in Section \ref{sec: moe}.

\subsection{Mixed Prior Guided Module}
\label{sec:mixed prior}
To update the local class prior, we incorporate the global prior into a mixed prior. The active party averages the local priors $\sum^{M}_{m=1}P(\mu^{m,t}_{z})$ to obtain the global prior probability $P(\mu^{g,t}_{z})$. This global prior is then combined with the local prior to balance the occurrence probability of each class globally, provided the local minority class sample size is not extremely small. However, to prevent minority classes from being assigned excessively high prior probabilities, especially when local data for minority classes is insufficient, we introduce a personalized parameter $\gamma^{m}$ to limit the influence of the global prior on the local priors. According to the law of the minimum principle \cite{de1994liebig}, we define $\gamma^{m}$ as $\frac{N^{m}_{\hat{z}}}{N^{m}}$ according to the number of local minority class $\hat{z}$. Since the labels of local unaligned data are unknown, the party adaptively calculates $\gamma^{m}$ based on the classification results from PTDC at each round.


Specifically, since maximizing the log-likelihood of unaligned samples is intuitive \cite{goldman2000enhancing} but directly optimizing the marginal likelihood is intractable, we first estimate local prior probability by designing an EM algorithm \cite{saerens2002adjusting} for inferring an iterative updating process (see Appendix \ref{sec:prior estimation}). We initialize local class prior probability as $P(\mu^{m,0}_{z}) = \frac{1}{Z}$ and update prior probability of the current sample distribution at round $t$ based on the similarity between its current features and prototypes:
\begin{equation}
\label{equ:estimate prior}
    P(\mu^{m,t}_{z}) = \frac{1}{B}\sum^{B}_{n=1}\pi_{\theta_{E}^{t-1}}(\mu^{m}_{z}|f^{m,n}_{u}),
\end{equation}
\begin{equation}
    \pi_{\theta_{E}}^{t-1}(\mu^{m}_{z}|f^{m,n}_{u}) = \frac{p(\mu^{m,t-2}_{z}){\rm exp}(\mu^{m,t-2}_{z}f^{m,n}_{u})}{\sum^{Z}_{z'=1}p(\mu^{m,t-2}_{z'}){\rm exp}(\mu^{m,t-2}_{z'}f^{m,n}_{u})},
\end{equation}
After receiving the global prior probability $P(\mu^{g,t}_{z})$ and estimating the local prior probability of round $t+1$, each party further updates the local prior probability by:
\begin{equation}
\label{equ:mixed prior}
    P(\mu^{m,t+1}_{z}) = \gamma^{m} P(\mu^{g,t}_{z}) + (1 - \gamma^{m})P(\mu^{m,t+1}_{z}).
\end{equation}
The probabilistic dual prototype learning scheme relies on this updated prior probability $P(\mu^{m,t+1}_{z})$ to achieve the selection of local unaligned samples and the optimization of extractor.

\IncMargin{1em}
\begin{algorithm}[t]
  \KwIn{communication round $T$; global prior probability influence ratio: $\gamma^{m}$; learning rate: $\eta$ and $\eta^{\prime}$. batch: $\mathcal{B}$. epoch: $\tau$.}
  \SetKwData{Left}{left}\SetKwData{This}{this}\SetKwData{Up}{up}
  \SetKwFunction{Union}{union}\SetKwFunction{FindCompress}{findcompress}
	 \BlankLine 
	 \For{$t \in \{1, ..., T\}$}{
    	 parties receive prototypes $\mu^{m}$ and global prior $P(\mu^{g})$ from active party\;
    	 \For{\rm{party} $m =1, ..., M$ \rm{in parallel}}{
                update local class prior via Eq.\ref{equ:mixed prior}\;
                compute $\mathcal{L}_{\mu \rightarrow f}$ and $\mathcal{L}_{f \rightarrow \mu}$\ via Eq.\ref{equ:Lf--mu} and Eq.\ref{equ:Lmu--f}\;
                 \For{j= 1, ..., $\tau$}{
                         Draw a sample batch $\mathcal{B}_{j-1}$\;
                 $\theta_{E^{m}} \leftarrow \theta_{E^{m}} - \eta\nabla_{\theta_{E^{m}}}\mathcal{L}_{{\rm local}}$\;
                 }
                 update representations of aligned samples $f_{a}^{m} \leftarrow E^{m}(X^{m}_{a})$
                    }
            active party ($m = 1$) receives $f_{a}^{m}$ and $P(\mu^{m})$ from party $m =1, ..., M$\;
                \For{$m \in \{1, ..., M\}$, \rm{active party}}{  
                         compute $\hat{f}_{a}^{m}$ via Eq.\ref{equ:adaptor update}\;
                         $W \leftarrow$ aggregated weight\;
                         aggregate $\hat{f}_{a}^{m}$ by weight $W$\;
                         compute $\mathcal{L}_{\rm global}$ via Eq.\ref{equ:cls}\;
                         \For{$j$= 1, ..., $\tau$}{
                         Draw a sample batch $\mathcal{B}_{j-1}$\;
                         $\theta_{R^{m},G,C} \leftarrow \theta_{R^{m},G,C} - \eta^{\prime}\nabla_{\theta_{R^{m}},G,C}\mathcal{L}_{\rm global}(f_{a},y_{a};\mathcal{B}_{j-1})$\;
                         }
                         update prototypes for each party via Eq.\ref{equ:prototype update}  \;
                         compute global prior $P(\mu^{g}_{z})$\;                 
            }
     }
 	\caption{Proto-EVFL}
        \label{algorithm flowchart}
 	 \end{algorithm}
 \DecMargin{1em}
 
\subsection{Adaptive Gated Feature Aggregation Strategy}
\label{sec: moe}
Given that the selected local unaligned samples varied in each communication round, the class probability distribution of the available training set also dynamically changed. It leads to the changing feature contribution inconsistency by aggregating representations from these extractors in the active party. Therefore, existing feature aggregation methods \cite{wu2022practical,wang2024unified,castiglia2023flexible} in VFL could not be directly applied. Motivated by the mixture of expert (MoE) \cite{yuksel2012twenty}, we first introduce MoE into the VFL aggregation process to dynamically concatenate each party's uploaded intermediate representations based on a parameterized gating network $G$. It adjusts the contribution of these representations by weights $W$. Specifically, the active party first introduces adaptors $R$ as expert models for each party. The aligned samples' representations $f^{m}_{a}$ of party $m$ are transformed by its adapter to produce new representations:
\begin{equation}
\label{equ:adaptor update}
    \hat{f}^{m}_{a} = R^{m}(f^{m}_{a}, \theta_{R^{m}}).
\end{equation}
To better preserve local information, the dimensions of transformed representations $\hat{f}^{m}_{a}$ are kept fixed. Therefore, the learning process of $W$ is expressed by:
\begin{equation}
    G(\hat{f}^{m}_{a}) = {\rm Softmax}(\hat{f}^{m}_{a} \cdot W),
\label{equ:weight aggregation}
\end{equation}
$W$ is a trainable weight matrix to dynamically adjust concatenation weights at each communication round on the active party side. 

Finally, we concatenate intermediate representations $\sum^{M}_{m=1}\hat{f}_{a}^{m}$ to obtain $\hat{f}_{a}$, aiming to train classifier. The optimization objective of the active party is written as:
\begin{equation}
\label{equ:cls}
    \mathcal{L}_{{\rm global}} = \omega((W \cdot \hat{f}_{a}, \theta_{C}), Y_{a}),
\end{equation}
where $\omega$ denotes the cross-entropy loss. After completing the above updating, at $t$-th communication round, the active party updates the class prototype $\mu^{m,t}_{z}$ of party $m$ using feature representations $\hat{f}^{m}_{a}$ by:
\begin{equation}
    \mu^{m,t}_{z} = \mu^{m,t-1}_{z} + \frac{\rho}{N_{a,z}}(\sum^{N_{a,z}}_{n=1}\hat{f}^{m}_{a,z}), 
\label{equ:prototype update}
\end{equation}
which aims to incorporate partial global feature information into local prototypes, facilitating local PDTC calculation and unaligned data selection in the next round. $\rho$ adjusts the update speed. Finally, the overall workflow of our proposed Proto-EVFL is shown in Algorithm \ref{algorithm flowchart}. 

\subsection{Convergence Analysis}
In this section, we provide a complete convergence analysis of our proposed Proto-EVFL. Existing VFL approaches \cite{he2022hybrid,wu2022practical,castiglia2023flexible,castiglia2022self,wang2024unified,castiglia2023less} require the active party to return gradient information to passive parties for local updates, introducing a high privacy leakage risk under current label inference attacks \cite{fu2022label} and feature inference attacks \cite{yang2023practical}. In contrast, Proto-EVFL utilizes class prototypes to avoid transmitting gradient information, which essentially eliminates the possibility of existing label and feature inference attacks. 

The optimization process of Proto-EVFL can be viewed as a bi-level optimization problem, for which we provide complete convergence proof in the following. For clarity, we introduce some new notations. Let $ {\Theta}:= \{ {\theta}_G, {\theta}_C, {\theta}_{R^1}, ..., {\theta}_{R^{M}}\}$ and ${\mathcal{E}}:=\{ {\theta}_{E^{1}}, ..., {\theta}_{E^{M}}\}$. Our bi-level optimization problem can be written as follows:
\begin{equation}
    \begin{aligned}
         \min_{ {\Theta}} F( {\Theta})&:= \mathcal{L}_{{\rm global}}( {\Theta}, {\mathcal{E}}^*( {\Theta}))=\mathbb{E}_{\xi}[l_{{\rm global}}( {\Theta}, {\mathcal{E}}^*( {\Theta});\xi)]\\
         &=\frac{1}{N_a}\sum_{i=1}^{N_a}l_{{\rm global}}( {\Theta}, {\mathcal{E}}^*( {\Theta});\xi_i)\\
     \text{s.t.} \quad \mathcal{E}^*(\Theta)&=\arg\min_{\mathcal{E}} \mathcal{L}_{{\rm local}}( \Theta,\mathcal{E})=\mathbb{E}_{\zeta}[l_{{\rm local}}( {\Theta}, {\mathcal{E}};\zeta)] \\
     &= \frac{1}{N_u}\sum_{i=1}^{N_u}l_{{\rm local}}( {\Theta}, {\mathcal{E}};\zeta_i),
    \end{aligned}\label{bilevel_opt}
\end{equation}
where $l_{{\rm local}}( {\Theta}, {\mathcal{E}};\zeta):=l_{{\rm local}}({\Theta}, {\mathcal{E}};\zeta) + \frac{\varphi}{2} \| \mathcal{E}\|_F^2$ and $\zeta:=\{X_{u}^1,X_{u}^2,...,X_{u}^M\}$ denote unaligned data from $M$ parties.

Let $\mathcal{Z}:= \{\Theta, \mathcal{E}\}$ denote all model parameters, we then make some standard assumptions \cite{ghadimi2018approximation,ji2020convergence,rajput2020closing} on $l_{{\rm global}}$ and $l_{{\rm local}}$ for our bi-level optimization problem Eq. \ref{bilevel_opt}.

\noindent \textbf{Assumption 1. (Lipschitz Condition)} The loss function $l_{{\rm global}}(\mathcal{Z};\xi)$ is $L_1$-Lipschitz for any given $\xi$.

\noindent \textbf{Assumption 2. (Smoothness)} The loss functions $l_{{\rm global}}(\mathcal{Z};\xi)$, $l_{{\rm local}}(\mathcal{Z};\xi)$ and $l(\mathcal{Z};\xi)$ are $L_2$-smooth, $L_2$-smooth and $L$-smooth for any give $\xi$ and $\zeta$, respectively.

\noindent \textbf{Assumption 3. (Lipschitz Condition for Second Derivatives)} The second derivatives $\nabla_{\Theta}\nabla_{\mathcal{E}}l_{{\rm local}}(\mathcal{Z};\zeta)$ and $\nabla_{\mathcal{E}}^2 l_{{\rm local}}(\mathcal{Z};\zeta)$ are $L_3$-Lipschitz and $L_4$-Lipschitz for any give $\zeta$, respectively.

\noindent \textbf{Assumption 4. (Bounded Variance)} Stochastic gradients $\nabla l_{{\rm local}}(\mathcal{Z};\zeta)$ has a bounded variance $\sigma^2$, i.e., for any $\mathcal{Z}$, we have
\begin{equation*}
    \mathbb{E}_{\zeta} \|\nabla l_{{\rm local}}(\mathcal{Z};\zeta) - \nabla \mathcal{L}_{{\rm local}}(\mathcal{Z})\|^2 \leq \sigma^2.
\end{equation*}

\noindent \textbf{Assumption 5. (Bounded Domain)} The parameter $\mathcal{E}$ is in a bounded domain with a diameter $\Delta$, i.e., for any $\mathcal{E}_1$ and $\mathcal{E}_2$, we have
\begin{equation*}
    \| \mathcal{E}_1 - \mathcal{E}_2\|\leq \Delta.
\end{equation*}

To characterize the convergence performance of our algorithm, we first introduce the following definition.

\noindent\textbf{Definition 1.} We call $\hat{\Theta}$ is an $\epsilon$-accurate stationary point for the objective function $F(\Theta)$ if $\mathbb{E}\|\nabla F(\hat{\Theta})\|^2 \leq \epsilon$.

Given the above definitions, we present our convergence:

\noindent\textbf{Theorem 1.} Under Assumptions 1-5, define $\alpha:=-L+\varphi$, choose step size $\eta$ to be $\frac{2}{L_2 + \alpha}$, $B=\mathcal{O}(\frac{1}{\sqrt{\epsilon}})$, $\tau=\mathcal{O}(\log\frac{1}{\epsilon})$, $\eta^\prime <\frac{1}{2L_0}$ and suppose $\alpha<L_2$, we have
\begin{equation}
    \begin{aligned}
        \frac{1}{\tau T}&\sum_{t=0}^{T-1}\sum_{j=0}^{\tau-1}\|\nabla F(\Theta_t^{j})\|^2 \leq \frac{F(\Theta_0^0) - \inf_{\Theta}F(\Theta)}{\tau T (\frac{\eta^{\prime}}{2} -L_0 \eta^{\prime 2}) }\\
        &+(\eta^\prime + 2\eta^{\prime 2}L_0)L_2^2\Delta^2 \frac{\tau -1}{\tau}+\mathcal{O}(\epsilon),
    \end{aligned}\label{theorem_1}
\end{equation}
where $\mathcal{O}(\cdot)$ indicates that the function's growth rate is proportional to or slower than $\epsilon$. We define $L_0$ in Eq. \ref{equ:L0}.

If we choose $\tau=1$ in Theorem 1, we can see our convergence rate is $\mathcal{O}(\frac{1}{\sqrt{T}})$ which is the classic rate in VFL.

We will further prove Theorem 1 and provide more general results about arbitrary $\tau$ and $B$ in the next. To provide the convergence analysis of our algorithm, we need to give some useful lemmas first.

Firstly, with $L$-smoothness of $l(\Theta,\mathcal{E};\zeta)$, we could give the strongly convexity property of $l_{{\rm local}}(\Theta,\mathcal{E};\zeta)$.

\noindent\textbf{Lemma 1.} Under Assumption 2, suppose $\alpha:=-L+\varphi >0$, $l_{{\rm local}}(\Theta,\mathcal{E};\zeta)$ is $\alpha$-strongly convex w.r.t $\mathcal{E}$. 

Based on Assumptions 1 and 2, we could give Lemma 2 directly.

\noindent\textbf{Lemma 2.} Under Assumption 1 and 2, the derivatives $\nabla l_{{\rm global}} (\mathcal{Z}; \xi)$, $\nabla_{\Theta}\nabla_{\mathcal{E}}l_{{\rm local}}(\mathcal{Z};\zeta)$ and $\nabla_{\mathcal{E}}^2 l_{{\rm local}}(\mathcal{Z};\zeta)$ have bounded variances, i,e., for any $\mathcal{Z}$, we have:
\begin{equation}
    \mathbb{E}_{\xi} \|\nabla l_{{\rm global}} (\mathcal{Z};\xi) - \nabla \mathcal{L}_{{\rm global}} (\mathcal{Z})\|^2 \leq L_1^2,\label{cls_bound}
\end{equation}
\begin{equation}
    \mathbb{E}_{\zeta} \|\nabla_{\Theta}\nabla_{\mathcal{E}}l_{{\rm local}} (\mathcal{Z};\zeta) - \nabla_{\Theta}\nabla_{\mathcal{E}}\mathcal{L}_{{\rm local}} (\mathcal{Z})\|^2 \leq L_2^2,\label{lv_bound1}
\end{equation}
\begin{equation}
    \mathbb{E}_{\zeta} \|\nabla_{\mathcal{E}}^2 l_{{\rm local}} (\mathcal{Z};\zeta) - \nabla_{\mathcal{E}}^2\mathcal{L}_{{\rm local}} (\mathcal{Z})\|^2 \leq L_2^2.\label{lv_bound2}
\end{equation}

As the proof of Lemma 1 and 2 is too trivial, we omit it. To show the smooth property of $F (\Theta)$, we first introduce the following lemma which is proposed in \cite{ghadimi2018approximation}.

\noindent\textbf{Lemma 3. (Lemma 2.2 in \cite{ghadimi2018approximation})} Under Assumptions 1, 2 and 3, $F (\Theta)$ is $L_0$-smooth where $L_0$ is given by:
\begin{small}
\begin{equation*}
    L_0:= L_2 +\frac{2L_2^2 + L_1^2 L_3}{\alpha}+ \frac{L_1L_2L_3+L_1L_2L_4+L_2^3}{\alpha^2}+\frac{L_1L_2^2L_4}{\alpha^3}.
    \label{equ:L0}
\end{equation*}
\end{small}

Tracking error $\mathbb{E}\|\mathcal{E}_t^{j-1} - \mathcal{E}^*(\Theta_t^0)\|$ is an important component in our convergence analysis. To give an upper bound on the tracking error, we utilized Lemma 9 in \cite{ji2021bilevel}.

\noindent\textbf{Lemma 4. (Lemma 9 in \cite{ji2021bilevel})} Under Assumptions 1, 2 and 4, with step size $\eta$ to be $\frac{2}{L_2 + \alpha}$, we have:
\begin{small}
\begin{equation}
    \mathbb{E}\|\mathcal{E}_t^{j-1} - \mathcal{E}^*(\Theta_t^0)\|^2 \leq \left(\frac{L_2 - \alpha}{L_2 +\alpha}\right)^{2(j-1)}\mathbb{E}\|\mathcal{E}_t^{0} - \mathcal{E}^*(\Theta_t^0)\|^2 + \frac{\sigma^2}{L_2\alpha B}. \label{tracking_error}
\end{equation}
\end{small}

With the help of the above lemmas, we could give the estimation property of the $\frac{\partial l_{{\rm global}} (\Theta_t^0,\mathcal{E}_t^\tau)}{\partial \Theta_t^0}$ approximating $\nabla F(\Theta_t^0)$. The result is presented in the following Proposition 1. For the sake of clarity, we introduce a new notation $\frac{\partial l_{{\rm global}} (\Theta_t^0,\mathcal{E}_t^\tau;\mathcal{B}_0)}{\partial \Theta_t^0}:= \frac{1}{B}\sum_{i \in \mathcal{B}_0} \frac{\partial l_{{\rm global}} (\Theta_t^0,\mathcal{E}_t^\tau;\xi_i)}{\partial \Theta_t^0}$. Other notations involved $\mathcal{B}_j$ or $\mathcal{B}_j$ such as $\nabla_{\Theta}\nabla_{\mathcal{E}}l_{{\rm local}}(\Theta_t^0,\mathcal{E}_t^{j-1};\mathcal{B}_{j-1})$ have similar meanings.

\noindent\textbf{Proposition 1.} Under Assumptions 1-5, choose step size $\eta$ to be $\frac{2}{L_2 + \alpha}$ and suppose $\alpha<L_2$, we have:
\begin{small}
\begin{equation}
    \begin{aligned}
        \mathbb{E}&\|\frac{\partial l_{{\rm global}} (\Theta_t^0,\mathcal{E}_t^\tau;\mathcal{B}_0)}{\partial \Theta_t^0} - \nabla F(\Theta_t^0)\|\leq (L_2 + \frac{L_2^2}{\alpha}\Big [\left(\frac{L_2 - \alpha}{L_2 +\alpha}\right)^{\tau}\\
        &\sqrt{\Delta}+ \frac{\sigma}{\sqrt{L_2\alpha B}}\Big ]+L_1\Big [\frac{L_2 (1-\frac{2}{L_2+\alpha} \alpha)^{\tau}}{\alpha}+\frac{1}{\alpha\sqrt{B}}(\frac{L_2^2}{\alpha}+L_2)\\
        &+\frac{\sigma}{\alpha\sqrt{L_2\alpha B}}(\frac{L_2L_4}{\alpha}+L_3)+\frac{2}{L_2+\alpha} (\frac{L_2L_4}{\alpha}+L_3)\sqrt{\Delta}\\
        &\frac{(1-\frac{2}{L_2+\alpha}\alpha)^\tau}{1-\frac{2}{L_2+\alpha}\alpha - \frac{L_2-\alpha}{L_2+\alpha}}\Big ]+\frac{L_1}{\sqrt{B}}.
    \end{aligned}\label{prop1}
\end{equation}
\end{small}

\noindent \textbf{Proof.}  Using the triangle inequality, we have:
\begin{small}
\begin{equation}
\begin{aligned}
    &\mathbb{E}\|\frac{\partial l_{{\rm global}} (\Theta_t^0,\mathcal{E}_t^\tau;\mathcal{B}_0)}{\partial \Theta_t^0} - \nabla F(\Theta_t^0)\|\\
    &\overset{(i)}{\leq} \frac{L_1}{\sqrt{B}}+\mathbb{E}\|\frac{\partial \mathcal{L}_{{\rm global}} (\Theta_t^0,\mathcal{E}_t^\tau)}{\partial \Theta_t^0} - \nabla F(\Theta_t^0)\|.
\end{aligned}\label{prop1_bound}
\end{equation}
\end{small}
where (i) follows from Eq. \ref{cls_bound}.

Then, we need to give an upper bound for $\mathbb{E}\|\frac{\partial \mathcal{L}_{{\rm global}} (\Theta_t^0,\mathcal{E}_t^\tau)}{\partial \Theta_t^0} - \nabla F(\Theta_t^0)\|$. Using:
\begin{small}
\begin{equation}
\begin{aligned}
    \nabla F(\Theta_t^0)&=\nabla \mathcal{L}_{{\rm global}}(\Theta_t^0,\mathcal{E}^*(\Theta_t^0)) \\
    &+ \frac{\partial \mathcal{E}^*(\Theta)_t^0}{\partial \Theta_t^0}\nabla_{\mathcal{E}}\mathcal{L}_{{\rm global}}(\Theta_t^0,\mathcal{E}^*(\Theta_t^0)),
\end{aligned}
\end{equation}
\end{small}
and
\begin{footnotesize}
\begin{equation}
\begin{aligned}
    \frac{\partial \mathcal{L}_{{\rm global}}(\Theta_t^0,\mathcal{E}_t^\tau)}{\partial \Theta_t^0}&=\nabla_{\Theta}\mathcal{L}_{{\rm global}}(\Theta_t^0,\mathcal{E}_t^\tau)+\frac{\partial \mathcal{E}_t^\tau}{\partial \Theta_t^0}\nabla_{\mathcal{E}}\mathcal{L}_{{\rm global}}(\Theta_t^0,\mathcal{E}_t^\tau),
\end{aligned}
\end{equation}
\end{footnotesize}
we have: 
\begin{small}
\begin{equation}
    \begin{aligned}
    &\mathbb{E}\|\frac{\partial \mathcal{L}_{{\rm global}} (\Theta_t^0,\mathcal{E}_t^\tau)}{\partial \Theta_t^0} - \nabla F(\Theta_t^0)\|\leq L_2 \mathbb{E}\|\mathcal{E}_t^\tau - \mathcal{E}^*(\Theta_t^0)\| \\
    &+ L_1 \mathbb{E}\|\frac{\partial \mathcal{E}_t^\tau}{\partial \Theta_t^0} - \frac{\partial \mathcal{E}^*(\Theta_t^0)}{\partial \Theta_t^0}\|+L_2\mathbb{E}(\|\frac{\partial \mathcal{E}^*(\Theta)_t^0}{\partial \Theta_t^0}\| \|\mathcal{E}_t^\tau - \mathcal{E}^*(\Theta_t^0)\|).
    \end{aligned} \label{approximate_bound}
\end{equation}
\end{small}

Now we first want to bound $\mathbb{E}\|\frac{\partial \mathcal{E}_t^\tau}{\partial \Theta_t^0} - \frac{\partial \mathcal{E}^*(\Theta_t^0)}{\partial \Theta_t^0}\|$. Recall the update method:
\begin{equation}
    \mathcal{E}_t^j=\mathcal{E}_t^{j-1} - \eta \nabla_{\mathcal{E}}l_{{\rm local}}(\Theta_t^0,\mathcal{E}_t^{j-1};\mathcal{B}_{j-1}),
\end{equation}
and use the chain rule on it, we have:
\begin{footnotesize}
\begin{equation}
    \begin{aligned}
        \frac{\partial \mathcal{E}_t^j}{\partial \Theta_t^0}&=\frac{\partial \mathcal{E}_t^{j-1}}{\partial \Theta_t^0}-\eta(\nabla_{\Theta}\nabla_{\mathcal{E}}l_{{\rm local}}(\Theta_t^0,\mathcal{E}_t^{j-1};\mathcal{B}_{j-1})\\
        &+\frac{\partial \mathcal{E}_t^{j-1}}{\partial \Theta_t^0} \nabla^2_{\mathcal{E}}l_{{\rm local}} (\Theta_t^0,\mathcal{E}_t^{j-1};\mathcal{B}_{j-1})).
    \end{aligned}\label{update_grad}
\end{equation}
\end{footnotesize}

For $\mathcal{E}^*(\Theta_t^0)$ is the optimal solution of $\mathcal{L}_{{\rm local}}(\Theta_t^0, \mathcal{E})$, we have $\nabla_{\mathcal{E}}\mathcal{L}_{{\rm local}}(\Theta_t^0, \mathcal{E}^*(\Theta_t^0))=0$. Then, using the chain rule, we have:
\begin{small}
\begin{equation}
\begin{aligned}
    \nabla_{\Theta}\nabla_{\mathcal{E}}\mathcal{L}_{{\rm local}}(\Theta_t^0, \mathcal{E}^*(\Theta_t^0)) + \frac{\partial\mathcal{E}^*(\Theta_t^0)}{\partial \Theta_t^0} \nabla_{\mathcal{E}}^2\mathcal{L}_{{\rm local}}(\Theta_t^0, \mathcal{E}^*(\Theta_t^0))=0. 
\end{aligned}
\label{optimal_condition}
\end{equation}
\end{small}

Combining Eq. \ref{update_grad} and Eq. \ref{optimal_condition}, the following equation holds:
\begin{small}
\begin{equation}
    \begin{aligned}
        &\frac{\partial \mathcal{E}_t^j}{\partial \Theta_t^0} -\frac{\partial \mathcal{E}^* (\Theta_t^0)}{\partial \Theta_t^0}=\frac{\partial \mathcal{E}_t^{j-1}}{\partial \Theta_t^0} -\frac{\partial \mathcal{E}^* (\Theta_t^0)}{\partial \Theta_t^0}-\eta(\nabla_{\Theta}\nabla_{\mathcal{E}}l_{{\rm local}}\\
        &(\Theta_t^0,\mathcal{E}_t^{j-1};\mathcal{B}_{j-1}) -\nabla_{\Theta}\nabla_{\mathcal{E}} \mathcal{L}_{{\rm local}}(\Theta_t^0,\mathcal{E}^*(\Theta_t^0)))\\
        &-\eta \left(\frac{\partial \mathcal{E}_{t}^{j-1}}{\partial \Theta_t^0} - \frac{\partial \mathcal{E}^* (\Theta_t^0)}{\partial \Theta_t^0} \right) \nabla_{\mathcal{E}}^2 l_{{\rm local}}(\Theta_t^0,\mathcal{E}_t^{j-1};\mathcal{B}_{j-1}) \\
        &+\eta \frac{\partial \mathcal{E}^*(\Theta_t^0)}{\partial \Theta_t^0}(\nabla^2_{\mathcal{E}}l_{{\rm local}}(\Theta_t^0, \mathcal{E}_t^{j-1};\mathcal{B}_{j-1})- \nabla^2_{\mathcal{E}}l_{{\rm local}}(\Theta_t^0,\mathcal{E}^*(\Theta_t^0))).
    \label{update_optimal_bound}
    \end{aligned}
\end{equation}
\end{small}

Based on Eq. \ref{optimal_condition}, we have:
\begin{equation}
    \| \frac{\partial \mathcal{E}^*(\Theta_t^0)}{\partial \Theta_t^0}\|\leq \frac{L_2}{\alpha}. \label{E_grad_bound}
\end{equation}

With the help of Assumption 3, Lemma 2, Eq. \ref{update_optimal_bound} and Eq. \ref{E_grad_bound}, the following bound holds:
\begin{small}
\begin{equation}
    \begin{aligned}
        \mathbb{E}&\|\frac{\partial \mathcal{E}_t^j}{\partial \Theta_t^0} - \frac{\partial \mathcal{E}^* (\Theta_t^0)}{\partial \Theta_t^0}\| \leq (1-\eta\alpha)\mathbb{E}\|\frac{\partial \mathcal{E}_t^{j-1}}{\partial \Theta_t^0} - \frac{\partial \mathcal{E}^*(\Theta_t^0)}{\partial \Theta_t^0}\| \\
        &+ \frac{\eta}{\sqrt{B}}(\frac{L_2^2}{\alpha}+L_2) + \eta (\frac{L_2 L_4}{\alpha}+L_3)\mathbb{E}\|\mathcal{E}_t^{j-1}-\mathcal{E}^*(\Theta_t^0)\|. \label{partial_bound}
    \end{aligned}
\end{equation}
\end{small}

For the choice of $\eta=\frac{2}{L_2+\alpha}$, Lemma 4 holds. With the result in Eq. \ref{tracking_error} and Assumption 5, we have:
\begin{small}
\begin{equation}
    \begin{aligned}
    \mathbb{E}&\|\mathcal{E}_t^{j-1} - \mathcal{E}^*(\Theta_t^0)\| \leq \left(\frac{L_2 - \alpha}{L_2 +\alpha}\right)^{j-1}\sqrt{\mathbb{E}\|\mathcal{E}_t^{0} - \mathcal{E}^*(\Theta_t^0)\|} \\
    &+ \frac{\sigma}{\sqrt{L_2\alpha B}} \leq \left(\frac{L_2 - \alpha}{L_2 +\alpha}\right)^{j-1}\sqrt{\Delta} + \frac{\sigma}{\sqrt{L_2\alpha B}}.
    \end{aligned}\label{E_bound}
\end{equation}
\end{small}

Telescoping Eq. \ref{partial_bound} over $j$ from 0 to $\tau$ and combining the result with Eq. \ref{E_bound} yields:
\begin{small}
\begin{equation}
    \begin{aligned}
        \mathbb{E}&\|\frac{\partial \mathcal{E}_t^\tau}{\partial \Theta_t^0} - \frac{\partial \mathcal{E}^*(\Theta_t^0)}{\partial \Theta_t^0}\| \\
        &\leq \frac{L_2 (1-\eta \alpha)^{\tau}}{\alpha}+\frac{1}{\alpha\sqrt{B}}(\frac{L_2^2}{\alpha}+L_2) +\frac{\sigma}{\alpha\sqrt{L_2\alpha B}}(\frac{L_2L_4}{\alpha}+L_3)\\
        &+\eta (\frac{L_2L_4}{\alpha}+L_3)\sqrt{\Delta} \frac{(1-\eta\alpha)^\tau}{1-\eta\alpha - \frac{L_2-\alpha}{L_2+\alpha}}.
    \end{aligned}
    \label{E_tau_bound}
\end{equation}
\end{small}

Take expedition of both sides of Eq. \ref{approximate_bound}, plugging Eq. \ref{E_grad_bound}, Eq. \ref{E_bound} and Eq. \ref{E_tau_bound} into it yields:
\begin{small}
\begin{equation}
    \begin{aligned}
        \mathbb{E}&\|\frac{\partial \mathcal{L}_{{\rm global}} (\Theta_t^0,\mathcal{E}_t^\tau)}{\partial \Theta_t^0} - \nabla F(\Theta_t^0)\| \leq (L_2 + \frac{L_2^2}{\alpha})\Big [\left(\frac{L_2 - \alpha}{L_2 +\alpha}\right)^{\tau}\\
        &\sqrt{\Delta} + \frac{\sigma}{\sqrt{L_2\alpha B}}\Big ] +L_1\Big [\frac{L_2 (1-\frac{2}{L_2+\alpha} \alpha)^{\tau}}{\alpha} +\frac{1}{\alpha\sqrt{B}}(\frac{L_2^2}{\alpha}\\
        &+L_2) + \frac{\sigma}{\alpha\sqrt{L_2\alpha B}}(\frac{L_2L_4}{\alpha}+L_3) +\frac{2}{L_2+\alpha}(\frac{L_2L_4}{\alpha}+L_3)\\
        &\sqrt{\Delta}\frac{(1-\frac{2}{L_2+\alpha}\alpha)^\tau}{1-\frac{2}{L_2+\alpha}\alpha - \frac{L_2-\alpha}{L_2+\alpha}}\Big ].
    \end{aligned}\label{prop1_prof}
\end{equation}
\end{small}

Finally, plugging Eq. \ref{prop1_prof} into Eq. \ref{prop1_bound} yields Eq. \ref{prop1}. Thus, we complete the proof of Proposition 1.

To handle multiple updates of the active party with a fixed $t$, we use a technology called virtual updates \cite{yang2022fastslowmo}. Specifically, we introduce a virtual parameter $\mathcal{E}_{t,j}^\tau$ obtained by local updates when $\Theta_t^{j}$ is given. Using Proposition 1, we can bound $\mathbb{E}\|\frac{\partial l_{{\rm global}} (\Theta_t^{j},\mathcal{E}_{t,j}^\tau;\mathcal{B}_{j})}{\partial \Theta_t^{j}} - \nabla F(\Theta_t^{j})\|$. Now, we aim to bound $\mathbb{E}\|\frac{\partial l_{{\rm global}} (\Theta_t^{j},\mathcal{E}_{t}^\tau;\mathcal{B}_{j})}{\partial \Theta_t^{j}} - \nabla F(\Theta_t^{j})\|$. The result follows:

\noindent\textbf{Proposition 2.} Following the conditions in Proposition 1, we have
\begin{small}
\begin{equation}
    \begin{aligned}
        \mathbb{E}&\|\frac{\partial l_{{\rm global}} (\Theta_t^{j},\mathcal{E}_t^\tau;\mathcal{B}_{j})}{\partial \Theta_t^{j}} - \nabla F(\Theta_t^{j})\|\leq (L_2 + \frac{L_2^2}{\alpha})\\
        &\Big [\left(\frac{L_2 - \alpha}{L_2 +\alpha}\right)^{\tau} \sqrt{\Delta}+ \frac{\sigma}{\sqrt{L_2\alpha B}}\Big ]+L_1\Big [\frac{L_2 (1-\frac{2}{L_2+\alpha} \alpha)^{\tau}}{\alpha}\\
        &+\frac{1}{\alpha\sqrt{B}}(\frac{L_2^2}{\alpha}+L_2) +\frac{\sigma}{\alpha\sqrt{L_2\alpha B}}(\frac{L_2L_4}{\alpha}+L_3) +\frac{2}{L_2+\alpha} \\
        &(\frac{L_2L_4}{\alpha}+L_3)\sqrt{\Delta}\frac{(1-\frac{2}{L_2+\alpha}\alpha)^\tau}{1-\frac{2}{L_2+\alpha}\alpha - \frac{L_2-\alpha}{L_2+\alpha}}\Big ]+\frac{L_1}{\sqrt{B}}+L_2 \Delta.
    \end{aligned}\label{prop2}
\end{equation}
\end{small}

\noindent\textbf{Proof.} Using the triangle inequality, we have:
\begin{small}
\begin{equation}
    \begin{aligned}
        &\|\frac{\partial l_{{\rm global}} (\Theta_t^{j},\mathcal{E}_t^\tau;\mathcal{B}_{j})}{\partial \Theta_t^{j}} - \nabla F(\Theta_t^{j})\| \leq \|\frac{\partial l_{{\rm global}} (\Theta_t^{j},\mathcal{E}_t^\tau;\mathcal{B}_{j})}{\partial \Theta_t^{j}} \\
        &- \frac{\partial l_{{\rm global}} (\Theta_t^{j},\mathcal{E}_{t,j}^\tau;\mathcal{B}_{j})}{\partial \Theta_t^{j}}\| +\| \frac{\partial l_{{\rm global}} (\Theta_t^{j},\mathcal{E}_{t,j}^\tau;\mathcal{B}_{j})}{\partial \Theta_t^{j}} - \nabla F(\Theta_t^{j})\|.
    \end{aligned}\label{virtual_decomp}
\end{equation}
\end{small}

Take the expectation of both sides of Eq. \ref{virtual_decomp}, with the help of Assumptions 2 and 5, the following result holds:
\begin{small}
\begin{equation}
    \begin{aligned}
        &\mathbb{E}\|\frac{\partial l_{{\rm global}} (\Theta_t^{j},\mathcal{E}_t^\tau;\mathcal{B}_{j})}{\partial \Theta_t^{j}} - \nabla F(\Theta_t^{j})\|\\
        &\leq L_2 \Delta + \mathbb{E} \| \frac{\partial l_{{\rm global}} (\Theta_t^{j},\mathcal{E}_{t,j}^\tau;\mathcal{B}_{j})}{\partial \Theta_t^{j}} - \nabla F(\Theta_t^{j})\|.
    \end{aligned}\label{prop2_proof}
\end{equation}
\end{small}

Plugging in the result in Proposition 1 into Eq. \ref{prop2_proof} yields Eq. \ref{prop2}. Thus, we complete the proof of Proposition 2.

\noindent\textbf{Proof for Theorem 1.} Based on the $L_0$-smoothness of $F(\Theta)$ established in Lemma 3, we have:
\begin{small}
\begin{equation}
    \begin{aligned}
        F&(\Theta_t^{j+1}) \leq F((\Theta_t^{j}) - (\frac{\eta^\prime}{2} - \eta^{\prime 2}L_0)\|\nabla F((\Theta_t^{j})\|^2\|\\
        &\frac{\partial l_{{\rm global}}(\Theta_t^{j},\mathcal{E}_t^\tau;B_{j})}{\partial \Theta_t^{j}} - \nabla F(\Theta_t^{j})\|^2.
    \end{aligned}\label{smooth_bound}
\end{equation}
\end{small}

With the help of Proposition 2, telescoping Eq. \ref{smooth_bound} over $j$ from 0 to $\tau-1$ and taking expectation of both sides yields:
\begin{small}
\begin{equation}
    \begin{aligned}
        \mathbb{E}&F(\Theta_t^{\tau})\leq \mathbb{E}F(\Theta_t^{0}) -(\frac{\eta^\prime}{2} - \eta^{\prime 2}L_0) \mathbb{E}\sum_{j=0}^{\tau -1}\|\nabla F(\Theta_t^{j})\|^2\\
        &+(\eta^\prime + 2\eta^{\prime 2}L_0)\tau\{(L_2 + \frac{L_2^2}{\alpha})\Big [\left(\frac{L_2 - \alpha}{L_2 +\alpha}\right)^{\tau}\sqrt{\Delta}+ \frac{\sigma}{\sqrt{L_2\alpha B}}\Big ]\\
        &+L_1\Big [\frac{L_2 (1-\frac{2}{L_2+\alpha} \alpha)^{\tau}}{\alpha}+\frac{1}{\alpha\sqrt{B}}(\frac{L_2^2}{\alpha}+L_2)+\frac{\sigma}{\alpha\sqrt{L_2\alpha B}}(\frac{L_2L_4}{\alpha}\\
        &+L_3)+\frac{2}{L_2+\alpha} (\frac{L_2L_4}{\alpha}+L_3)\sqrt{\Delta}\frac{(1-\frac{2}{L_2+\alpha}\alpha)^\tau}{1-\frac{2}{L_2+\alpha}\alpha - \frac{L_2-\alpha}{L_2+\alpha}}\Big ]\\
        &+\frac{L_1}{\sqrt{B}}\}^2+ (\eta^\prime + 2\eta^{\prime 2}L_0)L_2^2\Delta^2 (\tau-1).
    \end{aligned}\label{theor1_bound}
\end{equation}
\end{small}

Telescoping Eq. \ref{theor1_bound} over $t$ from 0 to $T-1$ yields the following result:
\begin{small}
\begin{equation}
    \begin{aligned}
        \frac{1}{\tau T}&\sum_{t=0}^{T-1}\sum_{j=0}^{\tau-1}\|\nabla F(\Theta_t^{j})\|^2 \leq \frac{F(\Theta_0^0) - \inf_{\Theta}F(\Theta)}{\tau T (\frac{\eta^{\prime}}{2} -L_0 \eta^{\prime 2}) }+(\eta^\prime + 2\eta^{\prime 2}L_0)\\
        &L_2^2\Delta^2 \frac{\tau -1}{\tau}+(\eta^\prime + 2\eta^{\prime 2}L_0)\tau\{(L_2 + \frac{L_2^2}{\alpha})\Big [\left(\frac{L_2 - \alpha}{L_2 +\alpha}\right)^{\tau} \\
        &\sqrt{\Delta}+ \frac{\sigma}{\sqrt{L_2\alpha B}}]+L_1[\frac{L_2 (1-\frac{2}{L_2+\alpha} \alpha)^{\tau}}{\alpha}+\frac{1}{\alpha\sqrt{B}}(\frac{L_2^2}{\alpha}+L_2)\\
        &+\frac{\sigma}{\alpha\sqrt{L_2\alpha B}}(\frac{L_2L_4}{\alpha}+L_3)+\frac{2}{L_2+\alpha} (\frac{L_2L_4}{\alpha}+L_3)\sqrt{\Delta}\\
        &\frac{(1-\frac{2}{L_2+\alpha}\alpha)^\tau}{1-\frac{2}{L_2+\alpha}\alpha - \frac{L_2-\alpha}{L_2+\alpha}}\Big ]+\frac{L_1}{\sqrt{B}}\}^2.
    \end{aligned}\label{general_theorem_1}
\end{equation}
\end{small}

This is a more general result of Theorem 1. Choose $B=\mathcal{O}(\frac{1}{\sqrt{\epsilon}})$ and $\tau=\mathcal{O}(\log\frac{1}{\epsilon})$, we complete proofs of Theorem 1. The full proof of Theorem 1 is shown in Appendix \ref{sec:convergence analysis}. 

\subsection{Privacy Analysis}
Proto-EVFL assumes the active party is trustworthy, allowing collaborative training with shared intermediate representations under GDPR \cite{gdpr2018general}. However, these representations can be vulnerable to privacy attacks, similar to reconstruction attacks in HFL \cite{geiping2020inverting}. This suggests that simply limiting data sharing is not enough to ensure privacy security. In our proposed Proto-EVFL, intermediate representations are in matrix form and not susceptible to the same analysis as gradients. Additionally, the absence of gradient transmission renders feature inference attacks ineffective. We also can further enhance privacy by treating these representations as released data and adding noise. As discussed in Section \ref{sec:privacy}, this method is a potential solution, but the noise should be strictly limited to a small scale to avoid significant performance loss.

Furthermore, the transfer of class prototypes could also leak labels. Existing label inference attacks in VFL \cite{fu2022label} usually depend on gradients, which are not used in Proto-EVFL. However, labels can still be leaked by calculating similarity distances between unaligned samples and class prototypes. In Section \ref{sec:privacy}, we introduce varying degrees of noise to the prototypes, mitigating label leakage risk. Experiments show that our method performs competitively within the VFL framework when noise is kept within a certain range.

\begin{table*}[t]
\centering
\caption{Detailed experimental settings.}
\setlength{\tabcolsep}{1.4mm}
{
\begin{tabular}{cccccccccccccc} \hline
\bottomrule
\multirow{2}{*}{} & \multirow{2}{*}{Size} & \multirow{2}{*}{Class Num.} & \multirow{2}{*}{Test Ratio} & \multirow{2}{*}{Aligned Ratio} & \multicolumn{2}{c}{Epoch} & \multicolumn{2}{c}{Architecture} & \multicolumn{2}{c}{Learning Rate} &\multirow{2}{*}{Optimizer} &\multirow{2}{*}{$\varphi$}&\multirow{2}{*}{$\rho$}\\ \cline{6-11}
                  &                    &           &                       &                                & Passive          & Active         & Passive         & Active         & Passive             & Active       \\ \hline
ModelNet-10                     & 11522             &   10                       &  0.200                              &  0.017                &     9     &  9    &          LeNet-5       &      MLP-3          &           1e-5           &   5e-4   &SGD & 1e-1 & 1e-1  \\
Fashion-MNIST           &   31432                   &     10                      &    0.075                            &      0.006            &     8     & 8 &           LeNet-5      &        MLP-3        &          1e-5            &      5e-4      &SGD& 1e-1 & 1e-1     \\
Credit                  &      30000         &     2                      &    0.200                            &      0.007         &       9    &   9  &           MLP-5      &        MLP-3        &          1e-5            &      3e-3      &SGD& 1e-1 & 1e-1  \\
Adult                     &    48842        &     2                      &    0.024                            &      0.004            &       6  &   6    &           MLP-5      &        MLP-3        &          1e-5            &      3e-3      &SGD& 1e-1   &  1e-1   \\
\hline
\bottomrule
\end{tabular}
}
\label{tab:detailed setting}
\end{table*}

\begin{table*}[t]
\centering
\caption{Class Imbalance Settings.}
\setlength{\tabcolsep}{2.6mm}
{
\begin{tabular}{ccccccccccccccccc} \hline
\bottomrule
\multirow{2}{*}{Party} & \multicolumn{4}{c}{ModelNet-10} & \multicolumn{4}{c}{FMNIST} & \multicolumn{4}{c}{Credit}      & \multicolumn{4}{c}{Adult}      \\ \cline{2-17}
                       & $\Gamma$   & S  & MID  & WCS  & $\Gamma$ & S & MID  & WCS  & $\Gamma$ & S & MID  & WCS  & $\Gamma$ & S & MID  & WCS  \\ \hline
1                      &  12:1          & 10   & 0.228      & 0.862      &   12:1   &10    &  0.232 &   0.842   &  4.7:1    &   2       & 0.331  &  0.878    &   9.5:1  &     2     & 0.540  &  1.0        \\
2                      &  24:1          & 10   &    0.146   &   0.862    &  11.5:1 &10       & 0.232  &  0.841    &  8.2:1    &    2      & 0.502  &  0.838   & 1.1:1     &     2     &  0.201 &  1.0        \\
3                      & 6:1           &  10  &   0.086    &   0.862    &    11:1 &10     & 0.232  & 0.836     &   5.2:1   & 2&    0.364     & 0.879  &   8.5:1           &2  &  0.513    & 1.0    \\
4                      & 5:1           &  10  &   0.138    &   0.842    &   10:1   &10    & 0.231  &   0.805   &   3.5:1   &2&  0.233        & 0.802     &  2.3:1    &      2    & 0.318  &   1.0      \\
\hline
\bottomrule
\end{tabular}
}
\label{tab:class imbalanced setting}
\begin{tablenotes}
        \footnotesize
        \item $\ast$ ${\rm MID} = 0, {\rm WCS} = 1$: the global data is strictly class-balanced; all of the local label distribution vectors align in the same direction. \\ 
        \quad${\rm MID} > 0, {\rm WCS} = 1$: the global data appears to be class-imbalanced; all of the local label distribution vectors are oriented in the same direction. \\
        \quad${\rm MID} = 0, {\rm WCS} < 1$: the global data is strictly class-balanced; the local label distribution vectors show discrepancies in direction. \\
        \quad${\rm MID} > 0, {\rm WCS} < 1$: the global data appears to be class-imbalanced; the local label distribution vectors exhibit discrepancies in direction.
      \end{tablenotes}
\end{table*}

\begin{table}[t]
\centering
\caption{\begin{tabular}[c]{@{}c@{}}Comparison of baseline frameworks. \\Assume that unaligned samples are also unlabeled.\end{tabular}}
\setlength{\tabcolsep}{2.0mm}
\scalebox{0.94}
{
\begin{tabular}{ccccc} \hline
\bottomrule
\multirow{2}{*}{Framework} & \multicolumn{2}{c}{Used Samples} & \multirow{2}{*}{Method} &\multirow{2}{*}{\begin{tabular}[c]{@{}c@{}}Imbalance\\Consideration \end{tabular}}  \\
 & Aligned & Unaligned & \\ \hline
Local Model  & $\checkmark$ &  $\times$ & Traditional & $\times$                      \\
Vanilla VFL$^{\ast}$               &  $\checkmark$   & $\times$                    & Traditional    & $\times$                         \\
SS-VFL$^{\ast}$                    & $\checkmark$                  &  $\checkmark$                  & Contrastive    & $\times$                     \\
FedHSSL$^{\ast}$                   & $\checkmark$                  & $\checkmark$                   & Contrastive      & $\times$                  \\
\textbf{Proto-EVFL}$^{\ast}$                &  $\checkmark$                 & $\checkmark$          & Probabilistic   & $\checkmark$                    \\
Upper Boundary$^{\ast}$                      &  $\checkmark$              & $\checkmark$                  & Traditional                 & $\times$                      \\ \hline
\bottomrule
\end{tabular}}

\label{tab:baselines}
\begin{threeparttable}
 \begin{tablenotes}
        \footnotesize
        \item $\ast$ denotes that this method is a VFL model.
      \end{tablenotes}

\end{threeparttable}
\end{table}

\begin{table*}[t]
\centering
\caption{Test accuracy ($\%$) of different methods on image datasets (ModelNet-10 and Fashion-MNIST).}
\setlength{\tabcolsep}{1.4mm}
{
\begin{tabular}{ccccccccccc} \hline
\bottomrule
\multirow{2}{*}{Dataset}  & \begin{tabular}[c]{@{}c@{}}\#Aligned data\\ (=labeled data)\end{tabular}& \multicolumn{3}{c}{\#200}  & \multicolumn{3}{c}{\#600}  & \multicolumn{3}{c}{\#1000}                                                     \\ \cline{2-11}
                          & Class imbalance      & zero-shot   & few-shot                       & normal                         & zero-shot            & few-shot             & normal                         & zero-shot            & few-shot             & normal                         \\ \hline
\multirow{7}{*}{ModelNet-10} & Local Model                  &  62.90±0.45                               & 63.70±0.51                            & 66.38±0.72
                               & 68.80±0.63                    & 69.71±0.49                      & 72.41±0.89                               & 70.12±0.88                     & 73.41±0.65                     & 75.31±0.74                               \\
                          & Vanilla VFL                  & 67.58±0.62                                & 68.14±0.52                               & 71.03±0.64                               & 73.33±0.48                     & 72.21±0.79                    & 75.12±0.55                                & 74.00±0.57                     & 74.78±0.74                     & 78.31±0.68                              \\
                          & SS-VFL                       & -                                & -                               & 69.74±0.54                               &  -                    & -                    &   74.12±0.72                             & -                     & -                     &  78.91±0.61                              \\
                          & FedHSSL                      & 64.93±1.01                      & 32.63±1.00                              & 70.71±0.90                     & 53.17±0.95           & 57.10±0.95                   & 80.60±0.80                  & 61.90±0.70          & 63.91±0.70                     & 83.32±0.60                     \\ \cline{2-11} 
                          & \cellcolor{gray!30}\textbf{Proto-EVFL}                   & \cellcolor{gray!30}\textbf{76.72±0.54}                     &\cellcolor{gray!30}\textbf{77.52±0.80}                   & \cellcolor{gray!30}\textbf{81.64±0.80}                    & \cellcolor{gray!30}\textbf{79.33±0.84}          & \cellcolor{gray!30}\textbf{79.65±0.62}           & \cellcolor{gray!30}\textbf{83.81±0.62}                     & \cellcolor{gray!30}\textbf{79.01±0.95}           & \cellcolor{gray!30}\textbf{80.95±0.40}           & \cellcolor{gray!30}\textbf{84.19±0.68}                     \\ \cline{2-11} 
                          & Upper Boundary               & \multicolumn{9}{c}{84.93±0.70}  \\ \hline
\multirow{6}{*}{\begin{tabular}[c]{@{}c@{}}Fashion-\\ MNIST\end{tabular}}   & Local Model                  & 50.79±0.74                                & 52.32±0.81                               & 58.97±0.66                               & 55.89±0.59                     & 56.79±0.78                     & 59.22±0.85                               & 56.35±0.76                     & 57.22±0.74                     & 60.56±0.84                               \\
                          & Vanilla VFL                  & 55.60±0.59                                & 58.99±0.67                               & 55.60±0.81                               & 58.90±0.74                     & 60.16±0.69                     & 63.90±0.88                               & 57.22±0.83                     & 57.42±0.65                     & 63.18±0.79                               \\
                          & SS-VFL                       & -                                & -                               & 52.54±0.72                               & -                     & -                     & 65.54±0.81                                & -                     & -                     & 66.97±0.66                              \\
                          & FedHSSL                      & 40.31±0.20                             & 46.33±0.20                               & 50.95±0.30                     & 64.61±0.30                     & 64.94±0.40                     & 69.02±0.05                     & 65.91±0.50                     & 67.82±0.20                     & 70.85±0.58 \\ \cline{2-11}
                          & \cellcolor{gray!30}\textbf{Proto-EVFL}                   & \multicolumn{1}{l}{\cellcolor{gray!30}\textbf{65.28±0.79}} & \multicolumn{1}{l}{\cellcolor{gray!30}\textbf{65.58±0.62}} & \multicolumn{1}{l}{\cellcolor{gray!30}\textbf{66.48±0.50}} & \multicolumn{1}{l}{\cellcolor{gray!30}\textbf{65.16±0.50}} & \multicolumn{1}{l}{\cellcolor{gray!30}\textbf{67.14±1.20}} & \multicolumn{1}{l}{\cellcolor{gray!30}\textbf{70.47±0.32}} & \multicolumn{1}{l}{\cellcolor{gray!30}\textbf{67.05±0.30}} & \multicolumn{1}{l}{\cellcolor{gray!30}\textbf{69.19±1.20}} & \multicolumn{1}{l}{\cellcolor{gray!30}\textbf{72.05±0.35}} \\ \cline{2-11} 
                          & Upper Boundary               & \multicolumn{9}{c}{73.17±0.42}    \\  \hline
\bottomrule      
\end{tabular}
}
\begin{tablenotes}
        \footnotesize
        \item $\ast$ The few-shot and zero-shot scenarios were designed by reducing the sample size of Class 3 in both datasets. For ModelNet-10, Class 3 corresponds to `desk', while for Fashion-MNIST, it represents `dress'.
      \end{tablenotes}
    
\label{tab:acc}
\end{table*}

\begin{table}[t]
\centering
\caption{Test accuracy ($\%$) of different methods on tabular datasets (Credit and Adult).}
\setlength{\tabcolsep}{2.4mm}
{
\begin{tabular}{ccccccc}\hline
\bottomrule
\multirow{3}{*}{Method} & \multicolumn{6}{c}{Dataset}                                 \\ \cline{2-7}
                        & \multicolumn{3}{c}{Credit}   & \multicolumn{3}{c}{Adult}    \\ \cline{2-7}
                        & \#200 & \#600 & \#1000 & \#200 & \#600 & \#1000 \\ \hline
FedHSSL                    & 60.41   & 75.70   & 78.15    & 74.80   & 75.43   & 76.92    \\
\cellcolor{gray!30}\textbf{Proto-EVFL}              & \cellcolor{gray!30}\textbf{77.96}   &\cellcolor{gray!30}\textbf{81.07}   & \cellcolor{gray!30}\textbf{82.42}    & \cellcolor{gray!30}\textbf{75.81}   &\cellcolor{gray!30}\textbf{77.90}   & \cellcolor{gray!30}\textbf{79.22}  \\  \hline
\bottomrule      
\end{tabular}
}
\label{tab:ACC on tabular data}
\end{table}

\begin{table}[]
\centering
\caption{Ablation study results}
\setlength{\tabcolsep}{3.3mm}{
\begin{tabular}{lcc} \hline
\bottomrule
\multicolumn{1}{c}{\multirow{2}{*}{Ablation setting}} & \multicolumn{2}{c}{Accuracy ($\%$)}       \\ \cline{2-3}
\multicolumn{1}{c}{}                                  & \multicolumn{1}{c}{\#200} & \#1000                 \\ \hline
Vanilla VFL               &        71.03   &                      78.31\\
\cellcolor{gray!30}\textbf{Proto-EVFL}                 &   \cellcolor{gray!30}\textbf{81.64}        &                       \cellcolor{gray!30}\textbf{84.19}\\ 
w/o prototype updating process&         79.42& 83.41 \\
w/o prototype learning scheme&         76.23& 82.71  \\
-----w/o mixed prior guided module &   80.17& 83.23 \\
w/o adaptive gated feature aggregation strategy &       78.12& 81.47  \\ \hline
\bottomrule
\end{tabular}
}
\label{tab:ablation}
\end{table}

\begin{table}[t]
\centering
\caption{Selection method comparisons for unaligned data}
\setlength{\tabcolsep}{4.4mm}{
\begin{tabular}{lccc} \hline
\bottomrule
Strategy                      & \#200 & \#600 & \#1000 \\\hline
Cosine similarity                        & 54.35    & 71.77    & 75.79     \\
Traditional optimal transport & 52.80  & 68.53    & 74.89     \\
PTDC (single direction)       & 48.17    & 65.41    & 73.28     \\
\cellcolor{gray!30}\textbf{PTDC}                          & \cellcolor{gray!30}\textbf{81.64}   & \cellcolor{gray!30}\textbf{83.81}   & \cellcolor{gray!30}\textbf{84.19}  \\ \hline
\bottomrule
\end{tabular}
}
\label{tab: OT ablation}
\end{table}

\section{Experiments}
\subsection{Experimental Settings}
In this section, we demonstrate the following settings: baselines, datasets, network architecture, hyper-parameter, and class imbalance, as shown in Tab. \ref{tab:detailed setting}, \ref{tab:class imbalanced setting} and \ref{tab:baselines}, respectively. For more details, please check our Appendix \ref{sec:exp details}. 

\textbf{Baselines.} We select five related and popular baselines, namely Local Model, Vanilla VFL \cite{sun2023communication}, SS-VFL \cite{castiglia2022self}, FedHSSL \cite{he2022hybrid} and Upper Boundary model. Among them, the Local Model only utilizes local all samples of the active party to train the classifier. Vanilla VFL applies aggregated intermediate features of aligned samples to complete training. SS-VFL and FedHSSL leverage local unaligned and unlabeled samples to augment available training sets. Upper Boundary model trains the target classifier of the active party with a full training set and assumes that all features of unaligned samples are known without FL. For fairness, all baselines use the same backbone network architectures and settings. 

\textbf{Datasets.} To evaluate the effectiveness of our proposed Proto-EVFL, we conduct experiments on two widely used image datasets including ModelNet-10 \cite{wu20153d} and Fashion-MNIST \cite{xiao2017fashion}) datasets in VFL. Meanwhile, two popular tabular datasets, including Credit \cite{misc_default_of_credit_card_clients_350} and Adult \cite{misc_adult_2}) datasets are used to evaluate the generality of Proto-EVFL. 

\textbf{Network architecture.}
For fair comparisons, we choose consistent network architectures and hyper-parameter settings for all methods in the same dataset. LeNet-5 is adopted as the local feature extractor, which comprises two convolutional layers, two pooling layers, and two linear layers. On the activate party side, we select a multi-layer perceptron (MLP) including three fully connected layers to serve as the classifier. 

\textbf{Hyper-parameters.}
All models in our experiments use the stochastic gradient descent (SGD) optimizer \cite{kinga2015method}. The learning rate for the extractor is set to 1e-5, and for the active-side classifier, it is 5e-4 for image datasets and 3e-3 for tabular datasets. The batch size is 64, and the default number of parties is 4. $\varphi$ is set to 1e-1 to implement regularization. Local prototypes are updated with $\rho$, defaulting to 1e-1.

\begin{table*}[]
\centering
\caption{Test accuracy (\%) of different models under zero-shot and few-shot settings on ModelNet-10 dataset.}
\setlength{\tabcolsep}{1.5mm}{
\begin{tabular}{ccccccccccc} \hline
\bottomrule
                         & \multicolumn{2}{c}{Class 1}                                     & \multicolumn{2}{c}{Class 2}       & \multicolumn{2}{c}{Class 3}       & \multicolumn{2}{c}{Class 4}       & \multicolumn{2}{c}{Class 5}                \\ \cline{2-11}
\multirow{-2}{*}{Method}       &Zero                            & Few                           & Zero            & Few             & Zero            & Few             & Zero            & Few             & Zero     & Few                             \\ \hline
FedHSSL                     & 62.41±0.82                        & 64.48±0.53                      & 67.12±0.51       & 69.23±0.38       & 64.40±0.55        & 66.42±0.57        & 65.74±0.33        & 68.15±0.25        & 63.26±0.18   & 67.22±0.82                        \\
\cellcolor{gray!30}\textbf{Proto-EVFL}                     &\cellcolor{gray!30}\textbf{73.56±0.61}                       &\cellcolor{gray!30}\textbf{77.22±0.37}                       &\cellcolor{gray!30}\textbf{74.09±2.13}        & \cellcolor{gray!30}\textbf{76.57±0.87}       &\cellcolor{gray!30}\textbf{79.51±0.56}        &\cellcolor{gray!30}\textbf{79.90±0.42}        &\cellcolor{gray!30}\textbf{78.83±0.85}        &\cellcolor{gray!30}\textbf{79.17±0.35}        &\cellcolor{gray!30}\textbf{74.03±1.02}  &\cellcolor{gray!30}\textbf{75.00±0.63}                       \\ \hline
                         & \multicolumn{2}{c}{Class 1,2}                                   & \multicolumn{2}{c}{Class 3,4}     & \multicolumn{2}{c}{Class 4,5}     & \multicolumn{2}{c}{Class 1,3}     & \multicolumn{2}{c}{Class 3,5}              \\ \cline{2-11}
\multirow{-2}{*}{Method} & Zero                            & Few                           & Zero            & Few             & Zero            & Few             & Zero            & Few             & Zero     & Few                             \\ \hline
FedHSSL                     & 58.92±0.52                        & 58.63±0.56                      & 64.85±0.40        & 65.25±0.79        & 61.73±0.81        & 60.93±1.11        & 58.78±0.46        & 59.42±0.31        & 61.10±0.36 & 61.44±0.50                        \\
\cellcolor{gray!30}\textbf{Proto-EVFL}                    & \cellcolor{gray!30}\textbf{66.09±0.54 }&\cellcolor{gray!30}\textbf{73.42±0.21} &\cellcolor{gray!30}\textbf{68.77±0.97}        &\cellcolor{gray!30}\textbf{75.13±0.63}        &\cellcolor{gray!30}\textbf{71.09±0.44}        &\cellcolor{gray!30}\textbf{71.54±0.82}        &\cellcolor{gray!30}\textbf{70.26±0.57}        &\cellcolor{gray!30}\textbf{74.72±0.23}        &\cellcolor{gray!30}\textbf{65.36±0.64} &\cellcolor{gray!30}\textbf{72.31±0.31} \\ \hline
                         & \multicolumn{2}{c}{Class 1,2,3}                                 & \multicolumn{2}{c}{Class 2,3,4}   & \multicolumn{2}{c}{Class 3,4,5}   & \multicolumn{2}{c}{Class 1,3,4}   & \multicolumn{2}{c}{Class 2,3,5}            \\ \cline{2-11}
\multirow{-2}{*}{Method} & Zero                            & Few                           & Zero            & Few             & Zero            & Few             & Zero            & Few             & Zero     & Few                             \\ \hline
FedHSSL                     & 55.34±0.32                        & 56.32±0.75                      & 61.32±0.33        & 61.17±0.31        & 58.83±0.56        & 59.58±0.39        & 57.41±0.52        & 58.03±1.00        & 57.36±0.61 & 57.73±1.06                        \\
\cellcolor{gray!30}\textbf{Proto-EVFL}                    &\cellcolor{gray!30}\textbf{60.02±0.55}                        &\cellcolor{gray!30}\textbf{71.44±0.20}                      &\cellcolor{gray!30}\textbf{65.79±0.34}        &\cellcolor{gray!30}\textbf{70.47±1.25}        &\cellcolor{gray!30}\textbf{65.22±0.37}        &\cellcolor{gray!30}\textbf{68.55±1.41}        &\cellcolor{gray!30}\textbf{65.26±0.50}        &\cellcolor{gray!30}\textbf{72.89±0.62}        &\cellcolor{gray!30}\textbf{60.37±0.71} &\cellcolor{gray!30}\textbf{67.31±0.96}                        \\ \hline
                         & \multicolumn{2}{c}{Class 1,2,3,4}                               & \multicolumn{2}{c}{Class 2,3,4,5} & \multicolumn{2}{c}{Class 1,3,4,5} & \multicolumn{2}{c}{Class 1,2,4,5} & \multicolumn{2}{c}{Class 1,2,3,5}          \\ \cline{2-11}
\multirow{-2}{*}{Method} & Zero                            & Few                           & Zero            & Few             & Zero            & Few             & Zero            & Few             & Zero     & Few                             \\ \hline
FedHSSL                     & 53.82±0.45                        & 54.32±0.33                      & 55.27±0.61        & 55.19±0.84        & 52.25±0.38        & 53.20±0.37        & 51.43±0.30        & 52.17±0.36        & 50.63±1.35 & 50.71±0.82                        \\ \cellcolor{gray!30}\textbf{Proto-EVFL}                     &\cellcolor{gray!30}\textbf{55.30±0.53}                        &\cellcolor{gray!30}\textbf{70.51±0.54}                      &\cellcolor{gray!30}\textbf{55.77±0.23}       &\cellcolor{gray!30}\textbf{64.74±0.79}        &\cellcolor{gray!30}\textbf{54.18±0.27}        &\cellcolor{gray!30}\textbf{67.71±0.90}        &\cellcolor{gray!30}\textbf{51.56±0.42}        &\cellcolor{gray!30}\textbf{67.64±0.66}        &\cellcolor{gray!30}\textbf{55.32±0.45} &\cellcolor{gray!30}\textbf{67.84±0.71}   \\ \hline
\bottomrule
\label{tab:multi-class}
\end{tabular}
}
\begin{tablenotes}
        \footnotesize
        \item $\ast$ In this table, few-shot and zero-shot experiments were conducted on categories 1 through 5 of the ModelNet-10 dataset. These categories correspond to the following classes: 1 (bed), 2 (chair), 3 (desk), 4 (dresser), and 5 (monitor).
      \end{tablenotes}
\end{table*}

\textbf{Class-imbalanced settings.} To assess the effectiveness of Proto-EVFL in class-imbalanced VFL scenarios, we design intra-class imbalance and inter-party class imbalance, as shown in Tab. \ref{tab:class imbalanced setting}. Specifically, we adopt four metrics to indicate the degree of class imbalance, including $\Gamma$, $S$, multi-class imbalance degree (MID), and weighted cosine similarity (WCS). $\Gamma$ indicates the ratio between the numbers of the majority and minority classes. $S$ denotes the number of classes. MID and WCS are applied to measure the global class-imbalanced degree and intra-party (local) class-imbalanced degree, respectively, as previous work \cite{xiao2021experimental}, where global class imbalance indicates that the data from all parties exhibit an overall class imbalance. We achieve inter-party class imbalance by having each party randomly select the same classes from all available classes as their local most prevalent classes.

\subsection{Overall Performance}
Tab. \ref{tab:acc} and Tab. \ref{tab:ACC on tabular data} compare the performance of Proto-EVFL with existing approaches on two benchmark image datasets and two tabular datasets. Tab. \ref{tab:acc} shows that Proto-EVFL outperforms other VFL frameworks, particularly in extremely class-imbalanced scenarios, i.e., certain classes are rare or invisible (few-shot and zero-shot settings). Proto-EVFL achieves at least 11.79\% improvements on the ModelNet-10 dataset, demonstrating its ability to improve model generalization through probabilistic dual prototype learning. The results on Fashion-MNIST further confirm Proto-EVFL's effectiveness in handling class-imbalanced challenges on image datasets. Moreover, Proto-EVFL shows consistent performance improvements across different levels of aligned data, indicating its scalability and ability to leverage limited aligned data. The Upper Boundary results suggest potential future improvements in exploiting unaligned and unlabeled samples.

Tab. \ref{tab:ACC on tabular data} compares the performance of Proto-EVFL and the FedHSSL (SOTA model) on the Credit and Adult tabular datasets across various aligned sample sizes. Since VFL tabular datasets are primarily binary classification, we focus on their performance under regular class-imbalanced settings. Proto-EVFL consistently outperforms FedHSSL. On the Credit dataset, it achieves 77.96\% accuracy on the smallest subset (\#200), surpassing FedHSSL's 60.41\%, and maintains superior performance on larger subsets (\#600 and \#1000). Similarly, on the Adult dataset, Proto-EVFL achieves 79.22\% accuracy on the largest subset (\#1000), compared to FedHSSL's 76.92\%. These results highlight Proto-EVFL's effectiveness with tabular data and limited aligned samples.

\subsection{Ablation Study}
Tab. \ref{tab:ablation} shows the performance of model ablations with our best configuration. The results indicate that removing key components, such as the local prototype update (Eq. \ref{equ:prototype update}), the PDTC calculation (Section \ref{sec:dual prototype}), mixed prior guided module (Eq. \ref{equ:mixed prior}), or the adaptive gated feature aggregation strategy (Eq. \ref{equ:weight aggregation}), leads to consistent performance drops averaging 1.50\%, 3.45\%, 1.22\%, and 3.03\% on the \#200 and \#1000 aligned data settings, respectively. The performance degradation becomes more pronounced as the number of aligned samples decreases, highlighting the robustness of Proto-EVFL with limited labeled data. Overall, these results underscore the importance of each component in maintaining the accuracy and stability of our model and demonstrate the effectiveness of all key components in our proposed Proto-EVFL.

\subsection{Selection Strategies Comparisons of Unaligned Data}
To compare the model performance under different unaligned data selection strategies, specifically including cosine similarity \cite{rahutomo2012semantic}, traditional OT distance \cite{frogner2015learning}, PTDC with single-direction, and PTDC, we conduct comparison experiments on the ModelNet-10 dataset. Tab. \ref{tab: OT ablation} demonstrates PTDC's superiority in selecting class-relevant unaligned samples across all sample sizes (\#200 to \#1000). In contrast, the other methods show lower performance, with accuracies ranging from 48.17\% to 75.79\%. These results highlight that incorporating the conditional probability-based dual optimal transport distance, which considers local class distribution, into the definition of model training loss can effectively mitigate local model bias. Consequently, PTDC demonstrates its superiority in leveraging unaligned data effectively in VFL. 

\begin{figure}[h]
  \centerline{\includegraphics[scale=0.3]{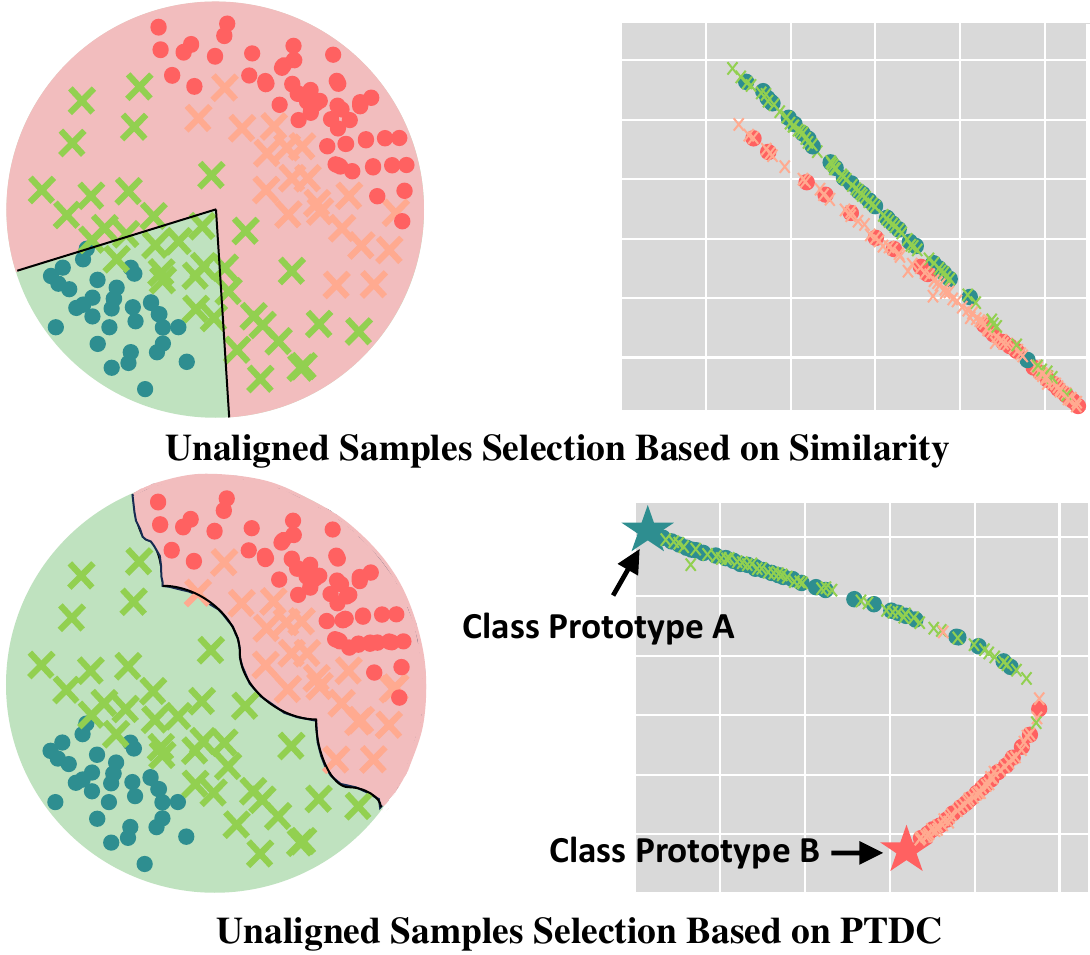}}
  \caption{Comparison of different methods for classifying unaligned and unlabeled samples under the class-imbalanced assumption. The red and green represent different classes, and the $\cdot$ represents aligned and labeled samples, while $\times$ denotes unaligned and unlabeled data. $\star$ is the class prototype.}
  \label{fig:example1}
\end{figure}

Furthermore, we provide a 2D visual example comparing the decision boundaries of cosine similarity and PTDC methods for classifying unaligned samples on ModelNet-10. As shown in Fig. \ref{fig:example1}, the left circle represents the instance space, while the right plot illustrates the latent feature space.
\noindent Most existing VFL methods \cite{kang2022fedcvt,yang2022multi,he2022hybrid} rely on adversarial self-supervised learning \cite{he2022hybrid,feng2022semi,liang2021self,castiglia2022self,li2022semi} or semi-supervised learning \cite{kang2022fedcvt,yang2022multi}. Assuming class imbalance, they do not generalize well to predicting rare classes of unaligned and unlabeled samples using traditional similarity calculations.

\subsection{Zero-Shot and Few-Shot}
To verify the robustness of Proto-EVFL in predicting various rare or unseen classes (zero-shot or few-shot), we simulate extreme class-imbalanced scenarios on the ModelNet-10 dataset. Tab. \ref{tab:multi-class} shows that Proto-EVFL consistently outperforms FedHSSL (SOTA model) across all single and multiple rare or unseen class settings. In few-shot scenarios, Proto-EVFL significantly outperforms FedHSSL, showing a notable increase of 17.10\% with rare classes 1 (bed), 2 (chair), 3 (desk), and 5 (monitor), indicating its effectiveness with very few samples. In zero-shot scenarios, Proto-EVFL remains competitive, with performance increases ranging from 11.16\% to 15.1\%. The consistent performance improvements with various rare or unseen classes demonstrate the robustness of Proto-EVFL in handling extreme class imbalance in VFL.

\begin{figure*}[t]
	\centering
	\subfloat[ModelNet-10] {\includegraphics[width=.23\textwidth]{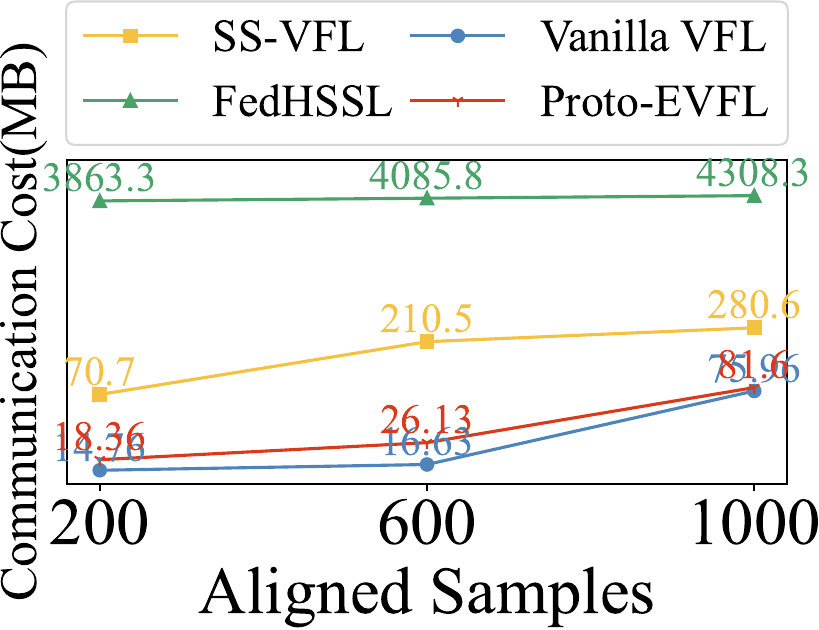}}
	\subfloat[Fashion-MNIST] {\includegraphics[width=.23\textwidth]{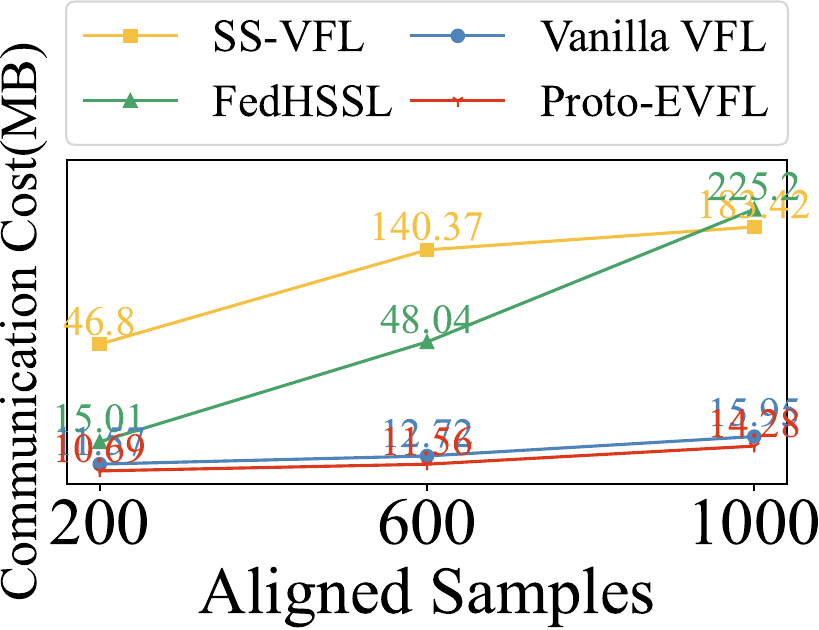}}
	\subfloat[Credit] {\includegraphics[width=.23\textwidth]{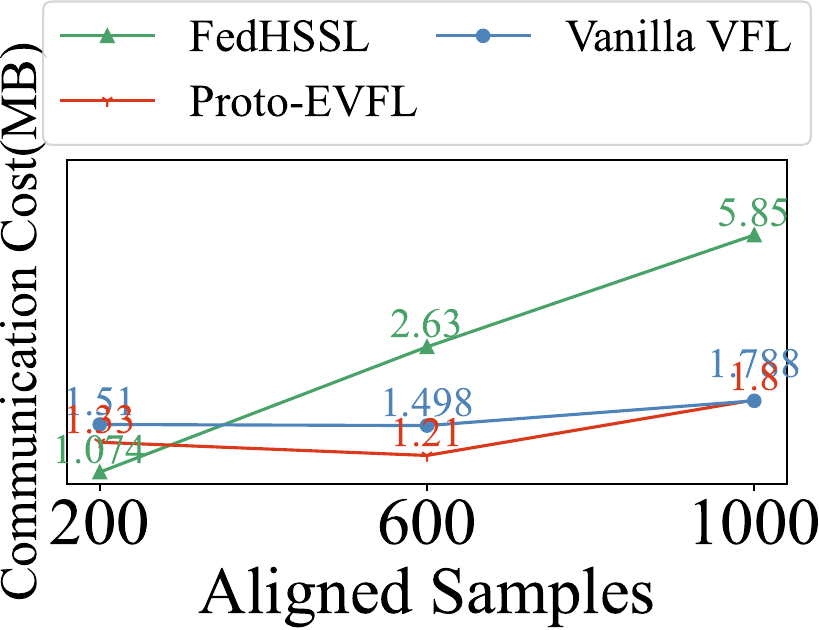}}
        \subfloat[Adult] {\includegraphics[width=.23\textwidth]{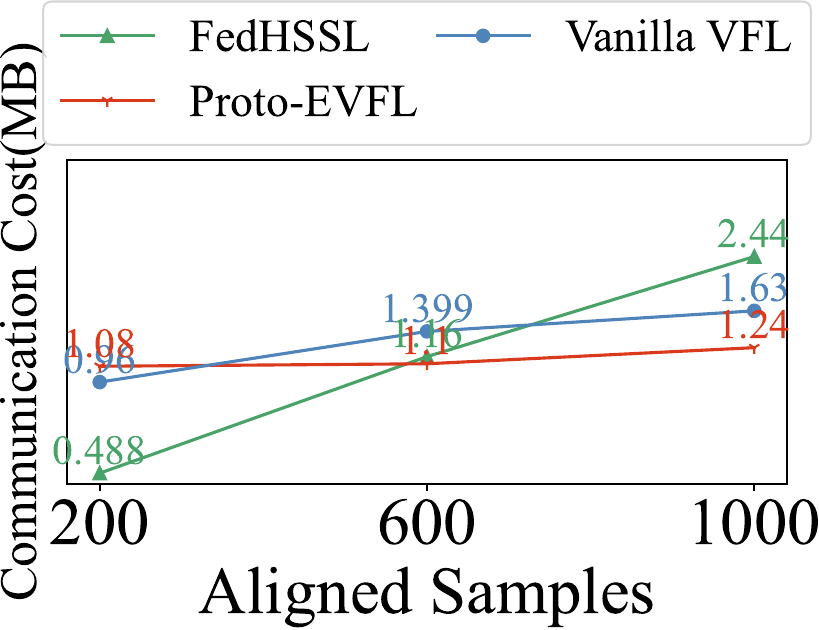}}
	\caption{Communication cost comparisons on four datasets.}
	\label{fig:cost}
\end{figure*}

\begin{figure*}[t]
	\centering
	\subfloat[ModelNet-10] {\includegraphics[width=.24\textwidth]{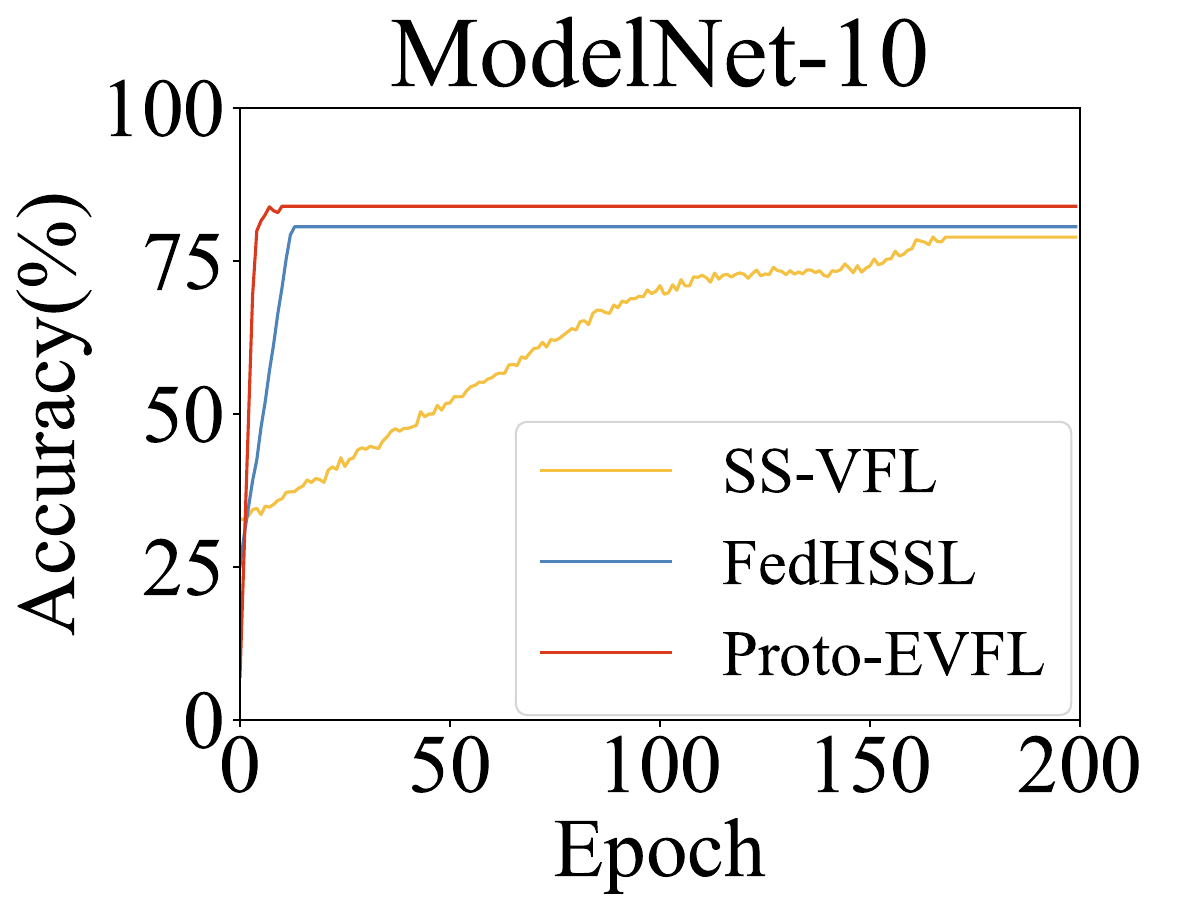}}
	\subfloat[Fashion-MNIST] {\includegraphics[width=.24\textwidth]{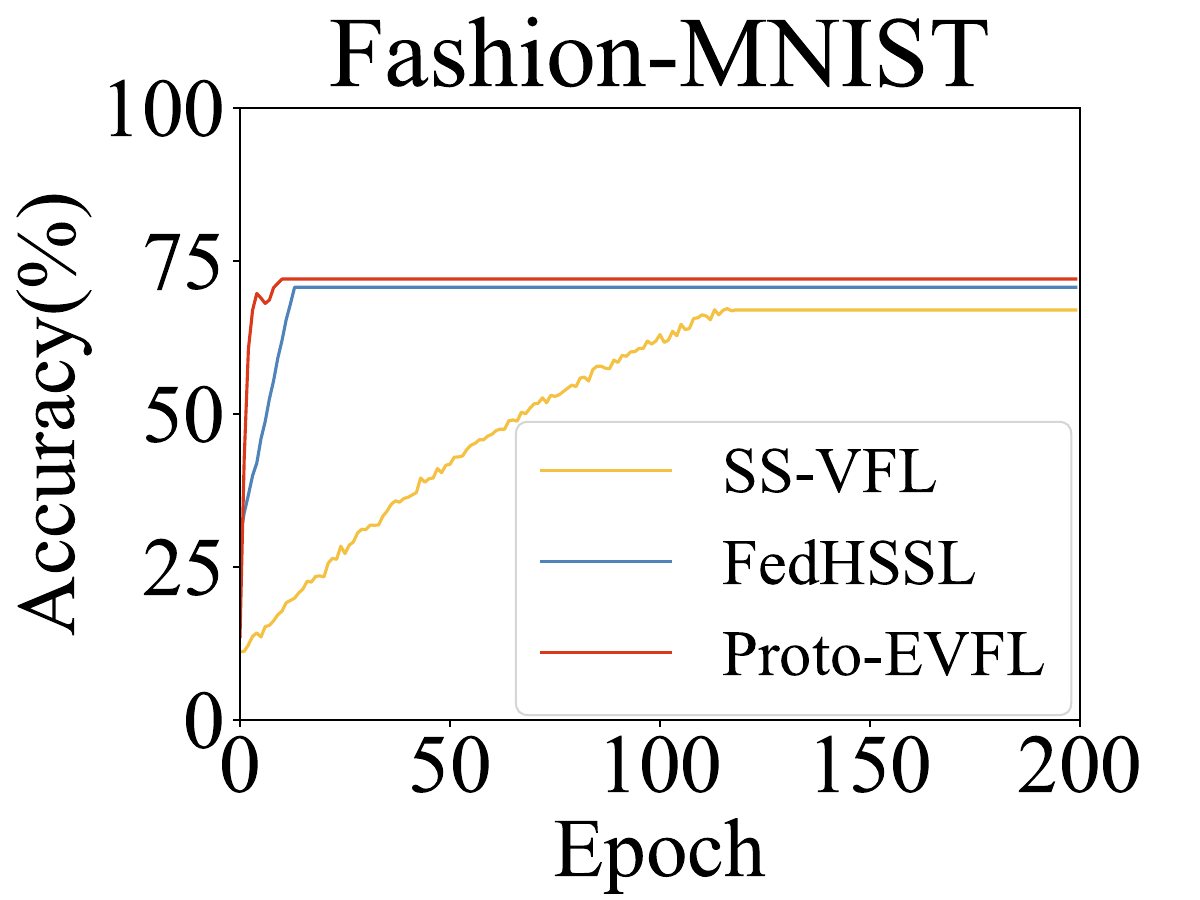}}
	\subfloat[Credit] {\includegraphics[width=.24\textwidth]{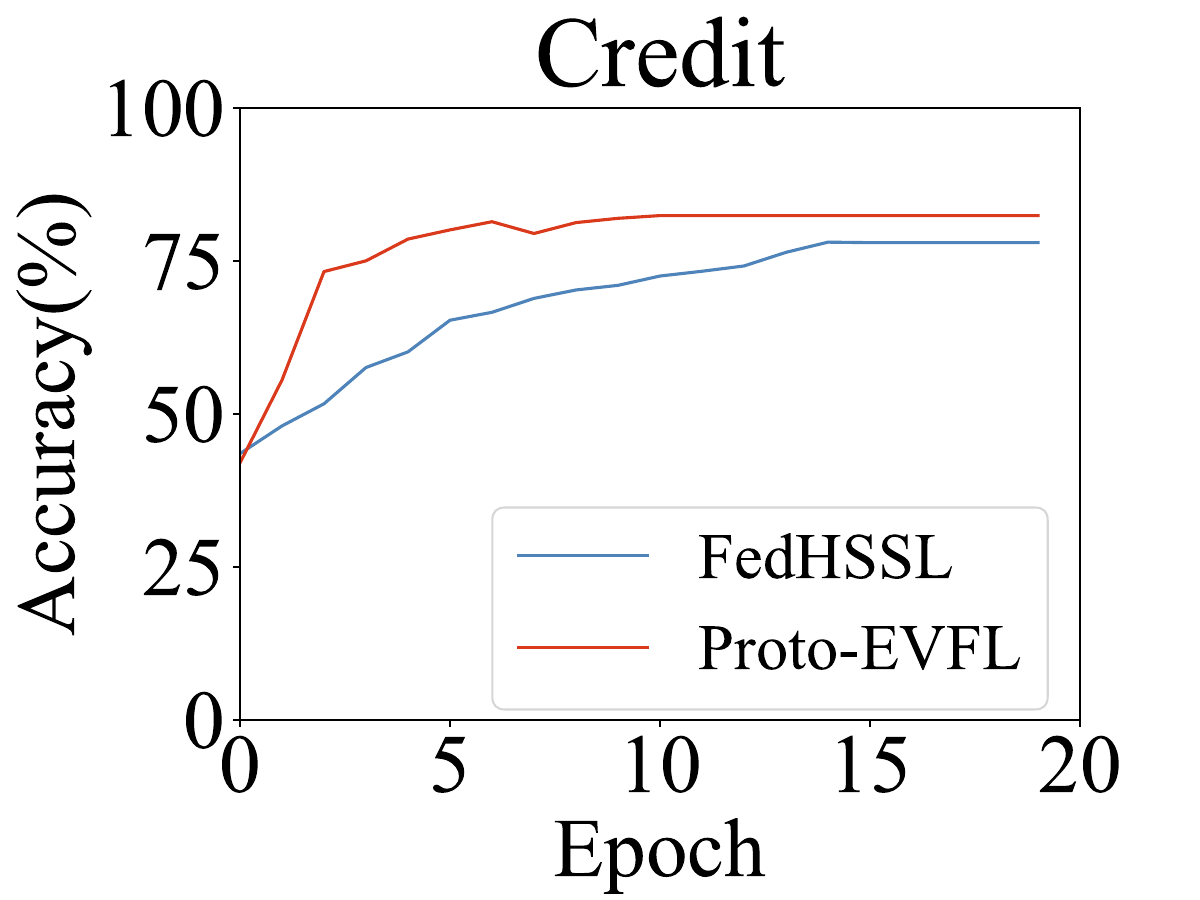}}
        \subfloat[Adult] {\includegraphics[width=.24\textwidth]{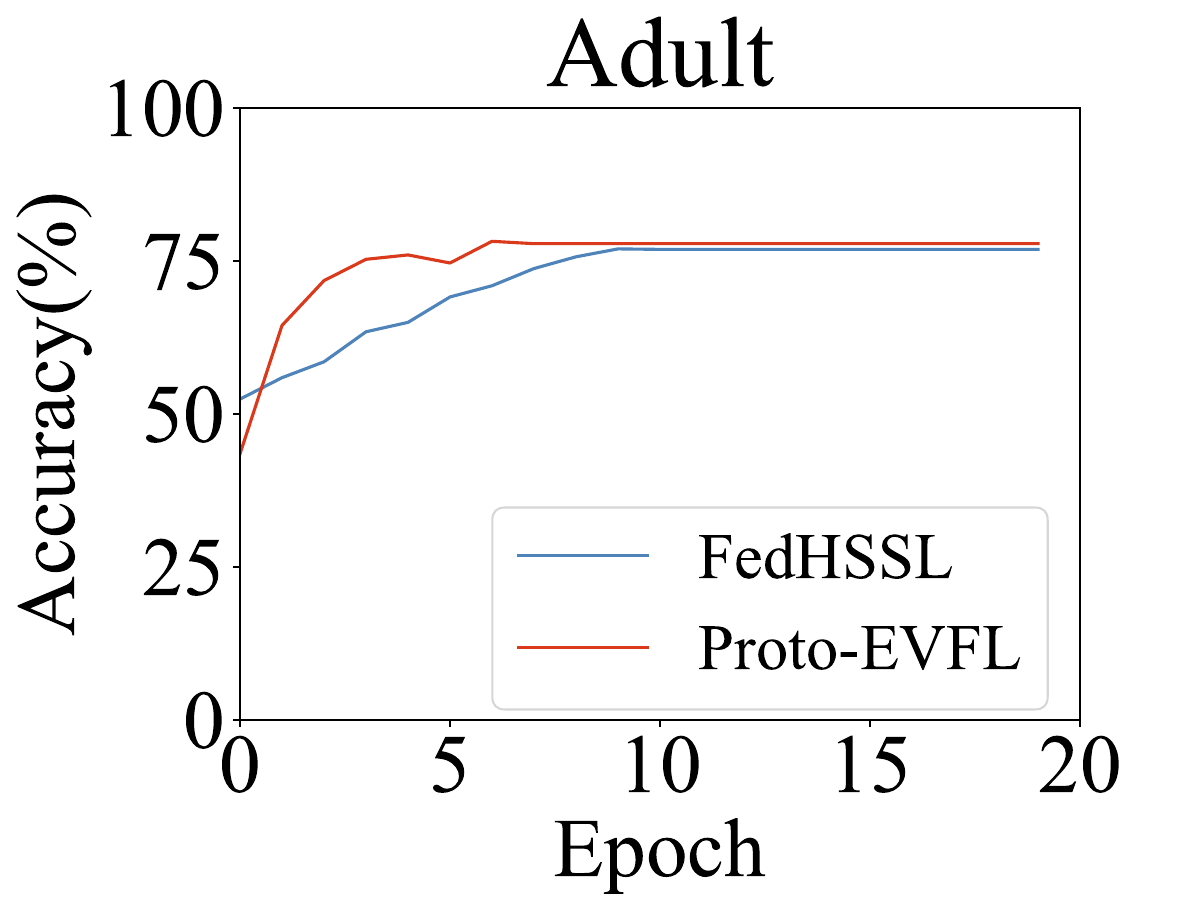}}
	\caption{Convergence comparisons on four datasets.}
	\label{fig:convergence}
\end{figure*}

\begin{figure*}[t]
	\centering
	\subfloat[ModelNet-10] {\includegraphics[width=.24\textwidth]{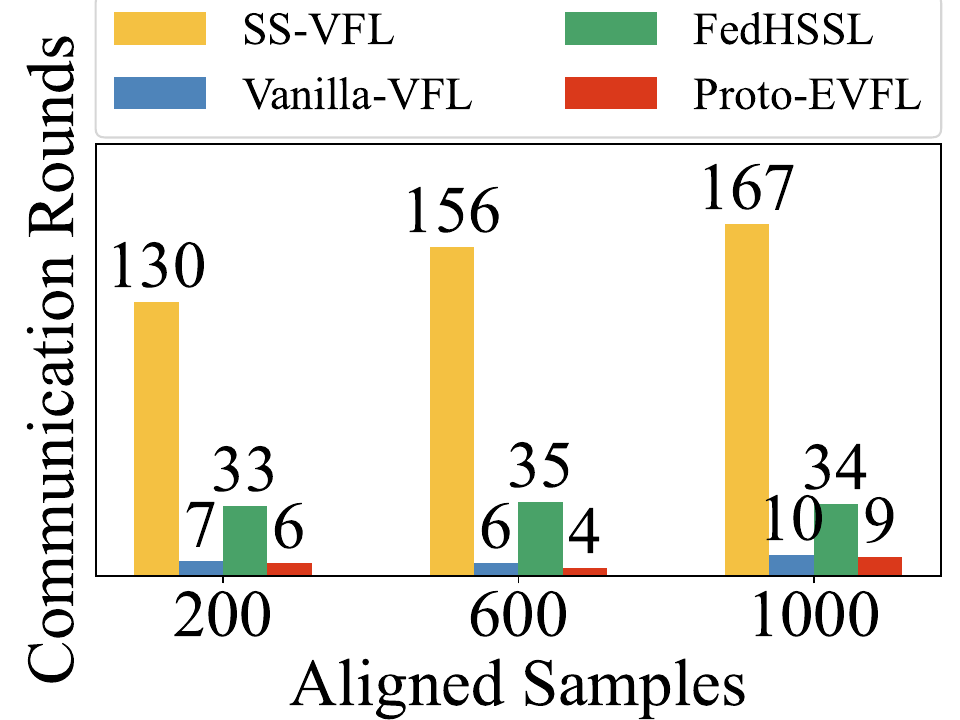}}
	\subfloat[Fashion-MNIST] {\includegraphics[width=.24\textwidth]{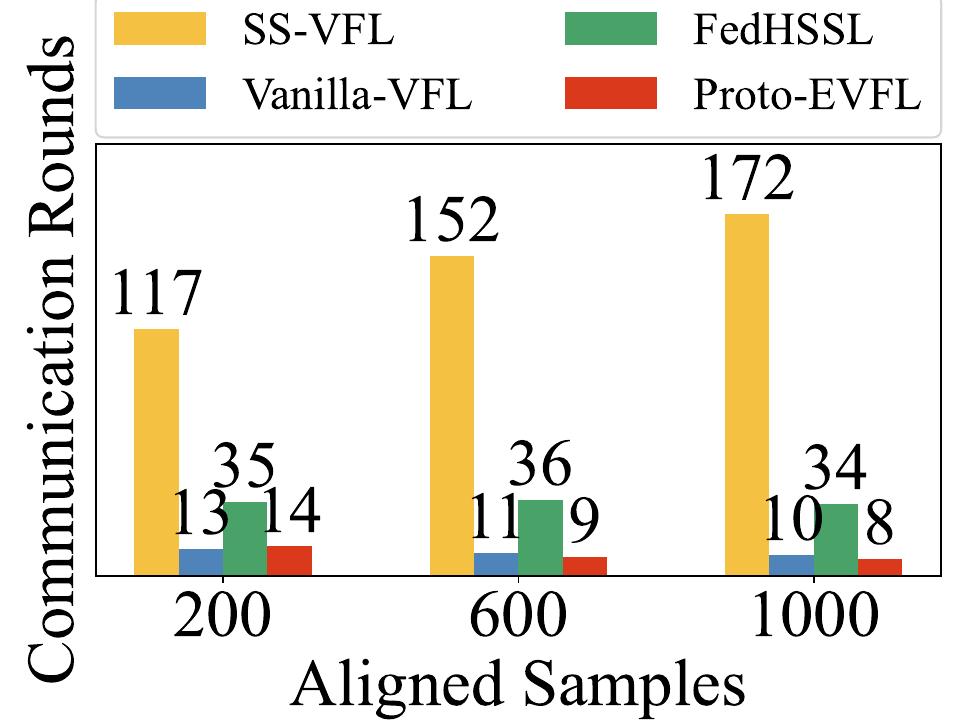}}
	\subfloat[Credit] {\includegraphics[width=.24\textwidth]{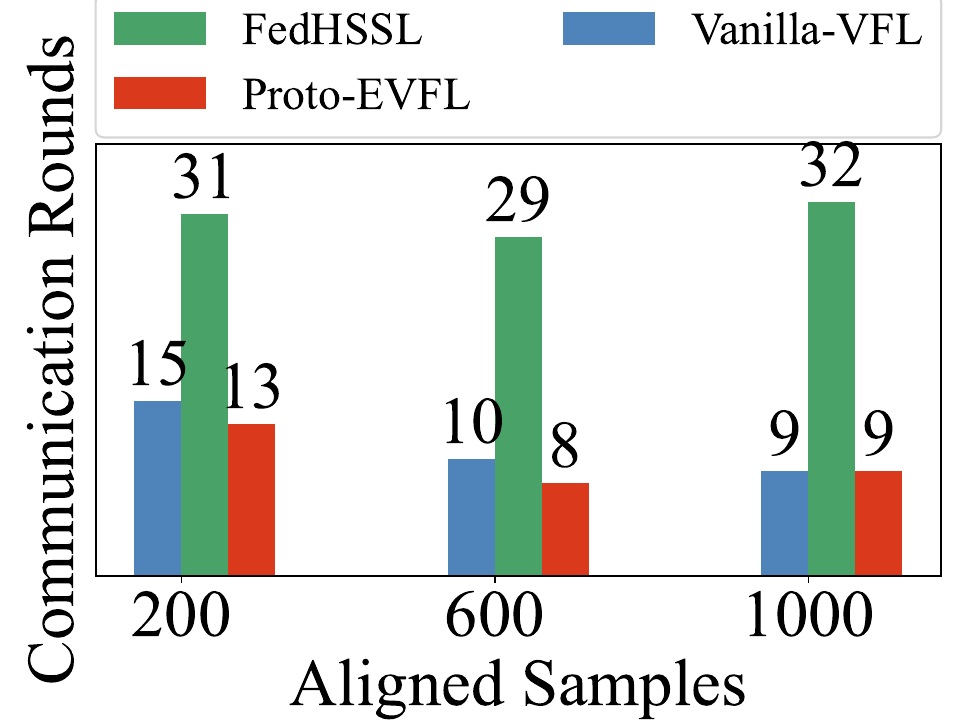}}
        \subfloat[Adult] {\includegraphics[width=.24\textwidth]{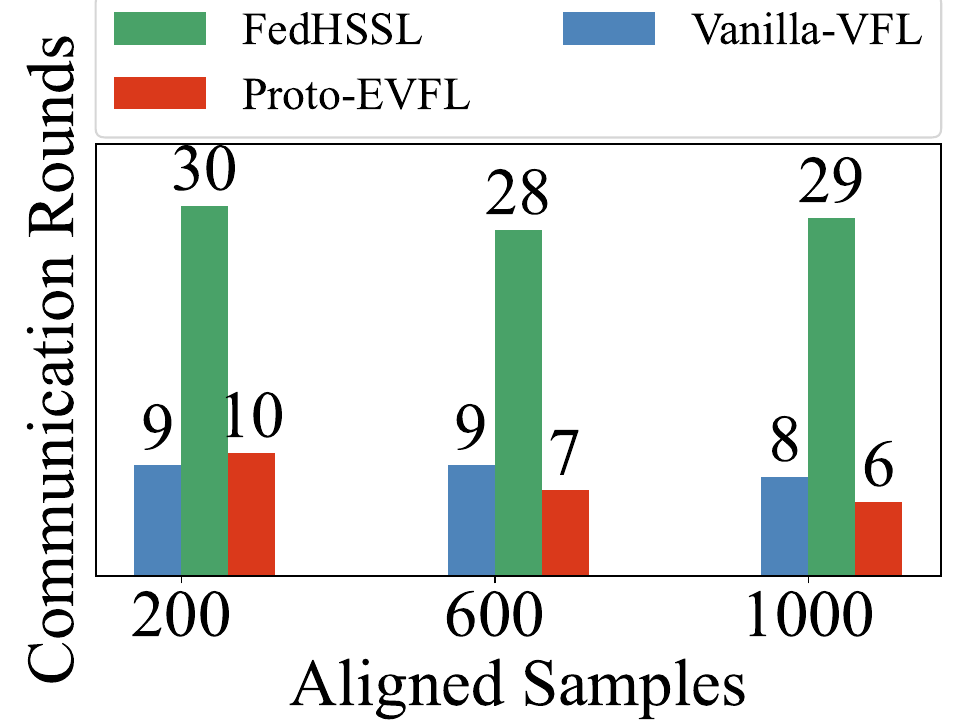}}
	\caption{Communication round comparisons on four datasets.}
	\label{fig:rounds}
\end{figure*}

\subsection{Communication Efficiency}
\textbf{Communication cost.} Fig. \ref{fig:cost} presents the communication cost for baselines to achieve optimal accuracy. Proto-EVFL has a much lower communication cost than SS-VFL and FedHSSL at \#200, \#600, and \#1000 aligned training samples. As target accuracy increases, the communication cost for SS-VFL and FedHSSL escalates substantially in Fig. \ref{fig:convergence}. Although Vanilla VFL has slightly lower costs in ModelNet-10 and Adult with \#600 aligned data, it fails to achieve the same accuracy level as Proto-EVFL, indicating a trade-off between communication efficiency and performance. Overall, Proto-EVFL achieves higher accuracy with lesser communication overhead, making it a more practical choice for real-world applications where communication costs are a limiting factor.

\textbf{Communication round.} 
Fig. \ref{fig:rounds} represents the number of communication rounds required for Proto-EVFL and baselines to reach peak accuracy on the ModelNet-10, Fashion-MNIST, Credit, and Adult datasets. The average rounds needed for Vanilla-VFL, SS-VFL, and FedHSSL are 11.00, 123.50, and 32.25 with \#200 aligned samples. Proto-EVFL outperforms the others, requiring only 10.75 rounds on average to converge while delivering better prediction performance. In scenarios with \#600 and \#1000 aligned data, Proto-EVFL also needs the fewest communication rounds compared to baseline methods.

\subsection{Training Time}
Tab. ~\ref{tab:communication time} presents the number of hours required for each baseline to converge on ModelNet-10. The results denote that Proto-EVFL's training times are significantly lower than the baselines. As the sample size increases to \#1000, Proto-EVFL still maintains a minimal increase in training cost, which indicates its practicability for real-world applications.

\subsection{Discussions}
\label{sec:privacy}
The above experimental results have confirmed that our proposed Proto-EVFL can achieve the SOTA results compared to baselines over several different class-imbalanced scenarios. In the following step, we will further discuss the effectiveness of our approach in more parties and privacy protection problems.

\begin{table}[t]
\centering
\caption{Training time (hours) comparisons between different methods}
\setlength{\tabcolsep}{2.8mm}{
\begin{tabular}{ccccc} \hline
\bottomrule
 \begin{tabular}[c]{@{}c@{}}Aligned\\ samples\end{tabular}  &Vanilla VFL                          & SS-VFL        & FedHSSL       & Proto-EVFL  \\ \cline{2-5} \hline            
\#200                                     & 13.59       & 38.00             & 2.08               & \textbf{0.57}    \\
\#600               &  18.08                &  38.96       &  2.17           & \textbf{0.48}  \\
\#1000      &  75.96                &  39.12      &  2.16             & \textbf{1.04}  \\ \hline
\bottomrule

\end{tabular}
}
\label{tab:communication time}
\end{table}

\begin{table}[]
\centering
\caption{Comparisons across different numbers of parties.}
\setlength{\tabcolsep}{0.3mm}{
\begin{tabular}{cccccc}\hline
\bottomrule
\multirow{2}{*}{}              & \multirow{2}{*}{\begin{tabular}[c]{@{}c@{}}Aligned data \\ number\end{tabular}} & \multicolumn{2}{c}{6 Party} & \multicolumn{2}{c}{8 Party} \\ \cline{3-6}
                               &                                                                                   & FedHSSL     & Proto-EVFL    & FedHSSL     & Proto-EVFL    \\ \hline
\multirow{3}{*}{ModelNet-10}   & \#200                                                                               & 34.40        & 81.92          & 25.74        & 77.53         \\
                               & \#600                                                                               & 71.51        & 82.76          & 65.49        & 79.31          \\
                               & \#1000                                                                              & 81.10          & 84.05         & 77.97          & 82.32         \\ \hline
\multirow{3}{*}{Fashion-MNIST} & \#200                                                                               & 45.72        & 67.23         & 34.45       & 61.04         \\
                               & \#600                                                                               & 62.44        & 69.28          & 57.81        & 65.43          \\
                               & \#1000   & 66.86        & 70.50          & 63.83        & 68.60              \\ \hline
\bottomrule
\label{tab:multi-party}
\end{tabular}
}
\end{table}

\noindent \textbf{More parties.}
Different from HFL, which typically occurs in cross-device scenarios and then can easily scale to hundreds of parties, VFL is primarily designed for enterprise scenarios, which typically involve 2 to 4 parties \cite{liu2022vertical,wu2022practical,he2022hybrid}. To demonstrate the effectiveness of Proto-EVFL in large-scale VFL environments, we evaluate the model performance in scenarios with 6 and 8 parties, as detailed in Tab. \ref{tab:multi-party}. The experimental results indicate that on both the ModelNet-10 and Fashion-MNIST datasets, our method provides superior performance compared to the baseline methods, particularly when the aligned sample size per party is limited. These findings highlight the potential of our approach to enhance performance in large-scale VFL applications.

\noindent \textbf{Privacy protection.}
We validate the privacy of Proto-EVFL from two perspectives: intermediate representations and class prototypes transferred between the active and passive parties.

For class prototypes, we compare the attack accuracy with baselines by label inference attack methods in VFL \cite{fu2022label}. The direct label inference attack method is widely applied in most VFL methods \cite{fu2022label}. It extends HFL gradient attack methods to VFL and proves that VFL methods with gradient feedback have a high probability of leaking the active party's label information. Unlike these VFL methods, Proto-EVFL avoids transferring raw gradient information to passive parties by prototypes, making existing label inference attacks based on gradient information inapplicable \cite{fu2022label}. To perform a label inference attack in Proto-EVFL, we calculate the cosine similarity between aligned data and transferred class prototypes, inferring aligned data's labels. As shown in Tab. \ref{tab:label inference attack}, the attack accuracy of Proto-EVFL is only 8.89\%, significantly lower than the baseline methods by at least 56.65\%.

To assess the privacy leakage risk of intermediate representations, we explore the effect of adding differential privacy (DP) noise \cite{dwork2006differential} to representations on the performance of Proto-EVFL. As shown in Tab. \ref{tab:DP representation}, introducing Gaussian noise to intermediate representations leads to a decrease in Proto-EVFL's accuracy as the noise scale ($\kappa$) increases. This trend highlights the trade-off between privacy protection and model effectiveness. Specifically, when a higher magnitude of noise (e.g., $\kappa = 0.2$) is applied, Proto-EVFL's performance drops below that of Vanilla VFL. This result suggests that while DP is effective in safeguarding intermediate representations, careful control of the noise level (e.g., $\kappa < 0.05$) is essential to preserve model performance. Furthermore, we provide additional insights by examining the impact of noise perturbation on class prototypes, as detailed in Tab. \ref{tab:DP class proto}.

\begin{table}[t]
\centering
\caption{Attack accuracy($\%$) of different methods under label inference attacks.}
\setlength{\tabcolsep}{3.8mm}
{
\begin{tabular}{ccc} \hline
\bottomrule
\multirow{2}{*}{Methods} & \multirow{2}{*}{\begin{tabular}[c]{@{}c@{}}Transfer Data\\ (Active to Passive Party)\end{tabular}} & \multirow{2}{*}{Attack Accuracy$\downarrow$} \\
                         &                                                                                                    &                                  \\ \hline
Vanilla VFL              & Embeddings, Gradients                                                                      & 93.69±0.23                       \\
FedHSSL                  & Models, Gradients                                                                                             & 65.54±0.75                       \\
\cellcolor{gray!30}\textbf{Proto-EVFL}               & \cellcolor{gray!30}\textbf{Embeddings, Prototypes}                                                                             & \cellcolor{gray!30}\textbf{8.89±0.24}     \\  \hline
\bottomrule      
\end{tabular}
}
\label{tab:label inference attack}
\end{table} 

\begin{table}[h]
\centering
\caption{Performance of Proto-EVFL with Gaussian noise of scale $\kappa$ on representations.}
\setlength{\tabcolsep}{2.0mm}{
\begin{tabular}{ccccccc}\hline
\bottomrule
\multirow{3}{*}{\begin{tabular}[c]{@{}c@{}}Aligned data \\ number\end{tabular}} & \multicolumn{6}{c}{Accuracy($\%$)}                                                    \\ \cline{2-7}
                                                                                & \multicolumn{5}{c}{Proto-EVFL with different $\kappa$} & \multirow{2}{*}{Vanilla VFL} \\ \cline{2-6}
                                                                                & 0.00      & 0.01      & 0.05     & 0.10     & 0.20     &                              \\\hline
\#200                                                                           &  81.64       &74.44          & 62.59         & 60.78         &  23.15       & 71.03                             \\
\#600                                                                           &  83.81         & 78.37        & 70.12         &61.39          &  39.49        & 75.12                             \\
\#1000                                                                          &  84.19         & 78.97          & 75.54         & 66.06         & 42.03         &  78.31                           \\  \hline
\bottomrule      
\end{tabular}
}
\label{tab:DP representation}
\end{table}

\begin{table}[h!]
\centering
\caption{Performance of Proto-EVFL with Gaussian noise of scale $\kappa$ on class prototypes.}
\setlength{\tabcolsep}{2.0mm}{
\begin{tabular}{ccccccc}\hline
\bottomrule
\multirow{3}{*}{\begin{tabular}[c]{@{}c@{}}Aligned data \\ number\end{tabular}} & \multicolumn{6}{c}{Accuracy($\%$)}                                                    \\ \cline{2-7}
                                                                                & \multicolumn{5}{c}{Proto-EVFL with different $\kappa$} & \multirow{2}{*}{Vanilla VFL} \\ \cline{2-6}
                                                                               & 0.00      & 0.01      & 0.05     & 0.10     & 0.20     &                               \\\hline
\#200                                                                           & 81.64          & 72.01          & 69.10         &  61.08        &  61.23        &  71.03                            \\
\#600                                                                           & 83.81          &  77.14         & 75.84         & 74.66         &  71.97        & 75.12                             \\
\#1000                                                                          & 84.19          & 78.43          & 76.43         &  75.11        &  74.68        &  78.31                           \\  \hline
\bottomrule      
\end{tabular}
}
\label{tab:DP class proto}
\end{table} 

\section{Conclusion}
In this paper, we mitigated class-imbalanced problems arising from using unaligned samples to alleviate the scarcity of aligned samples in VFL, where class-imbalanced problems include intra-party and inter-party class imbalances. Specifically, we proposed three novel components to tackle these challenges: (i) a probabilistic dual prototype learning scheme to obtain unbiased representations by incorporating class prior probability into loss function; (ii) a mixed prior guided module to mitigate intra-class imbalance by combining local and global priors; (iii) an adaptive gated feature aggregation strategy using a parametric gating network to weight features for the active party's target task. Extensive experiments showed our method outperforms baselines in augmenting feature representation and expanding model prediction space.

\section*{Acknowledgments}
This research work is supported by the National Key Research and Development Program of China under Grant No. 2021ZD0113602, the National Natural Science Foundation of China under Grant No. 62176014, the Fundamental Research Funds for the Central Universities.

\bibliographystyle{IEEEtran}
\bibliography{bare_jrnl_new_sample4}
\begin{IEEEbiography}[{\includegraphics[width=1in,height=1.25in,clip,keepaspectratio]{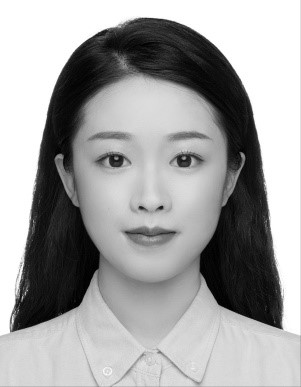}}]{Wei Guo}
She is a Ph.D. student at the Institute of Artificial Intelligence, Beihang University. She received his MSc degree from the School of Electronics and Computer Science at Southampton University. Her research interests primarily lie in federated learning and transfer learning.
\end{IEEEbiography}
\begin{IEEEbiography}[{\includegraphics[width=1in,height=1.25in,clip,keepaspectratio]{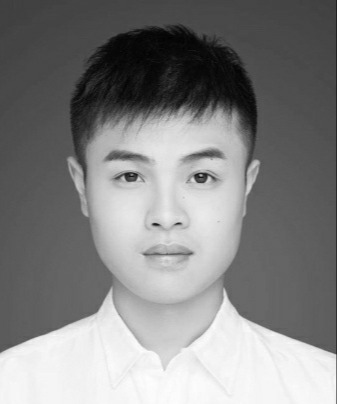}}]{Yiyang Duan}
He is a master's student at the College of Computing and Data Science (CCDS) at Nanyang Technological University. He obtained his bachelor's degree from the School of Artificial Intelligence at Beihang University. His research interest primarily lies in federated learning and recommendation systems.
\end{IEEEbiography}
\begin{IEEEbiography}[{\includegraphics[width=1in,height=1.25in,clip,keepaspectratio]{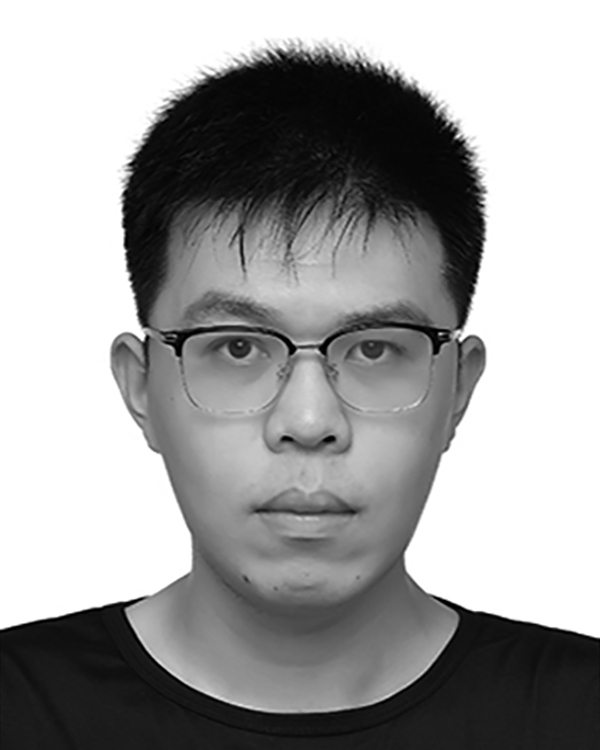}}]{Zhaojun Hu}
He is a Ph.D student at the School of Statistics, Renmin University of China. His research interests primarily lie in federated learning and high-dimensional statistics.
\end{IEEEbiography}
\begin{IEEEbiography}[{\includegraphics[width=1in,height=1.25in,clip,keepaspectratio]{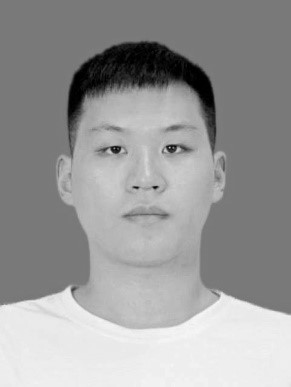}}]{Yiqi Tong}
He is a Ph.D. student at the School of Computer Science and Engineering, Beihang University. He received his MSc degree from the School of Informatics at Xiamen University. His research interests primarily lie in natural language processing and recommendation systems.
\end{IEEEbiography}
\begin{IEEEbiography}[{\includegraphics[width=1in,height=1.25in,clip,keepaspectratio]{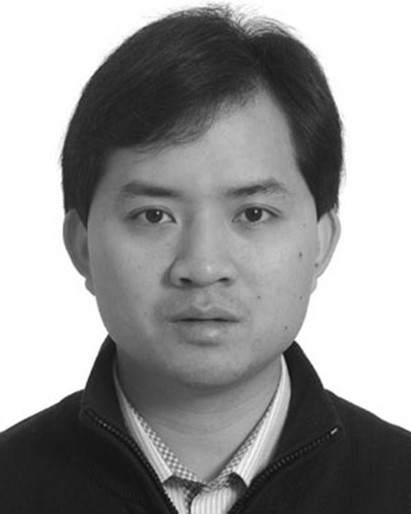}}]{Fuzhen Zhuang}
He received the BE degree from the College of Computer Science, Chongqing University, Chongqing, China, in 2006, and the PhD degree in computer science from the Institute of Computing Technology, Chinese Academy of Sciences, Beijing, China, in 2011. He is currently a full professor with the Institute of Artificial Intelligence, Beihang University. He has published more than 140 papers in some prestigious refereed journals and conference proceedings. His research interests include transfer learning, machine learning, data mining, multitask learning, knowledge graphs, and recommendation systems. He is a senior member of the CCF. He was a recipient of the Distinguished Dissertation Award of CAAI in 2013.
\end{IEEEbiography}
\begin{IEEEbiography}[{\includegraphics[width=1in,height=1.25in,clip,keepaspectratio]{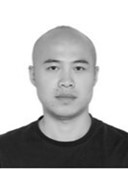}}]{Xiao Zhang}
He is an associate professor at the School of Computer Science and Technology, Shandong University. His research interests include distributed learning, federated learning, edge intelligence, and data mining. He has published more than 20 papers in prestigious refereed journals and conference proceedings, such as IEEE TKDE, TMC, UBICOMP, SIGKDD, SIGIR, IJCAI, ACM CIKM, and IEEE ICDM.
\end{IEEEbiography}
\begin{IEEEbiography}[{\includegraphics[width=1in,height=1.25in,clip,keepaspectratio]{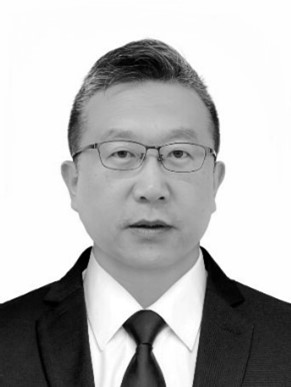}}]{Jin Dong}
He is the General Director of the Beijing Academy of Blockchain and Edge Computing (BABEC), Director of the National Blockchain Technology Innovation Center, Director of Beijing Advanced Innovation Center for Future Blockchain and Privacy Computing. He has been dedicated to technical research in the fields of blockchain, privacy computing, chip design, etc. for years. The team led by him developed the first of kind high performance hardware-software integrated blockchain system - ChainMaker around the globe, aiming to break through the performance and security bottlenecks of large-scale blockchain applications. This has been widely adopted by a variety of key economic and industrial applications in China. Dong Jin received his PhD degree from Tsinghua University and has filed more than forty US patents. 
\end{IEEEbiography}
\begin{IEEEbiography}[{\includegraphics[width=1in,height=1.25in,clip,keepaspectratio]{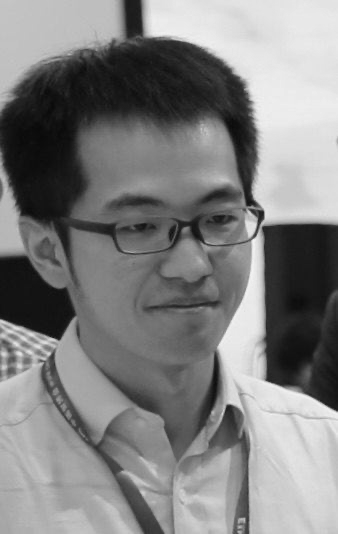}}]{Ruofan Wu}
He received his bachelor's and Ph.D. degrees from Fudan University in 2012 and 2017, respectively. His research interests include graph representation learning and privacy-preserving data analysis. He has published several research papers in machine learning and data mining conferences like NeurIPS, ICLR, KDD, WWW, AAAI, CIKM and ICDM. 
\end{IEEEbiography}
\begin{IEEEbiography}[{\includegraphics[width=1in,height=1.25in,clip,keepaspectratio]{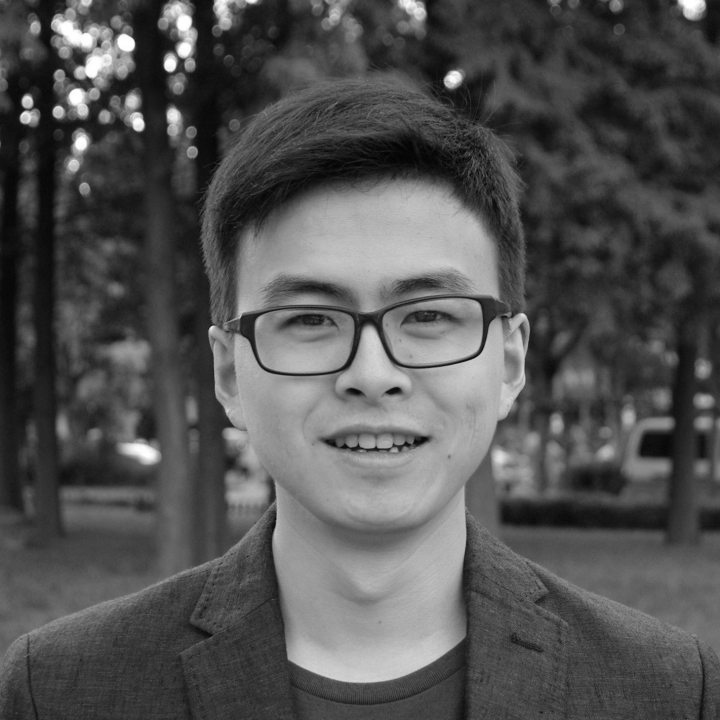}}]{Tengfei Liu}
He earned his Bachelor’s degree in Computer Science from the University of Science and Technology of China (USTC) and completed his Ph.D. in Computer Science and Engineering at the Hong Kong University of Science and Technology (HKUST).  His research has been featured in several leading conferences and journals, including but not limited to NeurIPS, KDD, AAAI, IJCAI, CIKM, AIJ, MLJ, and JAIR. 
\end{IEEEbiography}
\begin{IEEEbiography}[{\includegraphics[width=1in,height=1.25in,clip,keepaspectratio]{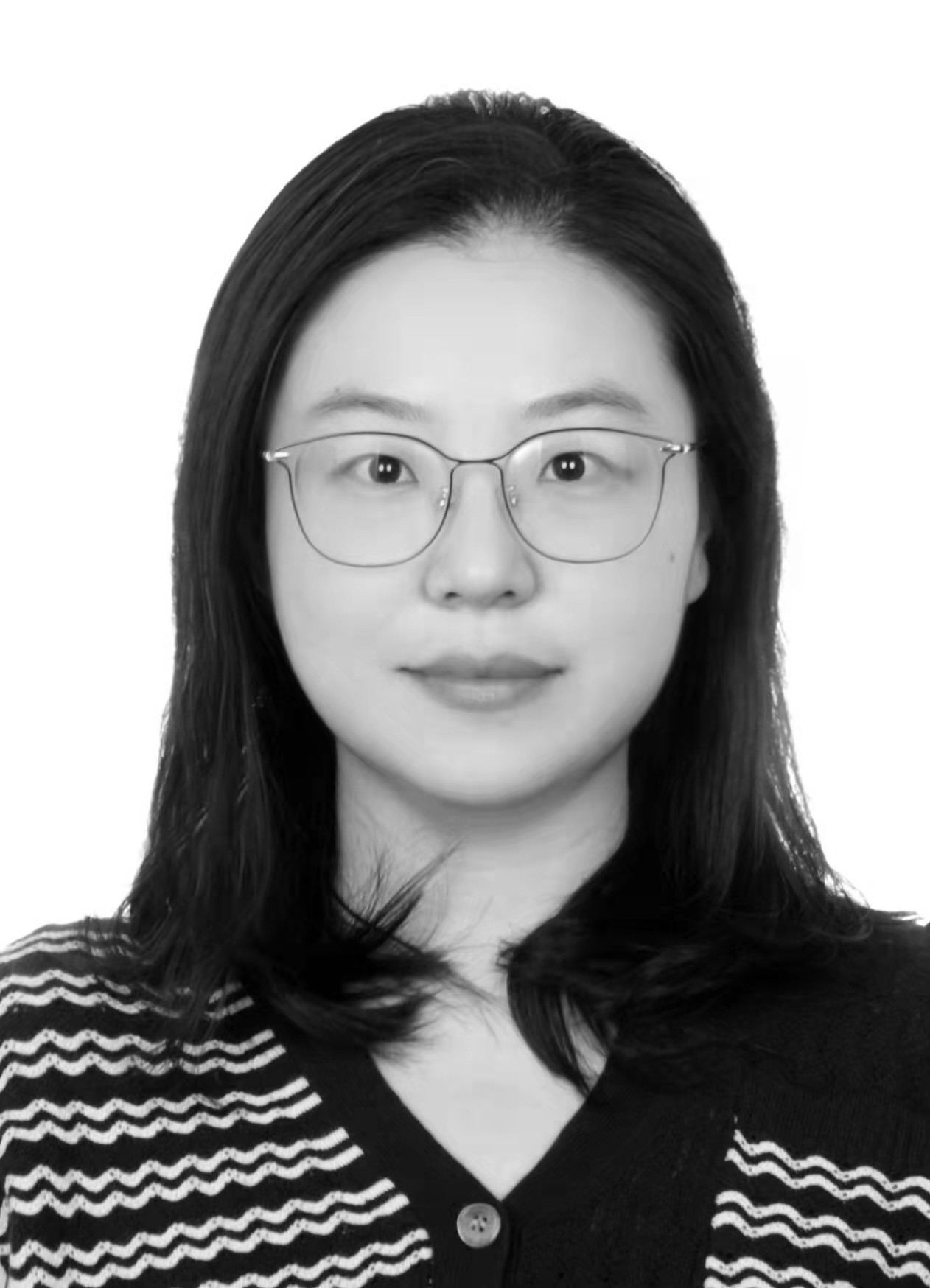}}]{Yifan Sun}
She is a professor at the School of Statistics, Renmin University of China, and a researcher at the Center for Applied Statistics. Her research interests include federated learning, statistical inference, and distributed optimization. Yifan Sun received her PhD degree from Beihang University.
\end{IEEEbiography}

\vfill

\clearpage
\appendix

\subsection{Notations}
\label{sec:notations}
For easier check, we have compiled our main variables in the Tab. \ref{tab:notations}.

\begin{table*}[]
\centering
\caption{Notations}
\setlength{\tabcolsep}{7mm}{
\begin{tabular}{clcl} \\  \hline
\bottomrule      
Symbol                            & \multicolumn{1}{c}{Definition}                                      & Symbol               & \multicolumn{1}{c}{Definition}                                  \\ \hline
$M$                               & Number of parties                              & $m$                  & Party $m$                                  \\
$N$                               & Number of samples                              & $a$                  & Aligned sample                             \\
$Z$                               & Number of classes                              & $u$                  & Unaligned sample                           \\
$T$                               & Number of communication round                  & $x$                  & Sample                                     \\
$D$                               & Feature dimensions                             & $y$                  & Label                                      \\
$X$                               & Sample set                                     & $\mu^{m}$            & Class prototype of $m$                     \\
$Y$                               & Ground truth label Set                         & $\mu^{g}$            & Global class prototype                     \\
$C$                               & Classifier                                      & $f$                  & Intermediate representation                \\
$E$                               & Extractor                                      & $\hat{f}$            & Updated intermediate representation by $R$ \\
$R$                               & Adaptor                                        & $p$                  & Probabilistic distribution                 \\
$G$                               & Gating network                                 & $t$                  & Communication round                        \\
$B$                               & Batch size                    & $\pi$                & Conditional distribution                   \\
$\mathcal{B}$                     & Batch                                & $\varphi$            & Hyperparameter                             \\
$P$                               & Prior probability                              & $\rho$               & Hyperparameter                             \\
$W$                               & Aggregation weight                             & $\varrho$            & Adaptive parameter                   \\
$\Theta$                          & Model parameter set of $G$,$C$,$R$            & $\gamma$             & Adaptive parameter                          \\
$\mathcal{Z}$                     & All model parameter of parties                 & $\omega$             & Cross-entropy loss                         \\
$\hat{\Theta}$                     & An $\epsilon$-accurate stationary point       & $\hat{z}$            & A local minority class                    \\
$\mathcal{E}$                     & Extractor parameter set of Extractor of parties & $\theta$             & Model hyperparameter                       \\
$\theta_{E^{*}}$                  & Optimal solution of $\theta_{E}$    & $\sigma$             & Variance                                   \\
$\mathcal{E}^{*}$                 & Optimal solution of $\mathcal{E}$               & $\zeta$              & Unaligned dataset of all parties           \\
$F(\cdot)$                        & Decision function                         & $\xi$                & A sample of active party                   \\
$\mathcal{L}_{f \rightarrow \mu}$ & PTDC from $f$ to $\mu$                          & $\tau$               & Epochs                                     \\
$\mathcal{L}_{\mu \rightarrow f}$ & PTDC from $\mu$ to $f$                             & $\eta$               & Learning rate of $E$                       \\
$\mathcal{L}_{{\rm local}}$       & Loss function of each passive/active party                & $\eta^{\prime}$      & Learning rate of $G$,$C$,$R$                   \\
$\mathcal{L}_{{\rm global}}$      & Loss function of active party             & $c(\cdot)$           & Cosine dissimilarity           \\  \hline
\bottomrule      
\end{tabular}
}
\label{tab:notations}
\end{table*}

\subsection{Detailed Probabilistic Dual prototype learning scheme}
\subsubsection{Connecting with Entropy Minimization}
\label{sec:entropy minimization}
To understand more clearly the expected cost from feature representation to class prototypes, we can think of it as an extension of entropy minimization \cite{grandvalet2004semi}, an effective form of regularization that is widely used in many previous prior domain adaptation studies \cite{saito2019semi,vu2019advent,saito2020universal,balaji2020robust}.

\textbf{Likelihood.} We begin by revisiting how semi-supervised learning challenges fit into the general supervised learning framework based on the maximum (conditional) likelihood estimation principle. The learning set is represented as $X_{N} = \{x_{n}, y'_{n}\}^{N}_{n=1}$, where $y' \in \{0,1\}^{Z}$ represents the dummy variable, which refers the labels that are actually available. Furthermore, $y$ still denotes the correct and complete class information. Assuming that $x_{n}$ is labeled as $\delta_{z}$, we can obtain $y'_{n,z} = 1$ and $y'_{n,l} = 0$ ($z \neq l$). If $x_{n}$ is not labeled, it can be represented as $y'_{i,z} = 1$ ($l = 1, ..., Z$). 

If the label is randomly missing, it means that $P(y'|x,\delta_{z}) = P(y'|x,\delta_{l})$ for all unlabeled examples, and then for any $(\delta_{z},\delta_{l})$, which shows:
\begin{equation}
    P(\delta_{z}|x,y') = \frac{y'_{z}P(\delta_{z}|x)}{\sum^{Z}_{l=1}y'_{l}P(\delta_{l}|x)}.
\end{equation}
Under the independent samples assumption, the conditional log-likelihood of the observed sample $(Y'|X)$ is:
\begin{equation}
\label{equ:log-likelihood}
    L(\theta,X_{N}) = \sum_{n=1}^{N}\log(\sum^{Z}_{z=1}y'_{n,z}\mathcal{L}_{z}(x_{n};\theta)) + h(y'_{n}),
\end{equation}
where $h(y'_{n})$ is not dependent on $P(X|Y)$, and only influenced by label missing mechanism, which refers to semi-supervised learning can be viewed as a missing data challenge \cite{mclachlan2005discriminant} in probabilistic framework. $\mathcal{L}_{z}(x_{n}$ denotes the model of $P(\delta_{z}|x)$, which is parameterized by $\theta$. 

\textbf{Probabilistic model optimization.}
Given that $\mathcal{F}_{z}(x;\theta)$ represents the model of $P(\delta_{z}|x)$, the model of $P(\delta_{z}|x,y')$ can be described as:
\begin{equation}
    g_{z}(x,y';\theta) = \frac{y'_{z}\mathcal{F}_{z}(x;\theta)}{\sum^{Z}_{l=1}y'_{l}\mathcal{F}_{l}(x,\theta)},
\end{equation}
where $g_{z}(x,y';\theta) = y'_{z}$ for labeled samples. And for unlabeled samples, $g_{z}(x,y';\theta) = \mathcal{F}_{z}(x;\theta)$. To simplify notation, the reference to parameter $\theta$ in $\mathcal{F}_{z}$ and $ g_{z}$ is dropped below. Therefore, the maximum a posteriori probability estimation (MAP) is achieved by:
\begin{footnotesize}
\begin{equation}
\begin{aligned}
    C(\theta,\beta;X_{N}) &= L(\theta,X_{N}) - \beta H_{{\rm emp}}(Y|X,Y';X_{N}) \\
    &= \sum^{N}_{n=1}\log(\sum^{Z}_{z=1}y'_{n,z}\mathcal{F}_{z}(x_{n})) \\
    &+ \beta\sum^{N}_{n=1}\sum^{Z}_{z=1}g_{z}(x_{n},y'_{n})\log g_{z}(x_{n},y'_{n}),
\end{aligned}
\end{equation}
\end{footnotesize}
\noindent where $L(\theta,X_{N})$ is only effected by labeled samples. $H(Y|X, Y'; X_{N})$ measures the conditional entropy of class overlap, which is related to the effectiveness of unlabeled data. Therefore, we can observe that maximizing the posterior probability requires minimizing entropy.

\textbf{In our method.} If the moving cost from point to point in the Eq. \ref{equ:Lf--mu} is denoted as:
\begin{equation}
    c(\mu^{m}_{z}, f^{m,n}_{u}) = -log{p}^{m,n}_{z} = -\log\frac{{\rm exp}(\mu^{m}_{z}f^{m,n}_{u})}{\sum^{Z}_{z'=1}{\rm exp}(\mu^{m}_{z'},f^{m,n}_{u})},
\end{equation}
and the conditional probability is represented as:
\begin{equation}
     \pi_{\theta_{E}}(\mu^{m}_{z}|f^{m,n}_{u})) = {p}^{m,n}_{z} = \frac{{\rm exp}(\mu^{m}_{z}f^{m,n}_{u})}{\sum^{Z}_{z'=1}{\rm exp}(\mu^{m}_{z'},f^{m,n}_{u})},
\end{equation}
and then the expected moving cost is: 
\begin{equation}
    \mathcal{L}_{f \rightarrow \mu} = -\mathbb{E}_{X^{m,n}_{u} \sim X^{m}_{u}}[\sum^{Z}_{z'=1}{p}^{m,n}_{z}\log{p}^{m,n}_{z}],
\end{equation}
which can be regarded as minimizing the entropy of feature representation of local unaligned samples. 

\subsubsection{Local prior probability estimation}
\label{sec:prior estimation}
To estimate local class proportions $\{P(\mu^{m,t}_{z})\}^{Z}_{z=1}$ of party $m$ in round $t$, we adopt the EM algorithm to infer the iterative updates of them. 

The target label of each data point, $Y^{m,n}$, can not be directly obtained by observation. We view it as a latent variable. Considering that the proportion of classes is important to match unaligned and unlabeled samples with class prototypes, we adopt a natural method by maximizing the likelihood on the local dataset:
\begin{equation}
\begin{aligned}    
    &L(\{P(\mu^{m,t}_{z})\}^{Z}_{z=1}|X^{m,1}, ..., X^{m, N^{m}}) \\
    &= \sum^{N^{m}}_{n=1}\ln[\sum^{Z}_{z=1}P(\mu^{m,t+1}_{z})p_{\theta,\mu} 
    (X^{m,j}|Y^{m,n}=z)],
\end{aligned}
\end{equation}
where $p_{\theta,\mu}(X^{m,j}|Y^{m,n}=z) = \frac{{\rm exp}(P(\mu^{m,t+1}_{z}))f^{m,n}_{u}}{V}$, and $V$ is a normalizing constant. Note that $P(\mu^{m,t}_{z})$ is obtained from the active party that uses classification loss to optimize the adaptor of each party, and then to update their prototypes. It aims to integrate the feature information of other parties into the local prototypes, and then achieve the selection of local unaligned samples in a global view.

Considering the difficulty of maximizing the log-likelihood of the unaligned samples, We employ the expectation maximization (EM) algorithm to iteratively optimize between the expectation and maximization steps. At the beginning of EM updates, the class prior probability is initialized with a uniform prior, $P(\mu_{z}^{0,t}) = \frac{1}{Z}, \forall z = 1, 2, ..., Z$. At each step $l$, we implement the E-step and M-step as shown in the below:

\textbf{E-step:} We use the previous estimates to calculate the posterior probability of the local unaligned and unlabeled samples belonging to class $z$. The posterior probabilities correspond to the weights of the transport cost of moving from unaligned samples to the class prototypes.     
\begin{equation}
  \begin{aligned}
    p_{\theta,\mu}(Y^{m,n} &= z|X^{m,n}_{u},p(\mu^{m,t}_{z})^{l}) \\
    &= \pi_{\theta}(\mu^{m,t}_{z}|f^{m,n}_{u},p(\mu^{m,t}_{z})^{l}) \\
    &= \frac{p(\mu_{z}^{m,t})^{l}{\rm exp}(\mu_{z}^{m,t+1}f^{m,n}_{u})}{\sum^{Z}_{z'=1}P(\mu^{m,t}_{z'})^{l}{\rm exp}(\mu^{m,t+1}_{z'},f^{m,n}_{u})}.
  \end{aligned}
\end{equation}

\textbf{M-step:} We obtain the log-likelihood of the full local dataset by $\sum^{B}_{n=1}\ln[p(\mu^{m,t}_{z})^{l+1}p_{\theta,\mu}(X^{m,n}_{a}|Y^{m,n})]$. Therefore, the expected log-likelihood can be calculated by:
\begin{equation}
\begin{aligned}
    p(\mu^{m,t}_{z})^{l+1} &= \mathop{\rm argmax}\limits_{p(\mu^{m,t}_{z})^{l+1}}L,\\
    &:= \mathop{\rm argmax}\limits_{p(\mu^{m,t}_{z})^{l+1}}\sum^{N^{m}_{a}}_{n=1}\sum^{Z}_{z=1}p_{\theta,\mu}(Y^{m,n})\\
    &=z|X^{m,n}_{z},p(\mu^{m,t}_{z})^{l+1})\ln[p(\mu^{m,t}_{z})^{l+1}\\
    &p_{\theta,\mu}(X^{m,n}_{u}|Y^{m,n}=z)] + \lambda(\sum^{Z}_{z=1}p(\mu^{m,t}_{z})^{l+1}-1),
\end{aligned}
\end{equation}

where $\lambda$ is a Lagrange multiplier to influence $p(\mu^{m}_{z})$ to be simplex. 
\begin{equation}
     \frac{\partial L}{\partial \lambda} = \sum^{Z}_{z=1}p(\mu^{m,t}_{z})^{l+1} - 1 = 0,
\end{equation}
\begin{equation}
\begin{aligned}
     \frac{\partial L}{\partial p(\mu^{m,t}_{z})^{l+1}} &= \sum^{N^{m}}_{n=1}p_{\theta,\mu}(Y^{m,n} \\
     &= z|X^{m,n},p(\mu_{z}^{m,t})^{l})\frac{1}{p(\mu^{m}_{z})^{l+1}} + \lambda = 0.
\end{aligned}
\end{equation}
Both sides of the above are multiplied by $p(\mu^{m}_{z})^{l+1}$, summed over $z$. Then, we can update: 
\begin{equation}
\begin{aligned}
    p(\mu_{z}^{m})^{l+1} &= \frac{1}{N^{m}}\sum^{N^{m}}_{n=1}p_{\theta,\mu}(Y^{m,n}=k|X^{m,n},p(\mu^{m}_{z})^{l}) \\
    &= \frac{1}{N^{m}}\sum^{N^{m}}_{n=1}\pi(\mu^{m}_{z}|f^{m,n}_{u},p(\mu^{m}_{z})^{l}),
\end{aligned}
\end{equation}
where $\lambda = - N^{m}$. In our experiments, we select a mini-batch $B$ to estimate this prior. 

\subsection{Convergence Analysis}
\label{sec:convergence analysis}
We provide the full proof of Theorem 1 and more general results about arbitrary $\tau$ and $B$ in the next. To give the convergence analysis of our algorithm, we need to give some useful lemmas first.

Firstly, with $L$-smoothness of $l(\Theta,\mathcal{E};\zeta)$, we could give the strongly convexity property of $l_{{\rm local}}(\Theta,\mathcal{E};\zeta)$.

\noindent\textbf{Lemma 1.} Under Assumption 2, suppose $\alpha:=-L+\varphi >0$, $l_{{\rm local}}(\Theta,\mathcal{E};\zeta)$ is $\alpha$-strongly convex w.r.t $\mathcal{E}$. 

Based on Assumptions 1 and 2, we could give Lemma 2 directly.

\noindent\textbf{Lemma 2.} Under Assumption 1 and 2, the derivatives $\nabla l_{{\rm global}} (\mathcal{Z}; \xi)$, $\nabla_{\Theta}\nabla_{\mathcal{E}}l_{{\rm local}}(\mathcal{Z};\zeta)$ and $\nabla_{\mathcal{E}}^2 l_{{\rm local}}(\mathcal{Z};\zeta)$ have bounded variances, i,e., for any $\mathcal{Z}$, we have:
\begin{equation}
    \mathbb{E}_{\xi} \|\nabla l_{{\rm global}} (\mathcal{Z};\xi) - \nabla \mathcal{L}_{{\rm global}} (\mathcal{Z})\|^2 \leq L_1^2,
\end{equation}
\begin{equation}
    \mathbb{E}_{\zeta} \|\nabla_{\Theta}\nabla_{\mathcal{E}}l_{{\rm local}} (\mathcal{Z};\zeta) - \nabla_{\Theta}\nabla_{\mathcal{E}}\mathcal{L}_{{\rm local}} (\mathcal{Z})\|^2 \leq L_2^2,
\end{equation}
\begin{equation}
    \mathbb{E}_{\zeta} \|\nabla_{\mathcal{E}}^2 l_{{\rm local}} (\mathcal{Z};\zeta) - \nabla_{\mathcal{E}}^2\mathcal{L}_{{\rm local}} (\mathcal{Z})\|^2 \leq L_2^2.
\end{equation}

As the proof of Lemma 1 and 2 is too trivial, we omit it. To show the smooth property of $F (\Theta)$, we first introduce the following lemma which is proposed in \cite{ghadimi2018approximation}.

\noindent\textbf{Lemma 3. (Lemma 2.2 in \cite{ghadimi2018approximation})} Under Assumptions 1, 2 and 3, $F (\Theta)$ is $L_0$-smooth where $L_0$ is given by:
\begin{small}
\begin{equation*}
    L_0:= L_2 +\frac{2L_2^2 + L_1^2 L_3}{\alpha}+ \frac{L_1L_2L_3+L_1L_2L_4+L_2^3}{\alpha^2}+\frac{L_1L_2^2L_4}{\alpha^3}.
\end{equation*}
\end{small}

Tracking error $\mathbb{E}\|\mathcal{E}_t^{j-1} - \mathcal{E}^*(\Theta_t^0)\|$ is an important component in our convergence analysis. To give an upper bound on the tracking error, we utilized Lemma 9 in \cite{ji2021bilevel}.

\noindent\textbf{Lemma 4. (Lemma 9 in \cite{ji2021bilevel})} Under Assumptions 1, 2 and 4, with step size $\eta$ to be $\frac{2}{L_2 + \alpha}$, we have:
\begin{small}
\begin{equation}
    \mathbb{E}\|\mathcal{E}_t^{j-1} - \mathcal{E}^*(\Theta_t^0)\|^2 \leq \left(\frac{L_2 - \alpha}{L_2 +\alpha}\right)^{2(j-1)}\mathbb{E}\|\mathcal{E}_t^{0} - \mathcal{E}^*(\Theta_t^0)\|^2 + \frac{\sigma^2}{L_2\alpha B}. 
\end{equation}
\end{small}

With the help of the above lemmas, we could give the estimation property of the $\frac{\partial l_{{\rm global}} (\Theta_t^0,\mathcal{E}_t^\tau)}{\partial \Theta_t^0}$ approximating $\nabla F(\Theta_t^0)$. The result is presented in the following Proposition 1. For the sake of clarity, we introduce a new notation $\frac{\partial l_{{\rm global}} (\Theta_t^0,\mathcal{E}_t^\tau;\mathcal{B}_0)}{\partial \Theta_t^0}:= \frac{1}{B}\sum_{i \in \mathcal{B}_0} \frac{\partial l_{{\rm global}} (\Theta_t^0,\mathcal{E}_t^\tau;\xi_i)}{\partial \Theta_t^0}$. Other notations involved $\mathcal{B}_j$ or $\mathcal{B}_j$ such as $\nabla_{\Theta}\nabla_{\mathcal{E}}l_{{\rm local}}(\Theta_t^0,\mathcal{E}_t^{j-1};\mathcal{B}_{j-1})$ have similar meanings.

\noindent\textbf{Proposition 1.} Under Assumptions 1-5, choose step size $\eta$ to be $\frac{2}{L_2 + \alpha}$ and suppose $\alpha<L_2$, we have:
\begin{equation}
    \begin{aligned}
        \mathbb{E}&\|\frac{\partial l_{{\rm global}} (\Theta_t^0,\mathcal{E}_t^\tau;\mathcal{B}_0)}{\partial \Theta_t^0} - \nabla F(\Theta_t^0)\|\\
        &\leq (L_2 + \frac{L_2^2}{\alpha}\Big [\left(\frac{L_2 - \alpha}{L_2 +\alpha}\right)^{\tau} \\
        &\sqrt{\Delta}+ \frac{\sigma}{\sqrt{L_2\alpha B}}\Big ]+L_1\Big [\frac{L_2 (1-\frac{2}{L_2+\alpha} \alpha)^{\tau}}{\alpha}\\
        &+\frac{1}{\alpha\sqrt{B}}(\frac{L_2^2}{\alpha}+L_2)\\
        &+\frac{\sigma}{\alpha\sqrt{L_2\alpha B}}(\frac{L_2L_4}{\alpha}+L_3)\\
        &+\frac{2}{L_2+\alpha} (\frac{L_2L_4}{\alpha}+L_3)\sqrt{\Delta}\\
        &\frac{(1-\frac{2}{L_2+\alpha}\alpha)^\tau}{1-\frac{2}{L_2+\alpha}\alpha - \frac{L_2-\alpha}{L_2+\alpha}}\Big ]+\frac{L_1}{\sqrt{B}}.
    \end{aligned}
\end{equation}

\noindent \textbf{Proof.}  Using the triangle inequality, we have:
\begin{equation}
\begin{aligned}
    &\mathbb{E}\|\frac{\partial l_{{\rm global}} (\Theta_t^0,\mathcal{E}_t^\tau;\mathcal{B}_0)}{\partial \Theta_t^0} - \nabla F(\Theta_t^0)\|\\
    &=\mathbb{E}\|\frac{\partial l_{{\rm global}} (\Theta_t^0,\mathcal{E}_t^\tau;\mathcal{B}_0)}{\partial \Theta_t^0}-\frac{\partial \mathcal{L}_{{\rm global}} (\Theta_t^0,\mathcal{E}_t^\tau)}{\partial \Theta_t^0} \\
    &+\frac{\partial \mathcal{L}_{{\rm global}} (\Theta_t^0,\mathcal{E}_t^\tau)}{\partial \Theta_t^0} - \nabla F(\Theta_t^0)\|\\
    &\leq \mathbb{E}\|\frac{\partial l_{{\rm global}} (\Theta_t^0,\mathcal{E}_t^\tau;\mathcal{B}_0)}{\partial \Theta_t^0}-\frac{\partial \mathcal{L}_{{\rm global}} (\Theta_t^0,\mathcal{E}_t^\tau)}{\partial \Theta_t^0}\|\\
    &+\mathbb{E}\|\frac{\partial \mathcal{L}_{{\rm global}} (\Theta_t^0,\mathcal{E}_t^\tau)}{\partial \Theta_t^0} - \nabla F(\Theta_t^0)\|\\
    &\overset{(i)}{\leq} \frac{L_1}{\sqrt{B}}+\mathbb{E}\|\frac{\partial \mathcal{L}_{{\rm global}} (\Theta_t^0,\mathcal{E}_t^\tau)}{\partial \Theta_t^0} - \nabla F(\Theta_t^0)\|.
\end{aligned}
\end{equation}
where (i) follows from Eq. \ref{cls_bound}.

Then, we need to give an upper bound for $\mathbb{E}\|\frac{\partial \mathcal{L}_{{\rm global}} (\Theta_t^0,\mathcal{E}_t^\tau)}{\partial \Theta_t^0} - \nabla F(\Theta_t^0)\|$. Using:

\begin{equation}
\begin{aligned}
    \nabla F(\Theta_t^0)&=\nabla \mathcal{L}_{{\rm global}}(\Theta_t^0,\mathcal{E}^*(\Theta_t^0)) \\
    &+ \frac{\partial \mathcal{E}^*(\Theta)_t^0}{\partial \Theta_t^0}\nabla_{\mathcal{E}}\mathcal{L}_{{\rm global}}(\Theta_t^0,\mathcal{E}^*(\Theta_t^0)),
\end{aligned}
\end{equation}
and
\begin{equation}
\begin{aligned}
    \frac{\partial \mathcal{L}_{{\rm global}}(\Theta_t^0,\mathcal{E}_t^\tau)}{\partial \Theta_t^0}&=\nabla_{\Theta}\mathcal{L}_{{\rm global}}(\Theta_t^0,\mathcal{E}_t^\tau)\\
    &+\frac{\partial \mathcal{E}_t^\tau}{\partial \Theta_t^0}\nabla_{\mathcal{E}}\mathcal{L}_{{\rm global}}(\Theta_t^0,\mathcal{E}_t^\tau),
\end{aligned}
\end{equation}
we have: 
\begin{equation}
    \begin{aligned}
    &\mathbb{E}\|\frac{\partial \mathcal{L}_{{\rm global}} (\Theta_t^0,\mathcal{E}_t^\tau)}{\partial \Theta_t^0} - \nabla F(\Theta_t^0)\|\\
    &= E\|\nabla_{\Theta}\mathcal{L}_{{\rm global}}(\Theta^{0}_{t},\mathcal{E}^{\tau}_{t}) + \frac{\partial \mathcal{E}_t^\tau}{\partial \Theta_t^0}\nabla_{\mathcal{E}}\mathcal{L}_{{\rm global}}(\Theta^{0}_{t},\mathcal{E}^{\tau}_{t})\\
    &-\nabla\mathcal{L}_{{\rm global}}(\Theta^{0}_{t},\mathcal{E}^*(\Theta^{0}_{t})) \\
    &+  \frac{\partial \mathcal{E}^*(\Theta_t^0)}{\partial \Theta_t^0}\nabla_{\mathcal{E}}\mathcal{L}_{{\rm global}}(\Theta_t^0,\mathcal{E}^*(\Theta_t^0))\| \\
    &\leq \mathbb{E}\|\nabla_{\Theta}\mathcal{L}_{{\rm global}}(\Theta_t^0,\mathcal{E}^\tau)-\nabla\mathcal{L}_{{\rm global}}(\Theta_t^0,\mathcal{E}^*(\Theta_t^0))\| \\
    &+\mathbb{E}\|\frac{\partial \mathcal{E}_t^\tau}{\partial \Theta_t^0}\nabla_{\mathcal{E}}\mathcal{L}_{{\rm global}}(\Theta_t^0,\mathcal{E}^\tau_{t})\\
    &-\frac{\partial \mathcal{E}^*(\Theta_t^0)}{\partial \Theta_t^0}\nabla_{\mathcal{E}}\mathcal{L}_{{\rm global}}(\Theta_t^0,\mathcal{E}^*(\Theta_t^0))\|\\
    &\leq L_2 \mathbb{E}\|\mathcal{E}_t^\tau - \mathcal{E}^*(\Theta_t^0)\| + L_1 \mathbb{E}\|\frac{\partial \mathcal{E}_t^\tau}{\partial \Theta_t^0} - \frac{\partial \mathcal{E}^*(\Theta_t^0)}{\partial \Theta_t^0}\|\\
    &+L_2\mathbb{E}(\|\frac{\partial \mathcal{E}^*(\Theta)_t^0}{\partial \Theta_t^0}\| \|\mathcal{E}_t^\tau - \mathcal{E}^*(\Theta_t^0)\|).
    \end{aligned}
\end{equation}

Now we want to bound $\mathbb{E}\|\frac{\partial \mathcal{E}_t^\tau}{\partial \Theta_t^0} - \frac{\partial \mathcal{E}^*(\Theta_t^0)}{\partial \Theta_t^0}\|$ first. Recall the update method:
\begin{equation*}
    \mathcal{E}_t^j=\mathcal{E}_t^{j-1} - \eta \nabla_{\mathcal{E}}l_{{\rm local}}(\Theta_t^0,\mathcal{E}_t^{j-1};\mathcal{B}_{j-1}),
\end{equation*}
and use the chain rule on it, we have:
\begin{equation}
    \begin{aligned}
        \frac{\partial \mathcal{E}_t^j}{\partial \Theta_t^0}&=\frac{\partial \mathcal{E}_t^{j-1}}{\partial \Theta_t^0}-\eta(\nabla_{\Theta}\nabla_{\mathcal{E}}l_{{\rm local}}(\Theta_t^0,\mathcal{E}_t^{j-1};\mathcal{B}_{j-1})\\
        &+\frac{\partial \mathcal{E}_t^{j-1}}{\partial \Theta_t^0} \nabla^2_{\mathcal{E}}l_{{\rm local}} (\Theta_t^0,\mathcal{E}_t^{j-1};\mathcal{B}_{j-1})).
    \end{aligned}
\end{equation}

For $\mathcal{E}^*(\Theta_t^0)$ is the optimal solution of $\mathcal{L}_{{\rm local}}(\Theta_t^0, \mathcal{E})$, we have $\nabla_{\mathcal{E}}\mathcal{L}_{{\rm local}}(\Theta_t^0, \mathcal{E}^*(\Theta_t^0))=0$. Then, using the chain rule, we have:
\begin{equation}
\begin{aligned}
    &\nabla_{\Theta}\nabla_{\mathcal{E}}\mathcal{L}_{{\rm local}}(\Theta_t^0, \mathcal{E}^*(\Theta_t^0)) \\
    &+ \frac{\partial\mathcal{E}^*(\Theta_t^0)}{\partial \Theta_t^0} \nabla_{\mathcal{E}}^2\mathcal{L}_{{\rm local}}(\Theta_t^0, \mathcal{E}^*(\Theta_t^0))=0. 
\end{aligned}
\end{equation}

Combining Eq. \ref{update_grad} and Eq. \ref{optimal_condition}, the following equation holds:
\begin{equation}
    \begin{aligned}
        \frac{\partial \mathcal{E}_t^j}{\partial \Theta_t^0} -\frac{\partial \mathcal{E}^* (\Theta_t^0)}{\partial \Theta_t^0}&=\frac{\partial \mathcal{E}_t^{j-1}}{\partial \Theta_t^0} -\frac{\partial \mathcal{E}^* (\Theta_t^0)}{\partial \Theta_t^0}\\
        &-\eta(\nabla_{\Theta}\nabla_{\mathcal{E}}l_{{\rm local}}(\Theta_t^0,\mathcal{E}_t^{j-1};\mathcal{B}_{j-1})\\
        &-\nabla_{\Theta}\nabla_{\mathcal{E}} \mathcal{L}_{{\rm local}}(\Theta_t^0,\mathcal{E}^*(\Theta_t^0)))\\
        &-\eta \left(\frac{\partial \mathcal{E}_{t}^{j-1}}{\partial \Theta_t^0} - \frac{\partial \mathcal{E}^* (\Theta_t^0)}{\partial \Theta_t^0} \right) \\
        &\nabla_{\mathcal{E}}^2 l_{{\rm local}}(\Theta_t^0,\mathcal{E}_t^{j-1};\mathcal{B}_{j-1})\\
        &+\eta \frac{\partial \mathcal{E}^*(\Theta_t^0)}{\partial \Theta_t^0} (\nabla^2_{\mathcal{E}}l_{{\rm local}}(\Theta_t^0, \mathcal{E}_t^{j-1};\mathcal{B}_{j-1})\\
        &- \nabla^2_{\mathcal{E}}l_{{\rm local}}(\Theta_t^0,\mathcal{E}^*(\Theta_t^0))).
    \end{aligned}
\end{equation}

Based on \eqref{optimal_condition}, we have:
\begin{equation}
    \| \frac{\partial \mathcal{E}^*(\Theta_t^0)}{\partial \Theta_t^0}\|\leq \frac{L_2}{\alpha}. 
\end{equation}

With the help of Assumption 3, Lemma 2, Eq. \ref{update_optimal_bound} and Eq. \ref{E_grad_bound}, the following bound holds:
\begin{equation}
    \begin{aligned}
        \mathbb{E}&\|\frac{\partial \mathcal{E}_t^j}{\partial \Theta_t^0} - \frac{\partial \mathcal{E}^* (\Theta_t^0)}{\partial \Theta_t^0}\|\\
        &\leq \mathbb{E}(\|I-\eta \nabla_{\mathcal{E}}^2 l_{{\rm local}}(\Theta_t^0,\mathcal{E}_t^{j-1};\mathcal{B}_{j-1})\|     \\
        &\|\frac{\partial \mathcal{E}_t^{j-1}}{\partial \Theta_t^0} - \frac{\partial \mathcal{E}^*(\Theta_t^0)}{\partial \Theta_t^0}\|)\\
        &+\eta \frac{L_2}{\alpha}(\mathbb{E}\|\nabla_{\mathcal{E}}^2l_{{\rm local}} (\Theta_t^0,\mathcal{E}_t^{j-1};\mathcal{B}_{j-1})\\
        &-\nabla_{\mathcal{E}}^2 l_{{\rm local}}(\Theta_t^0,\mathcal{E}^*(\Theta_t^0);\mathcal{B}_{j-1})\|\\
        &+\mathbb{E}\|\nabla_{\mathcal{E}}^2 l_{{\rm local}}(\Theta_t^0,\mathcal{E}^*(\Theta_t^0);\mathcal{B}_{j-1}) \\
        &- \nabla_{\mathcal{E}}^2 \mathcal{L}_{{\rm local}}(\Theta_t^0,\mathcal{E}^*(\Theta_t^0))\|)\\
        &+\eta(\mathbb{E}\|\nabla_{\Theta}\nabla_{\mathcal{E}}l_{{\rm local}}(\Theta_t^0,\mathcal{E}_t^{j-1};\mathcal{B}_{j-1})\\
        &- \nabla_{\Theta}\nabla_{\mathcal{E}}l_{{\rm local}}(\Theta_t^0,\mathcal{E}_t^*(\Theta_t^0);\mathcal{B}_{j-1})\|\\
        &+\mathbb{E}\|\nabla_{\Theta}\nabla_{\mathcal{E}}l_{{\rm local}}(\Theta_t^0,\mathcal{E}_t^*(\Theta_t^0);\mathcal{B}_{j-1}) \\
        &- \nabla_{\Theta}\nabla_{\mathcal{E}}\mathcal{L}_{{\rm local}}(\Theta_t^0,\mathcal{E}_t^*(\Theta_t^0))\|)\\
        &\leq (1-\eta\alpha)\mathbb{E}\|\frac{\partial \mathcal{E}_t^{j-1}}{\partial \Theta_t^0} - \frac{\partial \mathcal{E}^*(\Theta_t^0)}{\partial \Theta_t^0}\| \\
        &+ \frac{\eta}{\sqrt{B}}(\frac{L_2^2}{\alpha}+L_2)\\
        &+ \eta (\frac{L_2 L_4}{\alpha}+L_3)\mathbb{E}\|\mathcal{E}_t^{j-1}-\mathcal{E}^*(\Theta_t^0)\|.
    \end{aligned}
\end{equation}

For the choice of $\eta=\frac{2}{L_2+\alpha}$, Lemma 4 holds. With the result in eq. \ref{tracking_error} and Assumption 5, we have:
\begin{equation}
    \begin{aligned}
    \mathbb{E}\|\mathcal{E}_t^{j-1} - \mathcal{E}^*(\Theta_t^0)\| &\leq \left(\frac{L_2 - \alpha}{L_2 +\alpha}\right)^{j-1}\sqrt{\mathbb{E}\|\mathcal{E}_t^{0} - \mathcal{E}^*(\Theta_t^0)\|} \\
    &+ \frac{\sigma}{\sqrt{L_2\alpha B}}\\
    &\leq \left(\frac{L_2 - \alpha}{L_2 +\alpha}\right)^{j-1}\sqrt{\Delta} + \frac{\sigma}{\sqrt{L_2\alpha B}}.
    \end{aligned}
\end{equation}

Telescoping \eqref{partial_bound} over $j$ from 0 to $\tau$ and combining the result with Eq. \ref{E_bound} yields:
\begin{equation}
    \begin{aligned}
        \mathbb{E}&\|\frac{\partial \mathcal{E}_t^\tau}{\partial \Theta_t^0} - \frac{\partial \mathcal{E}^*(\Theta_t^0)}{\partial \Theta_t^0}\|\\
        &\leq (1-\eta \alpha)^\tau \mathbb{E}\|\frac{\partial \mathcal{E}_t^0}{\partial \Theta_t^0} - \frac{\partial \mathcal{E}^*(\Theta_t^0)}{\partial \Theta_t^0}\|\\
        &+\sum_{j=0}^{\tau-1}(1-\eta \alpha)^{\tau -1-j}\frac{\eta}{\sqrt{B}}(\frac{L_2^2}{\alpha}+L_2)\\
        &+\eta \left(\frac{L_2 L_4}{\alpha}+L_3 \right)\sum_{j=0}^{\tau-1}(1-\eta\alpha)^{\tau-1-j}\\
        &\Big [ \left(\frac{L_2-\alpha}{L_2+\alpha} \right)^j \sqrt{\Delta} + \frac{\sigma}{\sqrt{L_2\alpha B}}\Big ]\\
        &\leq \frac{L_2 (1-\eta \alpha)^{\tau}}{\alpha}+\frac{1}{\alpha\sqrt{B}}(\frac{L_2^2}{\alpha}+L_2)\\
        &+\frac{\sigma}{\alpha\sqrt{L_2\alpha B}}(\frac{L_2L_4}{\alpha}+L_3)\\
        &+\eta (\frac{L_2L_4}{\alpha}+L_3)\sqrt{\Delta} \frac{(1-\eta\alpha)^\tau}{1-\eta\alpha - \frac{L_2-\alpha}{L_2+\alpha}}.
    \end{aligned}
\end{equation}

Take expedition of both sides of Eq. \ref{approximate_bound}, plugging Eq. \ref{E_grad_bound}, Eq. \ref{E_bound} and Eq. \ref{E_tau_bound} into it yields:
\begin{equation}
    \begin{aligned}
        \mathbb{E}&\|\frac{\partial \mathcal{L}_{{\rm global}} (\Theta_t^0,\mathcal{E}_t^\tau)}{\partial \Theta_t^0} - \nabla F(\Theta_t^0)\| \\
        &\leq (L_2 + \frac{L_2^2}{\alpha})\Big [\left(\frac{L_2 - \alpha}{L_2 +\alpha}\right)^{\tau}\sqrt{\Delta} + \frac{\sigma}{\sqrt{L_2\alpha B}}\Big ]\\
        &+L_1\Big [\frac{L_2 (1-\frac{2}{L_2+\alpha} \alpha)^{\tau}}{\alpha}\\
        &+\frac{1}{\alpha\sqrt{B}}(\frac{L_2^2}{\alpha}+L_2)\\
        &+\frac{\sigma}{\alpha\sqrt{L_2\alpha B}}(\frac{L_2L_4}{\alpha}+L_3)\\
        &+\frac{2}{L_2+\alpha} (\frac{L_2L_4}{\alpha}+L_3)\sqrt{\Delta}\\
        &\frac{(1-\frac{2}{L_2+\alpha}\alpha)^\tau}{1-\frac{2}{L_2+\alpha}\alpha - \frac{L_2-\alpha}{L_2+\alpha}}\Big ].
    \end{aligned}
\end{equation}

Finally, plugging Eq. \ref{prop1_prof} into Eq. \ref{prop1_bound} yields Eq. \ref{prop1}. Thus, we complete the proof of Proposition 1.

To deal with multiple updates of the active party with a fixed $t$, we need to use a technology called virtual updates \cite{yang2022fastslowmo}. Specifically, we introduce a virtual parameter $\mathcal{E}_{t,j}^\tau$ which could be obtained by local updates when $\Theta_t^{j}$ is given. With the help of Proposition 1, we could obtain the bound for $\mathbb{E}\|\frac{\partial l_{{\rm global}} (\Theta_t^{j},\mathcal{E}_{t,j}^\tau;\mathcal{B}_{j})}{\partial \Theta_t^{j}} - \nabla F(\Theta_t^{j})\|$. Now, we want to give a bound for $\mathbb{E}\|\frac{\partial l_{{\rm global}} (\Theta_t^{j},\mathcal{E}_{t}^\tau;\mathcal{B}_{j})}{\partial \Theta_t^{j}} - \nabla F(\Theta_t^{j})\|$. The result is as follows:

\noindent\textbf{Proposition 2.} Following the conditions in Proposition 1, we have
\begin{equation}
    \begin{aligned}
        \mathbb{E}&\|\frac{\partial l_{{\rm global}} (\Theta_t^{j},\mathcal{E}_t^\tau;\mathcal{B}_{j})}{\partial \Theta_t^{j}} - \nabla F(\Theta_t^{j})\|\\
        &\leq (L_2 + \frac{L_2^2}{\alpha})\Big [\left(\frac{L_2 - \alpha}{L_2 +\alpha}\right)^{\tau} \\
        &\sqrt{\Delta}+ \frac{\sigma}{\sqrt{L_2\alpha B}}\Big ]+L_1\Big [\frac{L_2 (1-\frac{2}{L_2+\alpha} \alpha)^{\tau}}{\alpha}\\
        &+\frac{1}{\alpha\sqrt{B}}(\frac{L_2^2}{\alpha}+L_2)\\
        &+\frac{\sigma}{\alpha\sqrt{L_2\alpha B}}(\frac{L_2L_4}{\alpha}+L_3)\\
        &+\frac{2}{L_2+\alpha} (\frac{L_2L_4}{\alpha}+L_3)\sqrt{\Delta}\\
        &\frac{(1-\frac{2}{L_2+\alpha}\alpha)^\tau}{1-\frac{2}{L_2+\alpha}\alpha - \frac{L_2-\alpha}{L_2+\alpha}}\Big ]+\frac{L_1}{\sqrt{B}}+L_2 \Delta.
    \end{aligned}
\end{equation}

\noindent\textbf{Proof.} Using the triangle inequality, we have:
\begin{equation}
    \begin{aligned}
        &\|\frac{\partial l_{{\rm global}} (\Theta_t^{j},\mathcal{E}_t^\tau;\mathcal{B}_{j})}{\partial \Theta_t^{j}} - \nabla F(\Theta_t^{j})\|\\
        &\leq \|\frac{\partial l_{{\rm global}} (\Theta_t^{j},\mathcal{E}_t^\tau;\mathcal{B}_{j})}{\partial \Theta_t^{j}} - \frac{\partial l_{{\rm global}} (\Theta_t^{j},\mathcal{E}_{t,j}^\tau;\mathcal{B}_{j})}{\partial \Theta_t^{j}}\|\\
        &+\| \frac{\partial l_{{\rm global}} (\Theta_t^{j},\mathcal{E}_{t,j}^\tau;\mathcal{B}_{j})}{\partial \Theta_t^{j}} - \nabla F(\Theta_t^{j})\|.
    \end{aligned}
\end{equation}

Take the expectation of both sides of Eq. \ref{virtual_decomp}, with the help of Assumptions 2 and 5, the following result holds:
\begin{equation}
    \begin{aligned}
        &\mathbb{E}\|\frac{\partial l_{{\rm global}} (\Theta_t^{j},\mathcal{E}_t^\tau;\mathcal{B}_{j})}{\partial \Theta_t^{j}} - \nabla F(\Theta_t^{j})\|\\
        &\leq \mathbb{E}\|\frac{\partial l_{{\rm global}} (\Theta_t^{j},\mathcal{E}_t^\tau;\mathcal{B}_{j})}{\partial \Theta_t^{j}} - \frac{\partial l_{{\rm global}} (\Theta_t^{j},\mathcal{E}_{t,j}^\tau;\mathcal{B}_{j})}{\partial \Theta_t^{j}}\|\\
        &+\mathbb{E} \| \frac{\partial l_{{\rm global}} (\Theta_t^{j},\mathcal{E}_{t,j}^\tau;\mathcal{B}_{j})}{\partial \Theta_t^{j}} - \nabla F(\Theta_t^{j})\|\\
        &\leq L_2 \|\mathcal{E}_t^{\tau} - \mathcal{E}_{t,j}^\tau \| + \mathbb{E} \| \frac{\partial l_{{\rm global}} (\Theta_t^{j},\mathcal{E}_{t,j}^\tau;\mathcal{B}_{j})}{\partial \Theta_t^{j}} - \nabla F(\Theta_t^{j})\|\\
        &\leq L_2 \Delta + \mathbb{E} \| \frac{\partial l_{{\rm global}} (\Theta_t^{j},\mathcal{E}_{t,j}^\tau;\mathcal{B}_{j})}{\partial \Theta_t^{j}} - \nabla F(\Theta_t^{j})\|.
    \end{aligned}
\end{equation}

Plugging in the result in Proposition 1 into Eq. \ref{prop2_proof} yields Eq. \ref{prop2}. Thus, we complete the proof of Proposition 2.

\noindent\textbf{Proof for Theorem 1.} Based on the $L_0$-smoothness of $F(\Theta)$ established in Lemma 3, we have:
\begin{equation}
    \begin{aligned}
        F&(\Theta_t^{j+1})\leq F(\Theta_t^{j}) + \langle \nabla F(\Theta_t^{j}), \Theta_t^{j+1}-\Theta_t^{j} \rangle\\
        &+\frac{L_0}{2}\|\Theta_t^{j+1}-\Theta_t^{j}\|^2\\
        &\leq F(\Theta_t^{j}) -\eta^\prime \langle \nabla F(\Theta_t^{j}),\frac{\partial l_{{\rm global}}(\Theta_t^{j},\mathcal{E}_t^\tau;B_{j})}{\partial \Theta_t^{j}} \\
        &- \nabla F(\Theta_t^{j})\rangle - \eta^\prime \|\nabla F(\Theta_t^{j})\|^2+\eta^{\prime 2} L_0 \\
        &\|\nabla F(\Theta_t^{j})\|^2 +\eta^{\prime 2} L_0 \|\frac{\partial l_{{\rm global}}(\Theta_t^{j},\mathcal{E}_t^\tau;B_{j})}{\partial \Theta_t^{j}} \\
        &- \nabla F(\Theta_t^{j})\|^2\leq F((\Theta_t^{j}) - (\frac{\eta^\prime}{2} - \eta^{\prime 2}L_0) \\
        &\|\nabla F((\Theta_t^{j})\|^2\|\frac{\partial l_{{\rm global}}(\Theta_t^{j},\mathcal{E}_t^\tau;B_{j})}{\partial \Theta_t^{j}} - \nabla F(\Theta_t^{j})\|^2.
    \end{aligned}
\end{equation}

With the help of Proposition 2, telescoping Eq. \ref{smooth_bound} over $j$ from 0 to $\tau-1$ and taking expectation of both sides yields:
\begin{equation}
    \begin{aligned}
        \mathbb{E}F(\Theta_t^{\tau})&\leq \mathbb{E}F(\Theta_t^{0}) -(\frac{\eta^\prime}{2} - \eta^{\prime 2}L_0) \mathbb{E}\sum_{j=0}^{\tau -1}\|\nabla F(\Theta_t^{j})\|^2\\
        &+(\eta^\prime + 2\eta^{\prime 2}L_0)\tau\{(L_2 + \frac{L_2^2}{\alpha})\Big [\left(\frac{L_2 - \alpha}{L_2 +\alpha}\right)^{\tau} \\
        &\sqrt{\Delta}+ \frac{\sigma}{\sqrt{L_2\alpha B}}\Big ]+L_1\Big [\frac{L_2 (1-\frac{2}{L_2+\alpha} \alpha)^{\tau}}{\alpha}\\
        &+\frac{1}{\alpha\sqrt{B}}(\frac{L_2^2}{\alpha}+L_2)+\frac{\sigma}{\alpha\sqrt{L_2\alpha B}}(\frac{L_2L_4}{\alpha}+L_3)\\
        &+\frac{2}{L_2+\alpha} (\frac{L_2L_4}{\alpha}+L_3)\sqrt{\Delta}\\
        &\frac{(1-\frac{2}{L_2+\alpha}\alpha)^\tau}{1-\frac{2}{L_2+\alpha}\alpha - \frac{L_2-\alpha}{L_2+\alpha}}\Big ]+\frac{L_1}{\sqrt{B}}\}^2\\
        &+ (\eta^\prime + 2\eta^{\prime 2}L_0)L_2^2\Delta^2 (\tau-1).
    \end{aligned}
\end{equation}

Telescoping Eq. \ref{theor1_bound} over $t$ from 0 to $T-1$ yields the following result:
\begin{equation}
    \begin{aligned}
        \frac{1}{\tau T}&\sum_{t=0}^{T-1}\sum_{j=0}^{\tau-1}\|\nabla F(\Theta_t^{j})\|^2 \leq \frac{F(\Theta_0^0) - \inf_{\Theta}F(\Theta)}{\tau T (\frac{\eta^{\prime}}{2} -L_0 \eta^{\prime 2}) }\\
        &+(\eta^\prime + 2\eta^{\prime 2}L_0)L_2^2\Delta^2 \frac{\tau -1}{\tau}\\
        &+(\eta^\prime + 2\eta^{\prime 2}L_0)\tau\{(L_2 + \frac{L_2^2}{\alpha})\Big [\left(\frac{L_2 - \alpha}{L_2 +\alpha}\right)^{\tau} \\
        &\sqrt{\Delta}+ \frac{\sigma}{\sqrt{L_2\alpha B}}]+L_1[\frac{L_2 (1-\frac{2}{L_2+\alpha} \alpha)^{\tau}}{\alpha}\\
        &+\frac{1}{\alpha\sqrt{B}}(\frac{L_2^2}{\alpha}+L_2)+\frac{\sigma}{\alpha\sqrt{L_2\alpha B}}(\frac{L_2L_4}{\alpha}+L_3)\\
        &+\frac{2}{L_2+\alpha} (\frac{L_2L_4}{\alpha}+L_3)\sqrt{\Delta}\\
        &\frac{(1-\frac{2}{L_2+\alpha}\alpha)^\tau}{1-\frac{2}{L_2+\alpha}\alpha - \frac{L_2-\alpha}{L_2+\alpha}}\Big ]+\frac{L_1}{\sqrt{B}}\}^2.
    \end{aligned}
\end{equation}

This is a more general result of Theorem 1. Choose $B=\mathcal{O}(\frac{1}{\sqrt{\epsilon}})$ and $\tau=\mathcal{O}(\log\frac{1}{\epsilon})$, we complete proofs of Theorem 1.

\subsection{Experiment Details}
\label{sec:exp details}
\subsubsection{Datasets}
To fairly evaluate the effectiveness of our proposed Proto-EVFL, we conduct experiments on two widely used image datasets, namely ModelNet-10 \cite{wu20153d} and Fashion-MNIST \cite{xiao2017fashion}, and two popular tabular datasets in VFL, including Credit \cite{misc_default_of_credit_card_clients_350} and Adult \cite{misc_adult_2}, as shown in Tab. \ref{tab:detailed setting}. Among them, ModelNet-10 is a multi-view 3D image dataset. Following the procedure outlined in \cite{liu2022cross}, we created a total of 12 angles for each object. To simulate the VFL, we sequentially divide the views into 4 groups, and each party holds 3 angles of an object. For Fashion-MNIST, each image is divided into four quadrants (2$\times$2), and each party is assigned two of these quadrants, which is consistent with the implementation description in \cite{wu2022practical}. Moreover, we use the tabular data, including Credit and Adult, to further evaluate our methods. Each sample in Credit dataset contains 24 features used for predicting default payments. The Adult dataset is used to predict whether an individual's income exceeds \$50K per year based on 14 attributes. These features in these two datasets are divided into four parts, and each party is randomly assigned two parts. 

\subsubsection{Class imbalanced metric}
\label{sec:detailed class imbalanced formula}
The calculation formula of MID is as follows:
\begin{equation}
    {\rm MID} = \frac{{\rm LRID}}{{\rm LRID}_{extreme}} = \frac{logZ}{N}\sum_{z=1}^{Z}n_{z}\ln\frac{N}{Zn_{z}},
    \label{equ:MID}
\end{equation}
where $N$ refers to the total number of datasets. $n_{c}$ represents the number of samples in class $z$. LRID can be obtained by:
\begin{equation}
    {\rm LRID} = -2\sum^{Z}_{z=1}n_{z}\ln\frac{N}{Zn_{z}}.
\end{equation}
The larger LRID means that the dataset is more class-imbalanced. However, LRID is sensitive to the size of datasets. MID eliminates the impact of the size of dataset and is limited between 0 and 1. The calculation of WCS can be expressed as:
\begin{equation}
\begin{aligned}
    {\rm WCS} &= \sum^{M}_{m=1}\frac{{\|l_{m}\|}_{1}}{{\|L\|}_{1}}{\rm similarity}(L,l_{m})\\
    &= \sum^{M}_{m=1}\frac{{\|l_{m}\|}_{1}L\cdot l_{m}}{{\|L\|}_{1}{\|L\|}_{2}{\|l_{m}\|}_{2}}\\
    &=\frac{1}{{\|L\|}_{1}{\|L\|}_{2}}\sum^{M}_{m=1}\frac{{\|l_{m}\|}_{1}}{{\|l_{m}\|}_{2}}L\cdot l_{m},
    \label{equ:WCS}
\end{aligned}
\end{equation}
where $L$ and $l$ denote the global label distribution vector and local label distribution vector, respectively. ${\|l_{m}\|}_{1}$ refers to the number of samples of party $m$. 

\end{document}